\pdfoutput=1
\documentclass[10pt,twocolumn,letterpaper]{article}

\usepackage{iccv}
\usepackage{times}
\usepackage{times}
\usepackage{epsfig}
\usepackage{graphicx}
\usepackage{amsmath}
\usepackage{amssymb}
\usepackage{graphicx}
\usepackage{color}
\usepackage{algorithm}
\usepackage[noend]{algpseudocode}
\usepackage[font=footnotesize, caption=false]{subfig}
\usepackage[percent]{overpic}
\usepackage{enumitem}
\usepackage{pbox}
\usepackage{ulem}
\usepackage{filecontents}
\usepackage{lscape}
\usepackage[pagebackref=true,breaklinks=true,letterpaper=true,colorlinks,bookmarks=false]{hyperref}

\renewcommand{\etal}{ {\it et al.}}
\newcommand{\etalspace}{ {\it et al.} }

\DeclareMathOperator*{\argmin}{arg\,min}
\DeclareMathOperator*{\argmax}{arg\,max}

\iccvfinalcopy

\def\wacvPaperID{329} % *** Enter the wacv Paper ID here

\ificcvfinal\fi
\begin{document}

%%%%%%%%% TITLE
\title{Fast and robust curve skeletonization for real-world elongated objects}

% Authors at the same institution
%\author{First Author \hspace{2cm} Second Author \\
% Authors at different institutions

\author{Amy Tabb\\
USDA-ARS-AFRS\\
Kearneysville, West Virginia, USA\\
{\tt\small amy.tabb@ars.usda.gov}
\and
Henry Medeiros \\
Marquette University, Electrical and Computer Engineering\\
Milwaukee, Wisconsin, USA \\
{\tt\small henry.medeiros@marquette.edu}
}

\maketitle
\ificcvfinal\thispagestyle{empty}\fi

\newcommand{\diver}{\mathop{\mathrm{div}}\nolimits}

%%%%%%%%% ABSTRACT
\begin{abstract}
We consider the problem of extracting curve skeletons of three-dimensional, elongated objects given a noisy surface, which has applications in agricultural contexts such as extracting the branching structure of plants. We describe an efficient and robust method based on breadth-first search that can determine curve skeletons in these contexts. Our approach is capable of automatically detecting junction points as well as spurious segments and loops. All of that is accomplished with only one user-adjustable parameter. The run time of our method ranges from hundreds of milliseconds to less than four seconds on large, challenging datasets, which makes it appropriate for situations where real-time decision making is needed. Experiments on synthetic models as well as on data from real world objects, some of which were collected in challenging field conditions, show that our approach compares favorably to classical thinning algorithms as well as to recent contributions to the field.\footnote{The citation information for this paper is: A. Tabb and H. Medeiros, ``Fast and robust curve skeletonization for real-world elongated objects", 2018 IEEE Winter Conference on Applications of Computer Vision (WACV), Lake Tahoe, NV/CA.  DOI 10.1109/WACV.2018.00214}\footnote{Mention of trade names or commercial products in this publication is solely for the purpose of providing specific information and does not imply recommendation or endorsement by the U.S. Department of Agriculture.  USDA is an equal opportunity provider and employer.  A. Tabb acknowledges the support of US National Science Foundation grant number IOS-1339211.} 
\end{abstract}

%%%%%%%%% BODY TEXT
\section{Introduction}

The three-dimensional reconstruction of complex objects under realistic data acquisition conditions results in noisy surfaces. We describe a method to extract the curve skeleton of such noisy, discrete surfaces for the eventual purpose of making decisions based on the curve skeleton.  Much work has been done in the computer graphics community on the problem of skeletonization. In general, curve skeletonization converts a 3D model to a simpler representation, which facilitates editing or visualization \cite{Cornea2007Curve} as well as shape searching and structure understanding \cite{Aslan2008Disconnected,Goh2008Strategies,Macrini2008From}. However, the work in the computer graphics community usually assumes noise-less surfaces or surfaces with negligible noise levels.  In this work, we intend to use curve skeletons as an intermediate step between surface reconstruction and computing measurements of branches for automation applications in which robustness to noise and fast execution are important, such as in the automatic modeling of fruit trees in an orchard \cite{tabb2017robotic}.

There are two commonly-used types of skeletons, the medial axis transform (MAT) skeleton, and the curve skeleton. Skeletons using MAT are curves in 2D while in 3D they are locally planar. They allow for the original model to be reconstructed but are very sensitive to local perturbations \cite{Miklos2010Discrete}. Curve skeletons consist of one-dimensional curves for surfaces in 3D, which provides a simpler representation than MAT-type skeletons (see \cite{Cornea2007Curve} for a comprehensive review). However, because there are different definitions of curve skeletons, there is an abundance of methods, with different advantages and disadvantages.

The problem we explore in this paper is to compute curve skeletons of discrete 3D models represented by voxels, which may be sparse, noisy, and are characterized by elongated shapes. The curve skeleton must be thin and one-dimensional except in the case of junction points, which should be detected during the computation of the curve skeleton. The skeleton must also be centered, but because of noise and the use of voxels we use the relaxed centeredness assumption (see \cite{Cornea2007Curve} for more details). To make our approach robust to noise, we identify spurious curve skeleton segments in the course of the algorithm, which removes the need for a separate pruning step \cite{Bai2007Skeleton,SolisMontero2012Skeleton,Ward2010Groupwise}. Finally, since our work is mainly concerned with real trees, branch crossings occur and support poles may be attached to the trees via ties.  As a result, our datasets include loops and cycles and breaking a loop is not desirable. Hence, our method is able to determine that curve skeleton segments which do not terminate at a surface voxel are part of a loop.

We propose a path-based algorithm for the real-time computation of curve skeletons of elongated objects with noisy surfaces that takes into account all of the criteria above. Specifically, our contributions are: 1) our method is robust to noise and there is no requirement for additional pruning, 2) it has time complexity $\mathcal{O}(n^{\frac{7}{3}})$ -- where $n$ is the number of occupied voxels in 3D space -- and as a result is suitable for automation contexts, 3) the method can handle loops, 4) we provide an extensive evaluation on synthetic models as well as on real-world objects, and 5) we provide source code that is publicly available \cite{tabb2017code}.  

\section{Related work}
\label{s:related}
Using laser data, there has been some related work on the problem of extracting cylinders from point cloud data \cite{chaperon2001extracting,Liu2013Cylinder,medeiros2016modeling,rabbani2005efficient,Tran2015Extraction}.  Such methods cannot be used for the datasets we consider because these works assume that the underlying shapes are cylinders.  While elongated shapes may be locally cylindrical, they may have many curves and cylinder fitting may not be appropriate for these shapes.

In ~\cite{Arcelli2011Distance}, the authors compute MAT-style skeletons and combine those ideas with those of traditional thinning, for the purposes of object recognition and classification with an emphasis on reconstruction.  Their algorithm is efficient, but like many other algorithms it depends on a pruning step which requires parameter setting.  Other works which combine the ideas of thinning with computing skeletons are \cite{Couprie2007Discrete}, where a bisector function is used to compute a surface skeleton, and \cite{Bertrand2008Two} where new kernels are used for thinning.

In \cite{Wang2008Curve}, the authors provide an algorithm for computing a curve skeleton from a discrete 3D model, assuming that some noise is present.  This is accomplished by a shrinking step that preserves topology, as well as a thinning step to create 1D structures, and finally a pruning step.  While that method is able to deal with some noise, the shrinking step involved in the algorithm would result in missing some branches with small scale.

A recent approach that is most similar to the method we present was proposed by Jin \etal ~in ~\cite{Jin2014New,jin2016robust}. In these works, curve skeletons are extracted from medical data that contains noise, and once a seed voxel has been identified, new curve skeleton segments are found via searches based on the geodesic distance.  While our method shares a similar overall structure in that paths are iteratively discovered, we do not make assumptions concerning the thickness and lengths of neighboring branches.  In addition, that method was not designed to handle loops. Finally, in that work, the authors note the problems with computational speed in their approach because of the use of geodesic path computations. Our method was conceived to be executed in real-time and is hence computationally inexpensive. 

%%%%%%%%%%%%%%%%%%%%%%%%%%%%%%%%%%%%%%%%%%%%%%%% Method Starts %%%%%%%%%%%%%%%%%%%%%%%%%%%%%%%%%%%%%%%%%%%%%
\section{Method description}
\label{sec:method}

The proposed method to compute a curve skeleton is composed of four main steps:
\begin{enumerate}[leftmargin=*, nolistsep]  
	\item\label{step:vstar} Determine the seed voxel for the search for curve skeleton segments (Section \ref{ss:seepoint});
	\item Determine potential endpoints for curve skeleton segments (Section \ref{ss:add_curves});
	\item Determine prospective curve segments (Section \ref{ss:curve_segs});
	\item Identify and discard spurious segments and detect loops (Section \ref{ss:discard}). 
\end{enumerate}
Step \ref{step:vstar} is executed once at initialization, whereas the remaining steps are executed iteratively until all curve segments have been identified. These steps are described in detail below and the entire process is illustrated in Figure \ref{fig:big_picture_illustration}. In this description, we assume that there is only one connected component, but if there are additional connected components, all four steps are performed for each component.  

\newcommand*{\factorA}{0.16}
\begin{figure*}[!ht]
  \begin{center}
\subfloat[Step 1.1: Compute $d_i$]
{
  \includegraphics[height=\factorA\linewidth]{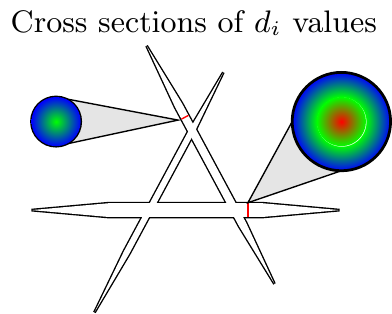} 
  \includegraphics[height=\factorA\linewidth]{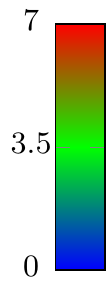}  
\label{subfig:step1_1}
} \hskip 0.4cm
\subfloat[Step 1.2: Locate $v^*$]
{
     \includegraphics[height=\factorA\linewidth]{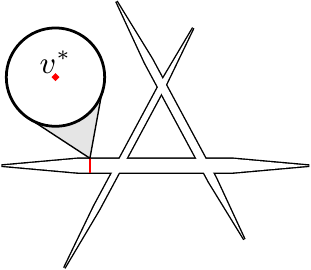} 
\label{subfig:step1_2}
} \hskip 0.4cm
\subfloat[Step 2.1: Compute BFS1 map from $\mathbb{C} = \lbrace v^* \rbrace$]
{
    \includegraphics[height=\factorA\linewidth]{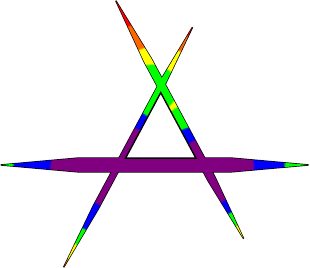} 
  \includegraphics[height=\factorA\linewidth]{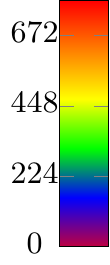}  
\label{subfig:step2_1}
} \hskip 0.4cm
\subfloat[Step 2.2: Locate potential endpoint $v_t$ from BFS1 labels]
{
    \includegraphics[height=\factorA\linewidth]{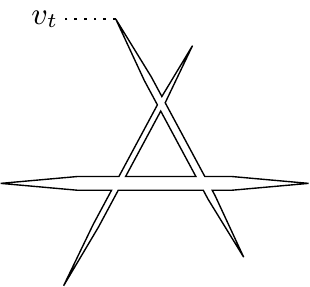} 
\label{subfig:step2_2}
}\\
\subfloat[Step 3.1: Compute BFS2 from $v_t$; black regions are currently unexplored.]
{
 \includegraphics[height=\factorA\linewidth]{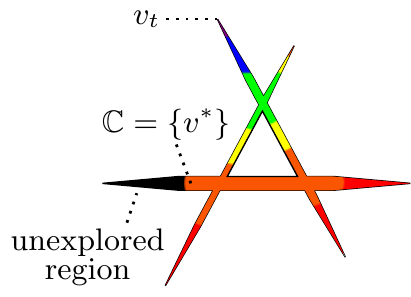}    
\includegraphics[height=\factorA\linewidth]{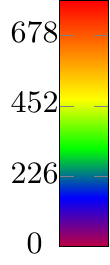}    
\label{subfig:step3_1}
} \hskip 0.4cm
\subfloat[Step 3.2: Compute path from BFS2 labels]
{
\includegraphics[height=\factorA\linewidth]{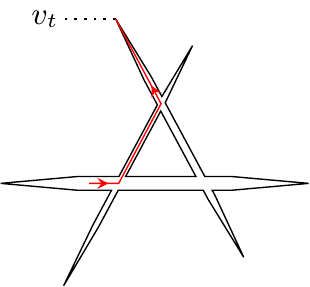}    
\label{subfig:step3_2}
} \hskip 0.4cm
\subfloat[Step 4.1: Accept path if not spurious]
{
\includegraphics[height=\factorA\linewidth]{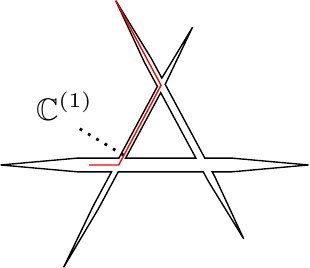}    
\label{subfig:step4_1}
}  \hskip 0.4cm
\subfloat[Step 4.2: Loop processing]
{
   \includegraphics[height=\factorA\linewidth]{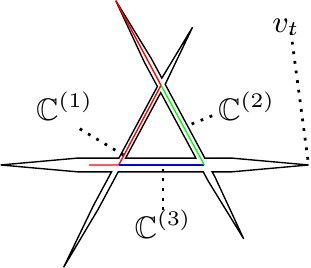}

\label{subfig:step4_2}
} 
\end{center}
    \caption{\textbf{Best viewed in color.} Illustration of the computation of curve skeletons with our proposed method on an artificially created three-dimensional object consisting of three intersecting segments of varying diameters. Legends for colormaps are indicated to the right of the figures. In step 1.1 shown in \ref{subfig:step1_1}, the distance labels $d_i$ are computed; the cross sections in \ref{subfig:step1_1} show the pattern of labels in the object's interior.  Then \ref{subfig:step1_2} shows step 1.2, where $v^*$ is selected from the local maxima of $d_i$; any voxel in the local maxima may be selected. In Step 2.1, the BFS1 map (Section \ref{sss:bfs1}) from $v^*$ given $d_i$ is computed (Figure \ref{subfig:step2_1}). In \ref{subfig:step2_2}, step 2.2, the maximal label from step 2.1 is selected as a proposed endpoint $v_t$. Figure \ref{subfig:step3_1} shows step 3.1 in which we compute the BFS2 labels (Section \ref{sss:bfs2}) from $v_t$ to the current curve skeleton, which is $\mathbb{C} = \lbrace v^* \rbrace$.  Step 3.2 in Figure \ref{subfig:step3_2} consists of tracing the path through the BFS2 labels from $v^*$ to $v_t$.  In step 4.1, we accept the path as part of the curve skeleton if it passes the spurious path test (Figure \ref{subfig:step4_1}). Finally, in Figure \ref{subfig:step4_2} we show the loop handling procedure; $v_t$ is found on the right-hand side, and the loop is incorporated into the curve skeleton. This completes the steps for iteratively adding a curve segment. Steps 2.1 through 4.2 are then repeated until there are no more proposed endpoints that pass the spurious path test.}
\label{fig:big_picture_illustration}
\end{figure*}

%%%%%%%%%%%%%%%%%%%%%%%%%%%%%%%% Preliminaries %%%%%%%%%%%%%%%%%%%%%%%%%%%%%%%%%%%%%%%%%%%%%%%%%%%%%%%%%%%%
\subsection{Preliminaries}
\label{ss:preliminaries}

The set of occupied voxels in 3D space is $\mathbb{V}$, and $n=|\mathbb{V}|$ is the number of occupied voxels. We note here that voxels are defined with respect to a uniform three-dimensional grid, and we let the number of voxels (occupied and empty) in such a grid be $N$.  However, in this work, we only operate on occupied voxels, and since the objects we treat are elongated, $n \ll N$.  In the remainder of the document, \textit{voxels} will mean occupied voxels.  We interpret the voxels as nodes in a graph and assume that voxel labels are binary: occupied or empty.  The edges of the graph are defined by a neighborhood relationship on the voxels.  We represent the set of occupied neighbors for a voxel $v_i$ as $\mathcal{N}_i$. In our implementation we use a 26-connected neighborhood. A surface voxel has $|\mathcal{N}_i| < 26$ and the set of surface voxels is $\mathbb{S} \subseteq \mathbb{V}$.  

Each segment of the curve skeleton is represented by a set of voxels. At each iteration $m$ of the skeletonization algorithm, a new curve skeleton segment $\mathbb{C}^{(m)}$ is discovered. Then, the overall skeleton is represented by the set $\mathbb{C}$ which is the set of skeleton segment sets: 
\begin{equation}
\mathbb{C} = \bigcup_m \lbrace  \mathbb{C}^{(m)} \rbrace.
\end{equation} 

%%%%%%%%%%%%%%%%%%%%%%%%%%%%%%%%%%%%%%%%%%%%%%%%%% BFS algorithm, general %%%%%%%%%%%%%%%%%%%%%%%%%%%%%%%%%%
\subsubsection{\bf Modified breadth-first search algorithm}
\label{bfs_general}

Our skeletonization method is heavily based on a proposed modification of breadth-first search (BFS) which allows one to alter the rate at which nodes are discovered according to a weighting function. We now discuss this BFS modification generally and then show its application to our approach to curve skeletonization in future sections.

In classic BFS, there are three sets of nodes: non-frontier, frontier, and undiscovered nodes and the result of the BFS is a label for each node in the graph. The starting node has a label of 0 and the labels of other nodes are the number of edges that need to be traversed on a shortest path from any node to the starting node. In our modified BFS algorithm, the labels represent the sum of pairwise distances along the shortest path to a given node.

Each voxel $v_i$ also has a weight assigned to it, $w_i$. Voxels with smaller values of $w_i$ are incorporated into the frontier before neighboring voxels with higher weight values.  The speed of discovery of nodes can thereby be altered to favor paths that go through voxels with characteristics which are desirable for a specific purpose as explained in detail in Section \ref{ss:add_curves}.

Our modified BFS is shown in Algorithm \ref{alg:l1norm_bfs}. The frontier for a particular iteration $k$ is $\mathbb{F}^{(k)}$, which is composed of two subsets, $\mathbb{F}_A^{(k)}$ and $\mathbb{F}_B^{(k)}$.  The initialization of $\mathbb{F}_A^{(0)}$ depends on the intended use of the algorithm (details are given in Sections \ref{sss:bfs1} and \ref{sss:bfs2}), and $\mathbb{F}_B^{(0)}$ is always initially empty.  The label of voxels is given by:
\begin{equation}
l_i = \begin{cases} \infty &\mbox{if } v_i \notin \mathbb{F}_A^{(0)} \\
0 & \mbox{if } v_i \in \mathbb{F}_A^{(0)} \end{cases} 
\label{eq:initial_conditions}
\end{equation}

The algorithm progresses as follows.  A voxel $v_i \in \mathbb{F}_A^{(k)}$ has neighbors $v_j$ which had been discovered previously as well as neighbors that were discovered later than itself as determined by the labels of $v_i$ and its neighbors.  In line \ref{alg1:line_foreach2}, only neighbors discovered later than the voxels in $\mathbb{F}_A^{(k)}$ are updated based on the label of $v_i$, the weight $w_i$, and the distance between the neighboring voxels. The set $\mathbb{N}^{(k)}$ in line \ref{alg1:set_N} is the set of frontier candidates beyond the current frontier at iteration $k$.  $\mathbb{N}^{(k)}$ is used to select a label threshold, $l_{min}$.  If the voxels in $\mathbb{F}^{(k)}$ have a label greater than this threshold, they remain in the frontier (specifically $\mathbb{F}_B^{(k + 1)}$) for the next iteration. If a voxel $v_i$ in $\mathbb{F}^{(k)}$ has a label smaller than this threshold, then those neighbors of $v_i$ which were discovered later than $v_i$ are placed in $\mathbb{F}_A^{(k + 1)}$ and $v_i$ is removed from the frontier for the next iteration.  The distinction between $\mathbb{F}_A^{(\cdot)}$ and $\mathbb{F}_B^{(\cdot)}$ allows for a more efficient implementation because only labels in $\mathbb{F}_A^{(\cdot)}$ must be updated in line \ref{alg1:set_lj}.

\begin{algorithm}
\caption{Modified BFS Algorithm} \label{alg:l1norm_bfs}
\begin{algorithmic}[1]
\Require{Set of occupied voxels $\mathbb{V}$, initial frontier voxels $\mathbb{F}_A^{(0)}$, voxel weights $w_i$, initial voxel labels $l_i$}
\Ensure{Updated voxel labels $l_i$}
\State {$k=0$, $\mathbb{F}_B^{(0)} = \emptyset$}
\While {$|\mathbb{F}_A^{(k)}| > 0$} \label{alg1:line_foreachF2}
\For {each voxel $v_i \in \mathbb{F}_A^{(k)}$} \label{alg1:line_foreach1}
\For {each voxel $v_j \in \mathcal{N}_i$ such that $(l_j > l_i)$} \label{alg1:line_foreach2} 
\State {$l_j = \min(l_j, l_i + w_j + ||v_j - v_i||)$} \label{alg1:set_lj}
\EndFor
\EndFor
\State {$\mathbb{F}^{(k)} = \mathbb{F}_A^{(k)} \cup \mathbb{F}_B^{(k)}$}
\State \parbox[t]{\dimexpr\linewidth-\algorithmicindent}{$\mathbb{N}^{(k)} = \lbrace v_j | l_j > l_i, \forall v_j \in \mathcal{N}_i, \forall v_j \notin \mathbb{F}^{(k)}, \forall v_i \in \mathbb{F}^{(k)}\rbrace$ \strut} \label{alg1:set_N}
\State {$l_{min} = \min_{v_j \in \mathbb{N}^{(k)}} l_j$} \label{alg1:set_lmin}
\State {$\mathbb{F}_A^{(k + 1)} = \lbrace v_j | l_i < l_{min}, \forall v_i \in \mathbb{F}^{(k)}, \forall v_j \in \mathbb{N}^{(k)} \rbrace$ \strut} \label{alg1:set_f2prime}
\State \parbox[t]{\dimexpr\linewidth-\algorithmicindent}{$\mathbb{F}_B^{(k + 1)} = \lbrace v_i | l_i \geq l_{min}, \forall v_i \in \mathbb{F}^{(k)}, |\mathcal{N}_i \cap \mathbb{N}^{(k)}| > 0\rbrace$ \strut} \label{alg1:set_f1prime}
\State {$k = k + 1$}
\EndWhile
\end{algorithmic}
\end{algorithm}

%%%%%%%%%%%%%%%%%%%%%%%%%%%% Compute distance labels %%%%%%%%%%%%%%%%%%%%%%%%%%%%%%%
\subsection{Determination of the seed voxel $v^*$}
\label{ss:seepoint}

As mentioned above, the first step of our skeletonization algorithm is the determination of the seed voxel. This step consists of two sub-steps: computation of distance labels, and localization of an extreme point as explained below.

\subsubsection{Distance label computation}
To compute the distance labels $d_i$, ${i=0,...,n-1}$, we compute the distance transform using Euclidean distances so that $d_i$ represents the  distance from $v_i$ to the closest voxel in $\mathbb{S}$, i.e., a surface voxel. To compute the distance labels efficiently, we use the linear-time algorithm of ~\cite{meijster2002general} on the occupied voxels in our graph representation (note that the pseudo-code in \cite{meijster2002general} considers regular grids instead). In addition, in our implementation, the three scans of the algorithm are executed in parallel.

\subsubsection{Finding the seed voxel $v^*$}
\label{ss:seed}
Once the distances $d_i$ are computed, we find the voxels with maximum distance label, $d_{max}$, which ensures that the curve skeleton goes through the thickest part of the object. There may be several voxels with $d_i = d_{max}$, and we arbitrarily select one of them to be $v^*$. If a point of the curve skeleton is known to be a desirable seed point for a specific application, that point may be selected to serve as $v^*$ without affecting the subsequent steps we describe in this paper.

%%%%%%%%%%%%%%%%%%%%%%%%%%%%%%%%%%%%%%%%%%%%%%%%%%%%%%%%%%%
\subsection{Determination of endpoint candidates}
\label{ss:add_curves}
%
%%%%%%%%%%%%%%%%%%%%%%%%%%%%%% initial conditions %%%%%%%%%%%%%%%%%

This section describes the first step to determine the curve skeleton segments $\mathbb{C}^{(m)}$: the identification of the endpoints of prospective segments that are connected to existing segments in the skeleton. This is done in two sub-steps. First, we compute the breadth-first search distances from the existing curve segments to potential endpoints. We call this step BFS1. Then, a candidate endpoint is identified from the extreme points in this set. These sub-steps are described in detail below.

%%%%%%%%%%%%%%%%%%%%%%%%%%% BFS1 %%%%%%%%%%%%%%%%%%%%%%%%%%%%%
\subsubsection{BFS1 Step}
\label{sss:bfs1}

At the first iteration of our algorithm, the curve skeleton consists of a single voxel,  $v^*$. We initialize $\mathbb{F}_A^{(0)} = \lbrace v^* \rbrace$ in Algorithm \ref{alg:l1norm_bfs}, and initialize the labels as in Equation \ref{eq:initial_conditions}. We then perform Algorithm \ref{alg:l1norm_bfs} using weights $w_i=d_{max}-d_i$, where $d_i$ and $d_{max}$ are the distance labels and the maximum distance label, respectively (shown in Figure \ref{subfig:step1_1}). These weights increase linearly according to a voxel's Euclidean distance to a surface voxel, i.e., surface voxels have $w_i = d_{max}$.  The overall effect of weighting the search in such a way is that paths which pass through the center of the object are explored first. This procedure finds the distances from each voxel to the existing curve skeleton along a centered path. As explained in detail below, points with maximal distance are endpoint candidates. 

At each subsequent iteration of the algorithm, new curve skeleton segments (which are identified as described in Sections \ref{ss:curve_segs} and \ref{ss:discard} below) are added to $\mathbb{F}_A^{(0)}$ and the BFS1 labels are updated accordingly. For improved efficiency, on subsequent iterations, BFS1 labels are simply updated instead of computed from scratch. Suppose that on iteration $m$ the set of approved curve skeleton segment voxels is $\mathbb{C}^{(m-1)}$, so that $\mathbb{F}_A^{(0)} =\mathbb{C}^{(m-1)}$.  We leave the existing BFS1 labels from iteration $m-1$ unchanged, except for those in $\mathbb{F}_A^{(0)}$:
\begin{equation}
l_i^{(m)} = \begin{cases} l_i^{(m - 1)} &\mbox{if } v_i \notin \mathbb{F}_A^{(0)} \\
0 & \mbox{if } v_i \in \mathbb{F}_A^{(0)} \end{cases} 
\label{eq:initial_conditions_BFS1}
\end{equation}
Then Algorithm \ref{alg:l1norm_bfs} progresses as usual given these labels.

%%%%%%%%%%%%%%%%%%%%%%%%%%%%%%%%%%%%%%%%%%%% tips %%%%%%%%%%%%%%%%%%%%%%%%%%%%%
\subsubsection{Identification of an endpoint candidate $v_t$ from BFS1}

We next identify candidate endpoint voxels of the curve skeleton.  A candidate endpoint voxel is a surface voxel which is not yet connected to the curve skeleton. An endpoint candidate is given by the voxel $v_t \in \mathbb{S}$ such that the label of $v_t$ is greater than or equal to any other BFS1 label for any other surface voxels, i.e., 
\begin{equation}
v_t = \underset{v_{i}\in\mathbb{S}}{\argmax}\left(l_i\right),
\label{eq:vt}
\end{equation}
where $l_i$ is the BFS1 label of $v_i$. There may be multiple voxels with the same maximum label value.  As in the seed voxel selection step in Section \ref{ss:seed}, $v_t$ may be chosen arbitrarily from the set of voxels with the maximum label value.

\subsection{Determination of prospective curve segments}
\label{ss:curve_segs}

The existing curve skeleton might be reachable from a proposed endpoint $v_t$ via more than one path (see Figure \ref{subfig:step4_2}, for example). We use the breadth-first search distance from the prospective endpoint to the existing curve skeleton to identify those branches and junction points. This is also done in two sub-steps. First, we compute the BFS distances from $v_t$, which we call the BFS2 step. Then we determine the curve skeleton segments by analyzing connected components. These sub-steps are explained in detail below.

\subsubsection{BFS2 Step}
\label{sss:bfs2}

In order to determine where a proposed curve skeleton segment intersects with the existing curve skeleton, we use Algorithm \ref{alg:l1norm_bfs} with weights $w_i= d_{max}-d_i$, as in the BFS1 step. Unlike the BFS1 step, however, for each iteration of the algorithm, the labels are now initialized using Equation \ref{eq:initial_conditions} with $\mathbb{F}_A^{(0)} = \lbrace v_t \rbrace$. When an existing curve skeleton section is encountered, the search is stopped for that region.  The output of this step are the BFS2 labels.

Once the BFS2 labels are computed, the frontier arcs are then analyzed and grouped by connected components.  The number of connected components in the frontier voxel set is the number of curve skeleton paths from $v_t$ to the existing curve skeleton.

\subsubsection{Identification of curve skeleton segments from BFS2}
\label{sss:paths}

Let a frontier connected component (FCC) from BFS2 be the set of voxels $\mathbb{F}_{CC}$.  We determine the voxels in $\mathbb{F}_{CC}$ that are neighbors of the existing curve skeleton $\mathbb{C}$ and denote these voxels as $\mathbb{F}_{CSN}$: 
\begin{equation}
\mathbb{F}_{CSN} = \lbrace v_i | (v_i \in \mathbb{F}_{CC}) \wedge (\exists v_j \in \mathcal{N}_i) \wedge (v_j \in \mathbb{C}) \rbrace.
\label{eq:fcsn}
\end{equation}
From $\mathbb{F}_{CSN}$ we determine the voxel with the smallest BFS2 label in the set and denote this voxel as $v_{s, 1}$, i.e., 
\begin{equation}
v_{s,1} = {\underset{v_{i}\in\mathbb{F}_{CSN}}{\argmin}}\left(l_i\right),
\label{eq:v_s1}
\end{equation}
where $l_i$ is the BFS2 label. As before, there may be many voxels with the same smallest label, and one may be chosen arbitrarily. The next step is to determine the path from $v_{s,1}$ to $v_t$ such that the path is centered.  We accomplish this with the BFS2 labels as well as the distance transform labels $d_i$.  This combination improves centeredness on curved portions as compared to only using BFS2 labels.

\begin{algorithm}[t]
\caption{Determination of curve skeleton segment from BFS2 and $d_i$} \label{alg:pseudo_code_path_trace}
\begin{algorithmic}[1]
\Require{Set of occupied voxels $\mathbb{V}$, BFS2 voxel labels $l_i$, voxel distance transforms $d_i$, proposed endpoint $v_t$, voxel in the existing curve skeleton $v_{s,1}$}
\Ensure{Curve skeleton segment $\mathbb{C}^{(m)}$}
\State {$v_c = v_{s,1}$}
\State {$\mathbb{C}^{(m)} = \lbrace v_{s,1} \rbrace$}
\While {$(v_c \neq v_t) \wedge (v_c \notin \mathbb{C})$}
\State \parbox[t]{\dimexpr\linewidth-\algorithmicindent}{Determine  $d^* = \underset{v_{i}\in\mathcal{N}_c \wedge l_c > l_i}{\max}\left(d_i\right)$ where $d_i$ is the distance transform of $v_i$ \strut}
\State \parbox[t]{\dimexpr\linewidth-\algorithmicindent}{Compute $v_n = \underset{v_{j}\in\mathcal{N}_c^*}{\argmin}\left(l_j\right)$ where $l_j$ is the BFS2 label of $v_j$ and $\mathcal{N}_c^* = \left\{v_j| v_j \in \mathcal{N}_c, d_j=d^* \right\} $ \strut}
\State {$\mathbb{C}^{(m)} = \mathbb{C}^{(m)} \cup \lbrace  v_n \rbrace$}
\State {$v_c = v_n$}
\EndWhile
\end{algorithmic}
\end{algorithm}
The process of determining a new curve skeleton segment is sketched in Algorithm \ref{alg:pseudo_code_path_trace}. The sequence of current voxels $v_c$ creates a new curve skeleton segment.  We start from the neighbor of the existing curve skeleton, $v_{s,1}$, and set $v_c$ equal to $v_{s,1}$.  We determine $d^*$ by examining $v_c$'s neighbors and the BFS2 labels.  Let the BFS2 label of $v_c$ be $l_c$.  $d^*$ is the maximum distance label of  $v_c$'s neighbors whose distance labels $l_i$ are less than $l_c$.  Then $v_n$ is determined as the voxel, out of $v_c$'s neighbors, with the smallest BFS2 label among those voxels with distance transform label equal to $d^*$.  The practical ramifications for these choices are as follows.  By choosing the next voxel $v_n$ as a voxel with a smaller BFS2 label value than $v_c$, we are guaranteed to be moving towards $v_t$ within the voxels that are occupied.  Secondly, by requiring that the next distance $d_n = d^*$, we are choosing the voxel most in the center of all of the neighboring voxels that are on a path towards $v_t$ in the occupied voxels. Algorithm \ref{alg:pseudo_code_path_trace} is performed for all FCCs.

\subsection{Identification of spurious segments and loops}
\label{ss:discard}

Due to the noisy nature of our surfaces, it is necessary to reject some of the curve skeleton segments identified in the previous step. In the next section, we describe our approach to classify curve skeleton segments as spurious or non-spurious using the frontier voxels from the BFS2 step.  When the number of FCCs is one, the proposed curve segment $\mathbb{C}^{(m)}$ undergoes the spurious curve segment classification described in the next section. When the number of FCCs is greater than one, a loop is present, and this case is handled using the approach described in Section \ref{sss:loop}.

\subsubsection{Spurious curve segment classification}
\label{sss:spurious}
Our classification approach assumes that the surface voxels are disturbed by additive three-dimensional Gaussian noise $\eta \sim (\mu, \Sigma)$ and checks whether the endpoint of the proposed curve segment $v_t$ belongs to this distribution.\footnote{Although we used a Gaussian model for mathematical convenience, our experiments showed that the real noise distribution has little impact on the performance of our approach. That is the case for real-world data or even when a substantial amount of shot-like noise is introduced into the models (See Figures \ref{fig:resultsB_detail}, \ref{fig:summed_error}, and the figures in the supplementary materials). } If it does not, the segment is considered spurious. To compute this distribution, note that $\mathbb{F}_{CC}$ is composed of interior and surface voxels.  Let the set of surface voxels from $\mathbb{F}_{CC}$ be $\mathbb{F}_{S}$, and let $v_{s, 0}$ be the closest voxel from the existing skeleton to the proposed segment. For each voxel $v_j \in \mathbb{F}_{S}$, we determine the difference vector $v_j^\prime =  v_j - v_{s,0}$. Then, the sample mean and the sample covariance of $\eta$ are given by $\mu=\frac{1}{N}\sum_{i=1}^{N}v_{j}^{\prime}$ and $\Sigma=\frac{1}{N}\sum_{i=1}^{N}\left(v_{j}^{\prime}-\mu\right)\left(v_{j}^{\prime}-\mu\right)^{T}$ where $N=\left|\mathbb{F}_{S}\right|$.

The squared difference vectors $||v_j^\prime||^2$ are $\chi^2$-distributed with three degrees of freedom.  In order to classify a curve segment $\mathbb{C}^{(m)}$, we compute the probability that the shifted segment endpoint $v_t^\prime =  v_t - v_{s,0}$ belongs to this $\chi^2$-distribution.  Let $x = (v_t^\prime - \mu)^T\Sigma^{-1}(v_t^\prime  - \mu)$, its probability density function is given by
\begin{equation}
f(x) = \frac{x^{1/2} e^{-x/2}}{2^{3/2} \Gamma(3/2)}.
\label{eq:chi_square}
\end{equation}
If $f(x) > t$, where $t$ is a user-supplied acceptance probability, then the curve segment's tip $v_t$ is considered part of the surface voxel's distribution and consequently discarded as a spurious segment.  Otherwise, $\mathbb{C}^{(m)}$ is incorporated into the existing curve skeleton. This approach allows curve segments to be classified as spurious or not depending on local conditions and not on absolute parameters.

\subsubsection{Loop handling for multiple frontier connected components}
\label{sss:loop}

The presence of multiple FCCs indicates that one or more tunnels are present in the surface, which corresponds to one or more loops in the curve skeleton.  In this case, we do not pursue the spurious curve classification step and instead handle the loop first (Algorithm \ref{alg:loop_handling}); once the loop has been located, the algorithm returns to spurious curve segment classification (\S \ref{sss:spurious}). Let the $j$-th FCC be represented by the set of voxels $\mathbb{F}_{CC, j}$, where $j=0,1,...,a-1$ and $a$ is the number of FCCs.  Then for each $\mathbb{F}_{CC, j}$, we compute a proposed curve skeleton segment using Algorithm \ref{alg:pseudo_code_path_trace}, and let this proposed curve skeleton segment be denoted $\mathbb{C}^{\prime}_j$.  The proposed curve skeleton segments $\mathbb{C}^{\prime}_j$s have some voxels in common.  In particular, all of the proposed segments travel through the region near the tip $v_t$, which is a surface voxel.  We remove this common region near the tip from all of the $\mathbb{C}^{\prime}_j$s (lines \ref{line:compute_common}, \ref{line:remove_common}).  Then, the $\mathbb{C}^{\prime}_j$s are processed such that there are no redundancies between $\mathbb{C}^{\prime}_j$s (line \ref{line:remove_redundacies}) and added to the set of curve skeleton segments $\mathbb{C}$ (line \ref{line:add_to_set}). A figure illustrating this process is in the supplemental materials.

\begin{algorithm}
\caption{Loop handling} \label{alg:loop_handling}
\begin{algorithmic}[1]
\Require{Proposed skeleton segments with common voxels $\mathbb{C}^{\prime}_j$, number of FCCs $a$}
\Ensure{Set of skeleton segments $\mathbb{C}$ with disjoint loop segments}
\State {$\mathbb{C}_t = \cap_{j \in [0, a-1]} \mathbb{C}^{\prime}_j$}\label{line:compute_common}
\For {$j = 0$ to ${a-1}$}
\State {$\mathbb{C}^{\prime}_j =\mathbb{C}^{\prime}_j \setminus \mathbb{C}_t$}\label{line:remove_common} 
\For {$k = j + 1$ to ${a-1}$}
\State {$\mathbb{C}^{\prime}_j =\mathbb{C}^{\prime}_j \setminus \mathbb{C}_k$} \label{line:remove_redundacies} 
\EndFor
\State {$\mathbb{C} = \mathbb{C} \cup \left\{ \mathbb{C}^{\prime}_j \right\} $}\label{line:add_to_set} 
\EndFor 
\end{algorithmic}
\end{algorithm}
\newcommand*{\wLH}{0.8}
Algorithm \ref{alg:pseudo_code_step_add_curve_segments} summarizes the complete proposed approach. Line \ref{alg4line1} corresponds to the seed localization step performed at initialization as described in Section \ref{ss:seepoint}. Lines \ref{alg4line3}-\ref{alg4line4} show the iterative endpoint localization method presented in Section \ref{ss:add_curves}. The determination of prospective segments of Section \ref{ss:curve_segs} is carried out by Lines \ref{alg4line5}-\ref{alg4line6}. Finally, lines \ref{alg4line7}-\ref{alg4line10} perform the spurious segment classification and loop handling routines described in Section \ref{ss:discard}.
\begin{algorithm}[t]
\caption{Proposed skeletonization algorithm} \label{alg:pseudo_code_step_add_curve_segments}
\begin{algorithmic}[1]
\Require{Set of occupied voxels $\mathbb{V}$ representing the object of interest, user-supplied acceptance probability $t$}
\Ensure{Object skeleton $\mathbb{C}$}
\State Determine seed voxel $v^*$ and make initial curve skeleton $\mathbb{C}^{(0)}=v^*$ \label{alg4line1}
\Repeat 
\State \parbox[t]{.90\dimexpr\linewidth-\algorithmicindent}{Update BFS1 labels using Alg. \ref{alg:l1norm_bfs} with $\mathbb{F}_A^{(0)} = \mathbb{C}^{(m-1)}$\strut}\label{alg4line3}
\State {Locate a proposed endpoint $v_t$ from BFS1}\label{alg4line4}
\State \parbox[t]{.90\dimexpr\linewidth-\algorithmicindent}{Create BFS2 labels using  Alg. \ref{alg:l1norm_bfs}  with $\mathbb{F}_A^{(0)} = \lbrace v_t \rbrace$\strut}\label{alg4line5}
\State \parbox[t]{.90\dimexpr\linewidth-\algorithmicindent}{Create $\mathbb{C}^{(m)}$ by tracing paths in BFS2 labels from existing curve skeleton $v_t$ according to Alg. \ref{alg:pseudo_code_path_trace}\strut}\label{alg4line6}
\If {(Number of FCCs == 1)}\label{alg4line7}
\State \parbox[t]{.90\dimexpr\linewidth-\algorithmicindent}{Accept or decline curve skeleton segments using the method of Section \ref{sss:spurious} according to the acceptance probability parameter $t$.  If accepted $\mathbb{C} = \mathbb{C} \cup \left\{ \mathbb{C}^{(m)} \right\} $ and $m = m  + 1$\strut}
\Else {}
\State {Check for loops using Alg. \ref{alg:loop_handling}.}\label{alg4line10}
\EndIf
\Until{No more endpoint hypotheses are found}
\end{algorithmic}
\end{algorithm}
The computational complexity of the entire method is $\mathcal{O}(n^{\frac{7}{3}})$. As a matter of fact, the method runs in $\mathcal{O}(||V_t|| \times d_{max} \times n)$, where $d_{max}$ is the largest voxel to surface distance, and $||V_t||$ is the number of proposed endpoints $v_t$. Since $||V_t||$ and $d_{max}$ tend to be orders of magnitude smaller than $n$ for elongated objects, the method runs extremely fast in practice. A detailed analysis of the computational complexity as well as a discussion of the topological stability of the proposed approach are included in the supplementary materials.

\section{Experiments}
\label{sec:exp}
\newcommand*{\factorTreeC}{0.15}
% trim left bottom right top
We evaluated our method and four comparison approaches that reflect the state of the art on curve skeletonization on real datasets consisting of nine different trees, denoted as trees A - I. These trees are real-world objects with an elongated shape. Most of the trees are three meters or taller, and the surfaces are noisy. The data for six out of the nine trees was acquired outdoors, and the reconstructions were generated using the method of \cite{Tabb2013Shape}, but other reconstruction algorithms could be applied (e.g., \cite{Pound2016Patch}). Table \ref{table:datasets} lists the main characteristics of the nine datasets. All five methods were evaluated with respect to their accuracy and robustness to noise as well as run times. We additionally performed a qualitative evaluation of the performance of our method in non-elongated synthetic models commonly used in the evaluation of skeletonization algorithms.

%=================================================================
\begin{table}  %[h]
\caption{Characteristics of the nine datasets used in our evaluation. $n$ is the number of occupied voxels/nodes, $N$ is the number of voxels in the grid, $d_{max}$ is the largest voxel to surface distance, and $||\mathbb{V}_t||$ is the number of proposed endpoints $v_t$.}
\begin{center}
\resizebox{0.8\linewidth}{!}
{
\begin{tabular}{lcccr}
\hline 
%\multicolumn{5}{c}{} \\
ID & $n$ & $N$ & $d_{max}$ & $||\mathbb{V}_t||$ \\
\hline \hline
A & \phantom{0}55,156 &  27,744,000 & 16 & 15  \\
\hline
B & \phantom{0}88,407 & 56,832,000 & 5 & 174  \\
\hline
C & \phantom{0}88,798 & 45,240,000 & 7 & 145  \\
\hline
D &  \phantom{0}92,892  & 80,640,000 & 6 & 154 \\
\hline 
E & \phantom{0}98,228  & 58,464,000 & 9 & 50 \\
\hline
F & 136,497 & 80,640,000 & 7 & 134   \\
\hline 
G & 158,686  & 64,512,000 & 9 & 128 \\
\hline
H & 176,820  & 80,640,000 & 10 & 97 \\
\hline 
I & 246,654  & 80,640,000 & 10 & 83 \\
\hline 
\end{tabular}
}
\end{center}
\label{table:datasets}
\end{table} 

The first comparison method is the classical medial-axis thinning algorithm of \cite{Lee1994Building}, which maintains the Euler characteristics of the object during its execution.  The second and third comparison methods, denoted PINK skel and PINK filter3d, are also medial-axis type thinning approaches based on the discrete bisector function \cite{Couprie2007Discrete} and critical kernels \cite{Bertrand2008Two}. The fourth comparison method is the approach of Jin \etal ~\cite{Jin2014New,jin2016robust}, discussed in Section \ref{s:related}.

We implemented our method in C/C++ on a machine with a 12 core Intel Xeon(R) 2.7 GHz processor and 256 GB RAM.\footnote{The source code is available at \cite{tabb2017code}.} For all the results shown in this section, the spurious branch probability, the only parameter of our algorithm, was set to $t= 1e^{-12}$. The implementation of the thinning algorithm from \cite{Lee1994Building} is provided through Fiji/ImageJ2 in the Skeletonize3D plugin, authored by Ignacio Arganda-Carreras. The implementations of \cite{Couprie2007Discrete} and \cite{Bertrand2008Two} were provided by the scripts `skel' and `skelfilter3d', respectively, from the PINK library \cite{Pink}. The implementation of the method of Jin \etal ~was kindly provided by the authors.  We did try to evaluate our datasets using the curve skeleton algorithm and implementation of ~\cite{Cornea2005Computing}, but that approach failed to return a skeleton.  We hypothesize that this failure is a result of the thinness of some of the structures in our datasets, which are sometimes only one voxel wide, since the discussion in \cite{Cornea2005Computing} specifically mentions that the algorithm may fail for thin structures.

\subsection{Accuracy and robustness to noise} \label{r1}

\newcommand*{\factorTreeB}{0.15}

In order to illustrate the accuracy of our method in comparison with the state-of-the-art approaches,  Figure \ref{fig:resultsB_detail} shows the original surface and curve skeletons computed with all five methods for Dataset B. As expected, the thinning algorithm, PINK skel, and PINK filter3d methods were not able to deal adequately with the noise in our datasets, and created many extra, small branches. In addition, the PINK filter3d method removes some branches. Jin \etal's method performed better than the thinning algorithms with respect to noise, although it still presented some small spurious branches. In addition, this method is unable to deal with loops or cycles in the original structure. Our method is robust to the noise in our datasets and also was able to deal with loops in the curve skeleton. High-resolution images and results for the other datasets are also available in the supplementary materials.

\renewcommand*{\factorTreeB}{0.15}
\newcommand{\tree}{1.65cm}
\newcommand{\lhs}{0cm}
\newcommand{\bt}{0.5cm}
\newcommand{\rhs}{0cm}
\newcommand{\toptrim}{0.5cm}

%trim = \lhs \bt \rhs \toptrim, clip=true
\begin{figure}[ht!]
\captionsetup[subfloat]{width=1.1cm}
\centering
\subfloat[Surface]  
	{  
\includegraphics[trim=2cm 5cm 2cm 5cm, clip=true, totalheight=\tree]{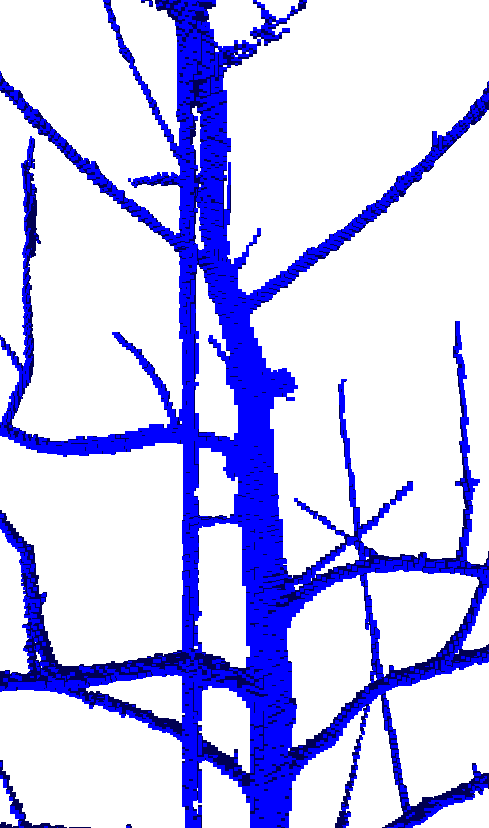}
}
\subfloat[Thinning]  
	{  
	\includegraphics[trim=2cm 5cm 2cm 5cm, clip=true, totalheight=\tree]{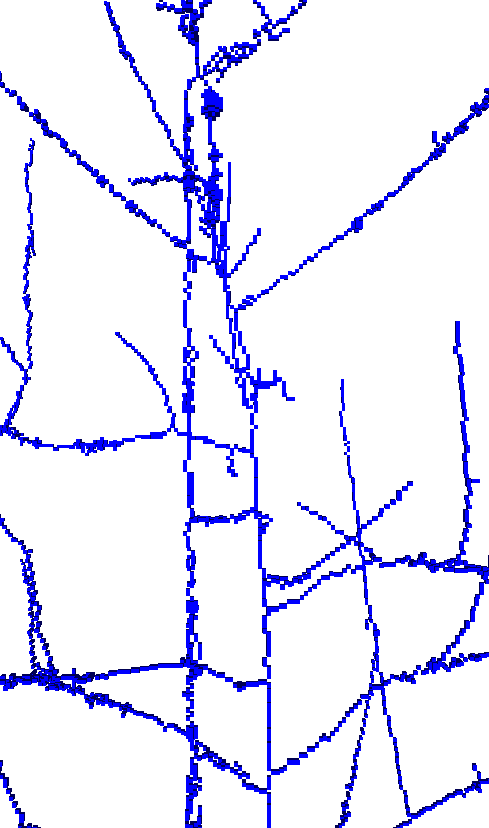}
}
\subfloat[PINK skel]  
	{  
	\includegraphics[trim=2cm 5cm 2cm 5cm, clip=true, totalheight=\tree]{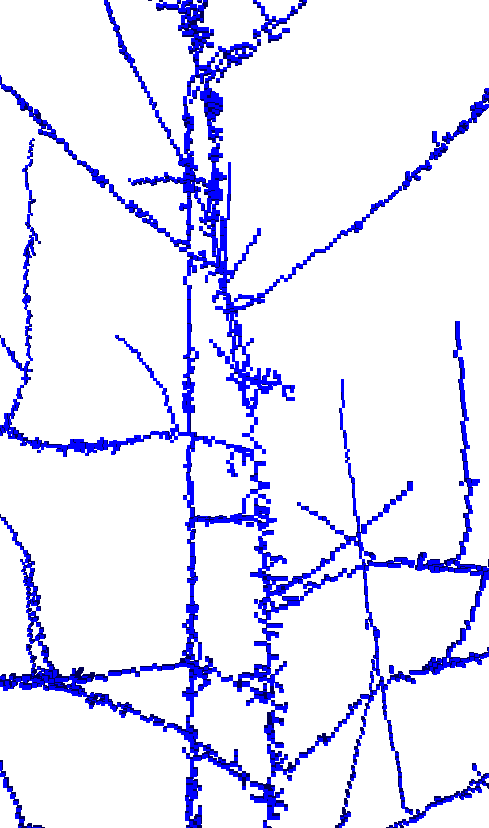}
} 
\subfloat[PINK filter3d]  
{  
	\includegraphics[trim=2cm 5cm 2cm 5cm, clip=true, totalheight=\tree]{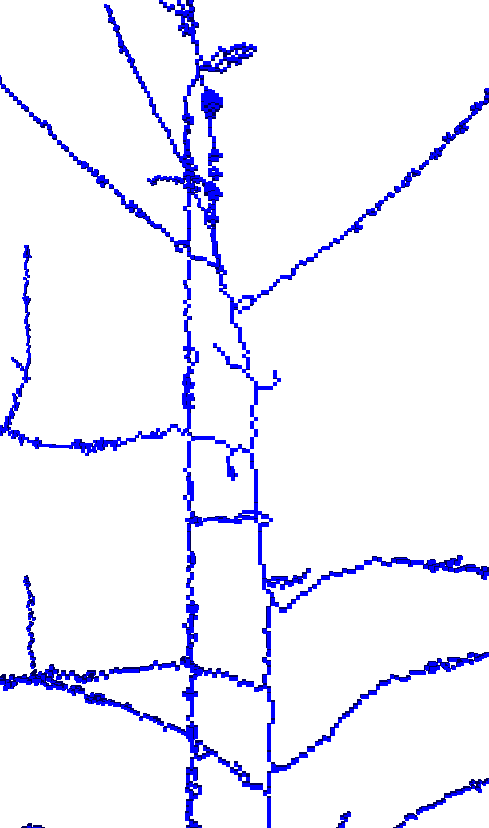}
}
\subfloat[Jin \etal]  
	{  
	\includegraphics[trim=2cm 5cm 2cm 5cm, clip=true, totalheight=\tree]{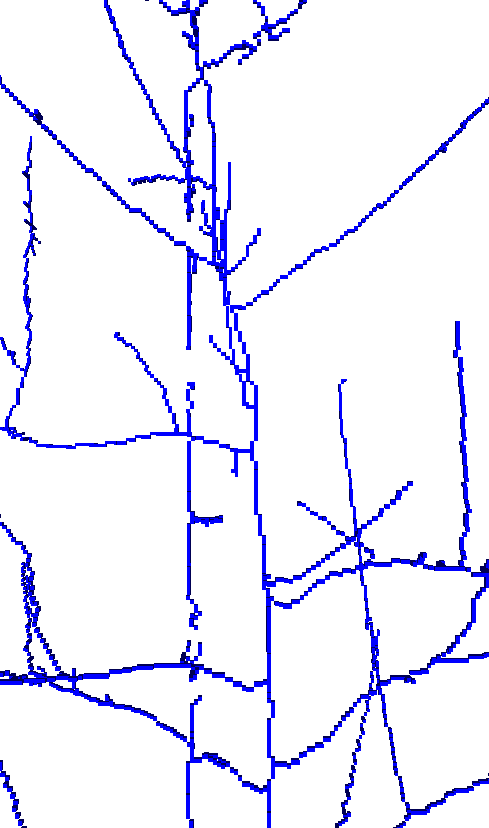}
}
\subfloat[Our method]  
	{  
	\includegraphics[trim=2cm 5cm 2cm 5cm, clip=true, totalheight=\tree]{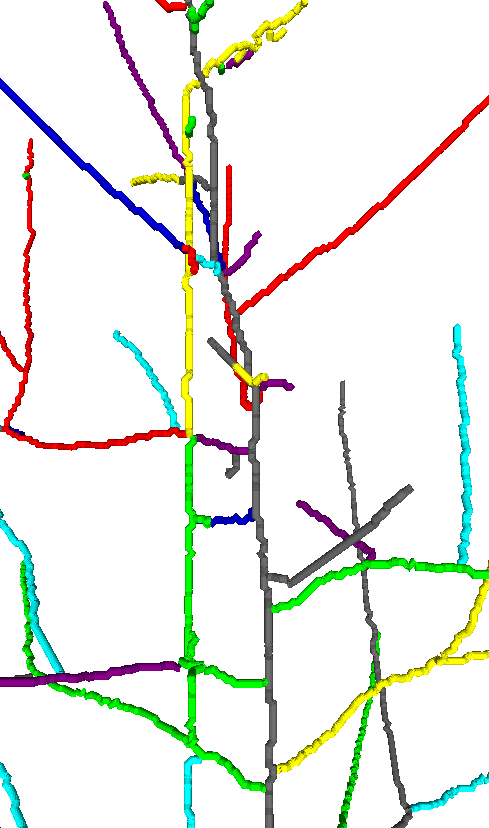}
\label{sf:ours_B}
}
\caption{\textbf{Best viewed in color.} Detailed view of the results from Dataset B: the surface reconstruction with noise of a real tree with a supporting metal pole, and curve skeletons computed with the thinning algorithm, PINK skel script, PINK filter3d script, Jin \etalspace method, and our proposed method.  The different colors in \ref{sf:ours_B} represent the curve skeleton segments identified during the course of the algorithm. Figures for Datasets A, the complete view of B, and C-I are given in the supplementary materials.}
\label{fig:resultsB_detail}
\end{figure}

We also assessed the performance of our method under increasingly noisy conditions on a synthetic 3D model, which serves as a ground truth. We iteratively add noise to the ground truth model by randomly choosing, with uniform probability, $(p/2)\times n$ surface voxels which have non-surface neighbors to be deleted and another set of the same size to which a new neighboring voxel is added. The parameter $p$ represents the proportion of voxels to be altered, and in our experiments $p=0.05$. For subsequent iterations, we repeat the process using the voxel occupancy map from the previous iteration such that, after $n_l$ iterations, the model has either noisy protrusions or depressions of at most $n_l$ voxels. A closeup view of the model without noise and with a noise level of $n_l=14$ as well as the corresponding curve skeletons computed using our method can be found in the supplementary materials.

\newcommand*{\factorSynthetics}{0.11}

\begin{figure}[!ht]
    \centering
	\includegraphics[width=0.85\linewidth]{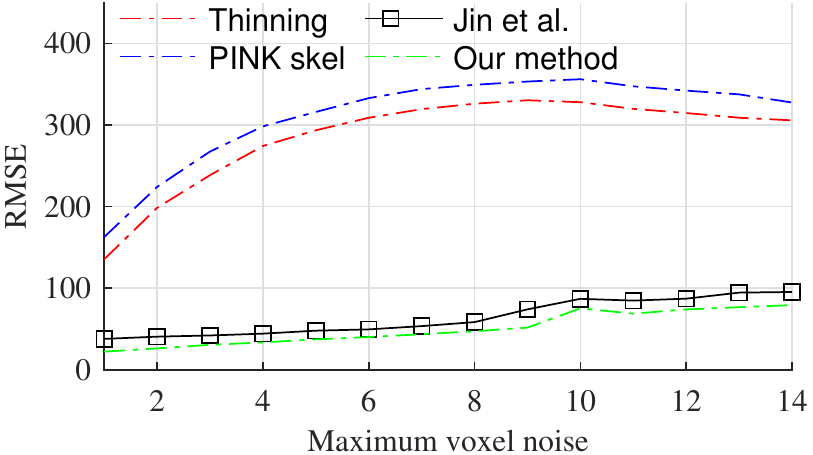}
\caption{\textbf{Best viewed in color.}  Root mean squared error of the curve skeletons computed with the comparison methods and our method, as compared to the ground truth curve skeleton. The PINK filter3d method is not reported, since its minimum error value is $3356$.} 
\label{fig:summed_error}
%\end{figure*}
\end{figure}

To quantify the effect of noise on each of the skeletonization methods, we compute the root mean squared error (RMSE) of the skeletons generated by each approach in comparison with the ground truth skeleton. That is, for each voxel in the curve skeleton of the ground truth model, we find the closest voxel in the curve skeleton of the noisy model and use the sum of the squared closest distances to compute the RMSE. Figure \ref{fig:summed_error} shows the corresponding results for our method and the comparison approaches.  The x-axis in the graph represents the maximum voxel noise according to the process described in the preceding paragraph, and the y-axis shows the RMSE in terms of voxel distances. As the figure indicates, our method outperforms all the other approaches by a significant margin. The second best approach is given by the method of Jin \etal, which has an average RMSE 20\% higher than our method.

\subsection{Computational efficiency} \label{r2}

\begin{figure}[ht]
\centering
	\includegraphics[width=.85\linewidth]{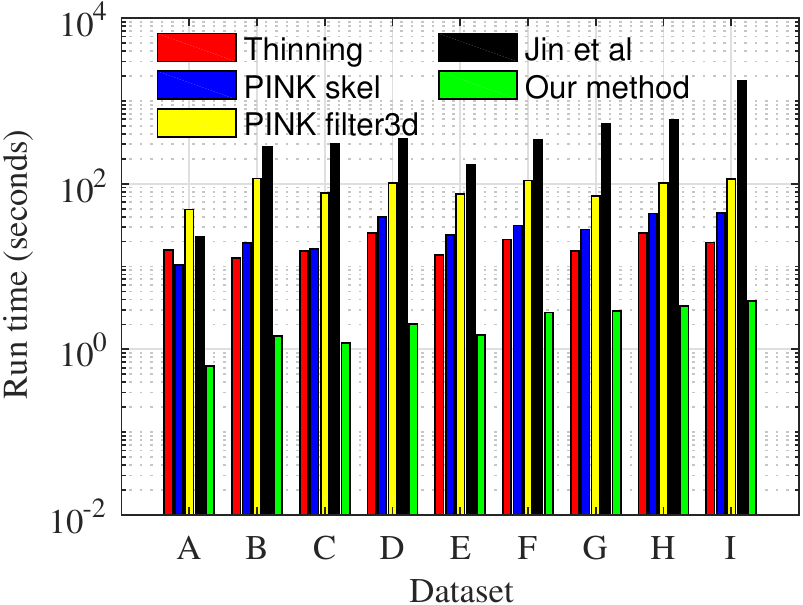}
\caption{\textbf{Best viewed in color.} Run times of our curve skeletonization method on the nine datasets in comparison with the existing approaches. The vertical axis is shown on a logarithmic scale due to the dramatic differences between our method and the existing approaches.   }\label{fig:runtimes}
\end{figure}

Figure \ref{fig:runtimes} summarizes the time performance of the methods on the nine trees. A general ordering with respect to increasing run time is: 1) our method, 2) thinning, 3) PINK skel, 4) PINK filter3d, and 4) the Jin \etalspace method.  The method proposed by Jin \etal , which has as one of its components geodesic path computation, has the longest run time of the methods.  Note that the run times of the comparison methods include loading and saving the results, whereas ours does not. The loading and saving portions required to run the thinning algorithm for dataset E, which has $N=58$ million voxels, is $1.2$ seconds.  Since we do not have access to the source code of Jin \etal, assessing the time spent loading and saving is difficult, but we assume that it is of the same order of magnitude. For the PINK scripts, intermediate results are loaded and saved in temporary locations, which affects run time. Nevertheless, our curve skeletonization method is able to compute curve skeletons one to three orders of magnitude faster than the other methods. It executes in less than four seconds even for very large models.  

\subsection{Results on traditional skeletonization datasets and influence of $t$ on the results} \label{r3}

While we are interested in real-world, elongated, noisy objects, we also evaluated our method on some commonly-used smooth models. The proposed curve skeleton method used in the context of smooth models produced the general structure of those objects and was also able to detect loops when they were present, for instance for the camel and seahorse examples. We also performed experiments with the only user-supplied threshold in our method.  When $t = 1$, no segments are discarded. The resulting curve skeleton resembles a more dense version of the thinning result. When $t = 0.0001$, the result resembles the Jin \etalspace result except that loops are preserved. Results for both of these items are found in the supplementary materials.

\newcommand*{\startrectx}{0.3}
\newcommand*{\startrecty}{0.3}
\newcommand*{\myendrectx}{1.2}
\newcommand*{\myendrecty}{1.5}

\section{Conclusions}

Understanding the structure of complex elongated branching objects in the presence of noise is a challenging problem with important real-world applications. In this paper, we presented a fast and robust algorithm to compute curve skeletons of such real-world objects. These curve skeletons provide most of the information necessary to represent the structure of these objects.  A large portion of the paper centered on how the ideas of BFS could be exploited to create an efficient curve skeletonization procedure. Our approach is able to detect connected segments and performs pruning in the course of the algorithm, so those steps do not need to be performed separately. The small run times of less than a few seconds make this method suitable for automation tasks where real-time decisions are required.

{\small
\bibliographystyle{ieee}
\bibliography{curveskeleton}
}

\clearpage
%\widetext

\begin{center}
\textbf{\large Fast and robust curve skeletonization for real-world elongated objects: Supplementary materials}
\end{center}

\setcounter{equation}{0}
\setcounter{figure}{0}
\setcounter{table}{0}
\makeatletter
\renewcommand{\theequation}{S\arabic{equation}}
\renewcommand{\thefigure}{S\arabic{figure}}

\section{Computational complexity}
\label{ss:complexity}

In this section, we analyze the computational complexity of the algorithms in this paper. For reference ease, we include all of the algorithms from the main paper.

\begin{algorithm}
\caption{Modified BFS Algorithm} \label{alg:l1norm_bfs}
\begin{algorithmic}[1]
\While {$|\mathbb{F}_A^{(k)}| > 0$} \label{alg1:line_foreachF2}
\For {each voxel $v_i \in \mathbb{F}_A^{(k)}$} \label{alg1:line_foreach1}
\For {each voxel $v_j \in \mathcal{N}_i$ such that $(l_j > l_i)$} \label{alg1:line_foreach2} 
\State {$l_j = \min(l_j, l_i + w_j + ||v_j - v_i||)$} \label{alg1:set_lj}
\EndFor
\EndFor
\State {$\mathbb{F}^{(k)} = \mathbb{F}_A^{(k)} \cup \mathbb{F}_B^{(k)}$}
\State \parbox[t]{\dimexpr\linewidth-\algorithmicindent}{{$\mathbb{N}^{(k)} = \lbrace v_j | l_j > l_i, \forall v_j \in \mathcal{N}_i, \forall v_j \notin \mathbb{F}^{(k)}, \forall v_i \in \mathbb{F}^{(k)}\rbrace$ \strut}} \label{alg1:set_N}
\State {$l_{min} = \min_{v_j \in \mathbb{N}^{(k)}} l_j$} \label{alg1:set_lmin}
\State {$\mathbb{F}_A^{(k + 1)} = \lbrace v_j | l_i < l_{min}, \forall v_i \in \mathbb{F}^{(k)}, \forall v_j \in \mathbb{N}^{(k)} \rbrace$ \strut} \label{alg1:set_f2prime}

\State \parbox[t]{\dimexpr\linewidth-\algorithmicindent}{{$\mathbb{F}_B^{(k + 1)} = \lbrace v_i | l_i \geq l_{min}, \forall v_i \in \mathbb{F}^{(k)}, |\mathcal{N}_i \cap \mathbb{N}^{(k)}| > 0\rbrace$ \strut}} \label{alg1:set_f1prime}
\State {$k = k + 1$}
\EndWhile
\end{algorithmic}
\end{algorithm}

\begin{algorithm}[t]
\caption{Determination of curve skeleton segment from BFS2 and $d_i$} \label{alg:pseudo_code_path_trace}
\begin{algorithmic}[1]
\State {$v_c = v_{s,1}$}
\State {$\mathbb{C}^{(m)} = \lbrace v_{s,1} \rbrace$}
\While {$(v_c \neq v_t) \wedge (v_c \notin \mathbb{C})$}
\State \parbox[t]{\dimexpr\linewidth-\algorithmicindent}{Determine  $d^* = \underset{v_{i}\in\mathcal{N}_c \wedge l_c > l_i}{\max}\left(d_i\right)$ where $d_i$ is the distance transform of $v_i$ \strut}
\State \parbox[t]{\dimexpr\linewidth-\algorithmicindent}{Compute $v_n = \underset{v_{j}\in\mathcal{N}_c^*}{\argmin}\left(l_j\right)$ where $l_j$ is the BFS2 label of $v_j$ and ${N}_c^* = \left\{v_j|\exists v_j \in \mathcal{N}_c, d_j=d^* \right\} $ \strut}
\State {$\mathbb{C}^{(m)} = \mathbb{C}^{(m)} \cup \lbrace  v_n \rbrace$}
\State {$v_c = v_n$}
\EndWhile
\end{algorithmic}
\end{algorithm}

\begin{algorithm}
\caption{Loop handling} \label{alg:loop_handling}
\begin{algorithmic}[1]
\State {$\mathbb{C}_t = \cap_{j \in [0, a-1]} \mathbb{C}^{\prime}_j$}\label{line:compute_common}
\For {$j = 0$ to ${a-1}$}
\State {$\mathbb{C}^{\prime}_j =\mathbb{C}^{\prime}_j \setminus \mathbb{C}_t$}\label{line:remove_common} 
\For {$k = j + 1$ to ${a-1}$}
\State {$\mathbb{C}^{\prime}_j =\mathbb{C}^{\prime}_j \setminus \mathbb{C}_k$} \label{line:remove_redundacies} 
\EndFor
\State {$\mathbb{C} = \mathbb{C} \cup \left\{ \mathbb{C}^{\prime}_j \right\} $}\label{line:add_to_set} 
\EndFor 
\end{algorithmic}
\end{algorithm}

\begin{algorithm}[t]
\caption{Proposed skeletonization algorithm} \label{alg:pseudo_code_step_add_curve_segments}
\begin{algorithmic}[1]
\State Determine seed voxel $v^*$ and make initial curve skeleton $\mathbb{C}^{(0)}=v^*$
\Repeat 
\State \parbox[t]{\dimexpr\linewidth-\algorithmicindent}{Update BFS1 labels using Alg. \ref{alg:l1norm_bfs} with $\mathbb{F}_A^{(0)} = \mathbb{C}^{(m-1)}$\strut}
\State {Locate a proposed endpoint $v_t$ from BFS1}
\State \parbox[t]{\dimexpr\linewidth-\algorithmicindent}{Create BFS2 labels using  Alg. \ref{alg:l1norm_bfs}  with $\mathbb{F}_A^{(0)} = \lbrace v_t \rbrace$\strut}
\State \parbox[t]{\dimexpr\linewidth-\algorithmicindent}{Create $\mathbb{C}^{(m)}$ by tracing paths in BFS2 labels from existing curve skeleton $v_t$ according to Alg. \ref{alg:pseudo_code_path_trace}\strut}
\If {(Number of FCCs == 1)}
\State \parbox[t]{\dimexpr\linewidth-\algorithmicindent}{Accept or decline curve skeleton segments using classification method.  If accepted $\mathbb{C} = \mathbb{C} \cup \left\{ \mathbb{C}^{(m)} \right\} $ and $m = m  + 1$\strut}
\Else {}
\State {Check for loops using Alg. \ref{alg:loop_handling}.}
\EndIf
\Until{No more endpoint hypotheses are found}
\end{algorithmic}
\end{algorithm}

\subsection{Complexity of the modified BFS algorithm}

We now analyze the computational complexity of Algorithm \ref{alg:l1norm_bfs}. Note that a voxel is a member of $\mathbb{F}_A^{(k)}$ for any $k$ exactly once. On line \ref{alg1:set_f2prime}, voxels in $\mathbb{F}_A^{(k + 1)} \subseteq \mathbb{N}^{(k)}$ and a voxel in $\mathbb{N}^{(k)}$ cannot be in $\mathbb{F}^{(k)}$: $\mathbb{F}^{(k)} \cap \mathbb{N}^{(k)} = \emptyset$ (from line \ref{alg1:set_N}). In addition, the condition $l_j > l_i$ ensures that a voxel in $\mathbb{F}_A^{(k + 1)}$ has not already been discovered, meaning that voxels in $\mathbb{F}_A^{(k + 1)}$ cannot be members of the frontier $\mathbb{F}^{(m)}$ at any previous iteration $m$, $m = 0, 1, ... , k-1$.  Hence $\mathbb{F}_A^{(k)} \cap \mathbb{F}_A^{(j)} = \emptyset$ for $k \neq j$. Finally, since all voxels are eventually discovered, the sets $\mathbb{F}_A^{(k)}$ form a partition of $\mathbb{V}$, implying 
\begin{equation}
\bigcup_k \mathbb{F}_A^{(k)} = \mathbb{V}
\end{equation}
and
\begin{equation}
\sum_k |\mathbb{F}_A^{(k)}| = |\mathbb{V}| = n
\label{equation_f2}
\end{equation}
Consequently, for all iterations of Algorithm \ref{alg:l1norm_bfs}, line \ref{alg1:line_foreach1} will be performed $n$ times.  

On the other hand, a voxel $v_i$ may be a member of $\mathbb{F}_B^{(k)}$ multiple times, waiting for the condition $l_i < l_{min}$ to be true so that its neighbors are moved into $\mathbb{F}_A^{(k)}$ (line \ref{alg1:set_f2prime}). Because the conditions for computing set $\mathbb{N}^{(k)}$, and therefore $l_{min}$ (lines \ref{alg1:set_N} - \ref{alg1:set_lmin}), for the members of $\mathbb{F}_B^{(k)}$ are dependent on the composition of $\mathbb{F}_A^{(k)}$, $l_{min}$ cannot be stored from previous iterations and still be relevant.

Finding set $\mathbb{N}^{(k)}$ in Algorithm \ref{alg:l1norm_bfs} takes time $|\mathbb{F}_A^{(k)}| + |\mathbb{F}_B^{(k)}|$ per iteration $k$, and line \ref{alg1:set_lmin} can be computed while $\mathbb{N}$ is found.  Lines \ref{alg1:set_f2prime} and \ref{alg1:set_f1prime} also take time $|\mathbb{F}_A^{(k)}| + |\mathbb{F}_B^{(k)}|$ per iteration $k$.

Since we know from Equation \ref{equation_f2} that $\sum_k|\mathbb{F}_A^{(k)}| = n$, the asymptotic lower bound is $\Omega(n)$, and it is a strict bound.  For instance, consider $\mathbb{F}_A^{(0)}$ contains one voxel, which is an endpoint of a 1-dimensional line in 3D space.  In this case, $|\mathbb{F}_A^{(k)}| = 1$ and $|\mathbb{F}_B^{(k)}| = 0$ for all $k$, and the maximum value of $k$ is $n$.

Determining the value of $\sum_k |\mathbb{F}_B^{(k)}|$ in general is problematic as it depends strongly on the shape involved, on the composition of $\mathbb{F}_A^{(0)}$, and weights $w_i$.

\subsection{Complexity of modified BFS algorithm with zero weights}
\label{ss:zero}
The Euclidean distance between neighboring voxels is $d \in \lbrace 1, \sqrt{2}, \sqrt{3} \rbrace$.  We claim that a voxel can only be in one of the frontier sets at most three times; first, it enters the frontier through $\mathbb{F}_A^{(k)}$, and if the voxel remains in the frontier, the only other sets it may belong to are $\mathbb{F}_B^{(k + 1)}$ and $\mathbb{F}_B^{(k + 2)}$, at which point it exits the frontier sets. Hence 
\begin{equation}
\sum_k (|\mathbb{F}_A^{(k)}| + |\mathbb{F}_B^{(k)}|) \leq 3n
\end{equation}
The asymptotic upper bound of Algorithm \ref{alg:l1norm_bfs} is then $\mathcal{O}(n)$.  

Algorithm \ref{alg:l1norm_bfs} using zero weights has many of the same characteristics as the breadth-first distance transform described in Section V of \cite{Silvela2001Breadth}.  One major difference is that in \cite{Silvela2001Breadth} neighbors are assumed to have the same distance from each other, which is not an assumption we share when working with 26-connected voxels in 3D. However, the algorithm with zero weights retains the asymptotic upper bound of $\mathcal{O}(n)$ as the method in \cite{Silvela2001Breadth}.

\subsection{Complexity of modified BFS with non-zero weights}
\label{ss:nonzero}
The complexity of the modified BFS algorithm with non-zero weights is $\mathcal{O}(n d_{max})$, assuming that the weights are computed based on a distance transform $w_i = d_{max} - d_i$. 

Our discussion of Algorithm \ref{alg:l1norm_bfs} for zero weights provided an asymptotic upper bound $\mathcal{O}(n)$, and determining this bound came down to determining how many times a voxel could possibly be a member of the frontier.  For the analysis of Algorithm \ref{alg:l1norm_bfs} when weights are non-zero, we return to similar questions. From Algorithm \ref{alg:l1norm_bfs}, voxels are in $\mathbb{F}_A$ once, and are in $\mathbb{F}_B$ multiple times, until the condition $l_i < l_{min}^{(k)}$ is satisfied.

Let us determine the maximum number of times a voxel resides in the frontier sets.  The maximum distance label $d_{max}$ is the maximum difference between any two distance labels, and the L2 norms between any two neighboring voxels belong to the set $\lbrace 1, \sqrt{2}, \sqrt{3} \rbrace$.  Then, the maximum number of times a voxel can possibly be in the frontier sets is $3d_{max}$, leading to asymptotic upper bound $\mathcal{O}(n d_{max})$. We consider shapes that produce the largest values of $\frac{d_{max}}{n}$.  The shape that maximizes $\frac{d_{max}}{n}$ is a sphere.  In that scenario, $d_{max}$ is the radius of the sphere, and the relationship between $n$ and $d_{max}$ is $n = \frac{4}{3} \pi \left(d_{max}\right)^3$.  Then, performing the relevant substitutions we have the asymptotic upper bound of this step as  $\mathcal{O}(n^{\frac{4}{3}})$.

\subsection{Computational complexity of the curve skeleton method}

\begin{figure}
\centering
	\includegraphics[width=0.93\linewidth]{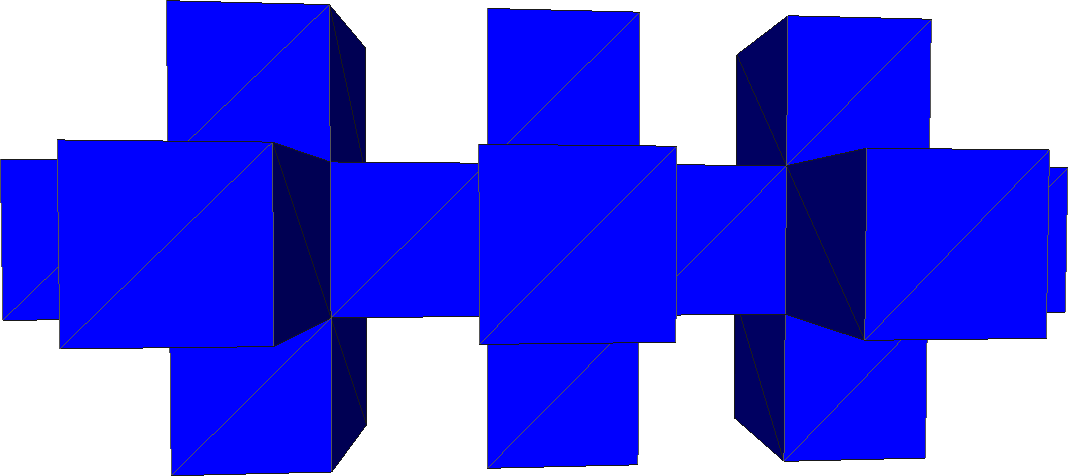}
\caption{Example of a shape with a high proportion of proposed endpoints $t$ relative to $n$.   }\label{fig:complexity_example}
\end{figure}

We now consider the complexity of the curve skeleton method of Algorithm \ref{alg:pseudo_code_step_add_curve_segments}, excluding the complexity of the distance transform step, which is $\mathcal{O}(n)$ using the approach of ~\cite{meijster2002general}. Let the number of proposed endpoints, or iterations of Algorithm \ref{alg:pseudo_code_step_add_curve_segments}, be $||\mathbb{V}_t||$.  Computing the BFS1 labels (step 2.1) and BFS2 labels (step 3.1) per iteration has asymptotic upper bound $\mathcal{O}(d_{max}n)$.  The computation of the BFS1 and BFS2 labels is repeated $||\mathbb{V}_t||$ times, giving $\mathcal{O}(||\mathbb{V}_t|| d_{max} n)$ as an upper bound.  As in Section \ref{ss:nonzero}, we try to determine shapes that maximize $\frac{||\mathbb{V}_t||}{n}$ to find an asymptotic upper bound on $||\mathbb{V}_t||$. We have given an example of a particular shape in Figure \ref{fig:complexity_example} where $||\mathbb{V}_t|| = \frac{2}{3}(n + 2)$. More generally, the number of extremities can be no larger than the number of voxels, hence we can assume that $||\mathbb{V}_t|| = \mathcal{O}(n)$.  In summary, the asymptotic upper bound of the complete algorithm is $\mathcal{O}(n^{\frac{7}{3}})$.

\begin{figure}
\centering
	\includegraphics[width=0.90\linewidth]{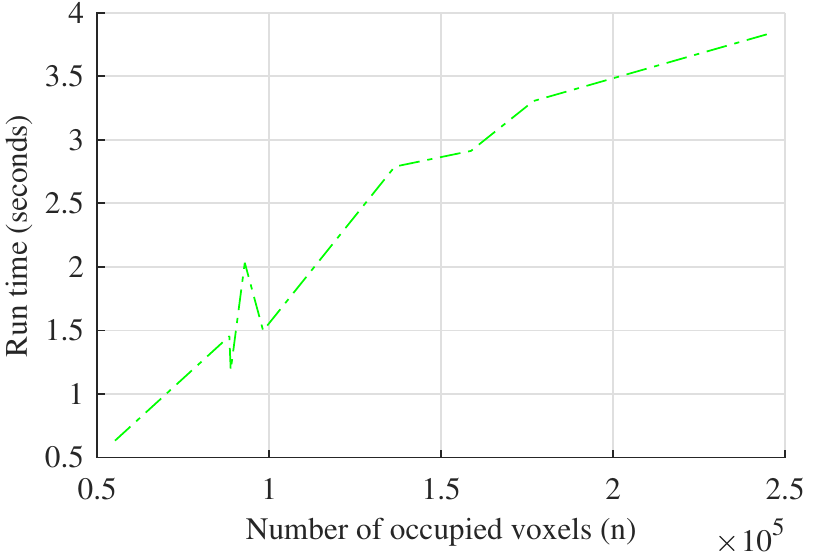}
\caption{Execution time of our algorithm as a function of the number of occupied pixels in each dataset.   }\label{fig:nversust}
\end{figure}

We note that in our experiments involving elongated objects, both $||\mathbb{V}_t||$ and $d_{max}$ are extremely small relative to $n$ as shown in Table \ref{table:datasets}.  In those datasets, the maximum value of $||\mathbb{V}_t||$ relative to $n$ is $||\mathbb{V}_t|| \leq 0.002 n$ (Dataset B), and the greatest value of $d_{max}$ is $16$, or in terms of $n$, $d_{max} \leq 0.0003 n$ (Dataset A). Consequently, while the asymptotic upper bound is greater than quadratic, if  $||\mathbb{V}_t||$ and $d_{max}$ can be assumed to be small constants as in the case of elongated objects, in practice the algorithm runs quickly. This is highlighted in Figure \ref{fig:nversust}, which shows the runtime of our method as a function of the number of occupied voxels $n$ for each of the datasets in Table \ref{table:datasets}. As the figure indicates, despite a four times increase in the number of occupied voxels between the smallest and the largest dataset, the run time of our algorithm increases slowly.

\begin{table}[h]
\caption{Characteristics of the nine datasets used in our evaluation. $n$ is the number of occupied voxels/nodes, $N$ is the number of voxels in the grid, $d_{max}$ is the largest voxel to surface distance, and $||\mathbb{V}_t||$ is the number of proposed endpoints $v_t$.}
\begin{center}
%\resizebox{\linewidth}{!}
{
\begin{tabular}{lcccr}
\hline 
\multicolumn{5}{c}{} \\
ID & $n$ & $N$ & $d_{max}$ & $||\mathbb{V}_t||$ \\
\hline \hline
A & \phantom{0}55,156 &  27,744,000 & 16 & 15  \\
\hline
B & \phantom{0}88,407 & 56,832,000 & 5 & 174  \\
\hline
C & \phantom{0}88,798 & 45,240,000 & 7 & 145  \\
\hline
D &  \phantom{0}92,892  & 80,640,000 & 6 & 154 \\
\hline 
E & \phantom{0}98,228  & 58,464,000 & 9 & 50 \\
\hline
F & 136,497 & 80,640,000 & 7 & 134   \\
\hline 
G & 158,686  & 64,512,000 & 9 & 128 \\
\hline
H & 176,820  & 80,640,000 & 10 & 97 \\
\hline 
I & 246,654  & 80,640,000 & 10 & 83 \\
\hline 
\end{tabular}
}
\end{center}
\label{table:datasets}
\end{table} 

\section{Topological Stability}

There are three steps in the algorithm where decisions may be made deterministically or randomly in the case of equal labels:(1) the selection of $v^*$ in Section 3.2, (2) endpoint candidates $v_t$ in 3.3.2, and (3) frontier connecting voxel $v_{s,1}$ in 3.4.2.  If topological stability is desired, the following protocol could be employed to deterministically select a voxel when there are multiple voxels with the same label.  The voxels of equal label are placed into a vector, and then the $x, y, z$ coordinates would be sorted as specified by the user.  One such ordering would be to sort the coordinates by $x$ value, and then in case of ties by $y$ value, and then in case of ties by $z$.  This kind of repeatable ordering would produce reproducible results that would be more topologically stable than a random ordering.

\section{Additional figures for loop handling step (section 3.5.2 in the main paper)}

Figure \ref{fig:loop_handling} shows the loop handling procedure.  As shown in Figure \ref{sf:loopG}, only regions with loops are recovered.

\renewcommand*{\wLH}{0.8}
\begin{figure}[!ht]
\centering
\subfloat[]{  
	\includegraphics[trim  = 2cm 8.5cm 0cm 0cm, clip, width=\wLH\linewidth]{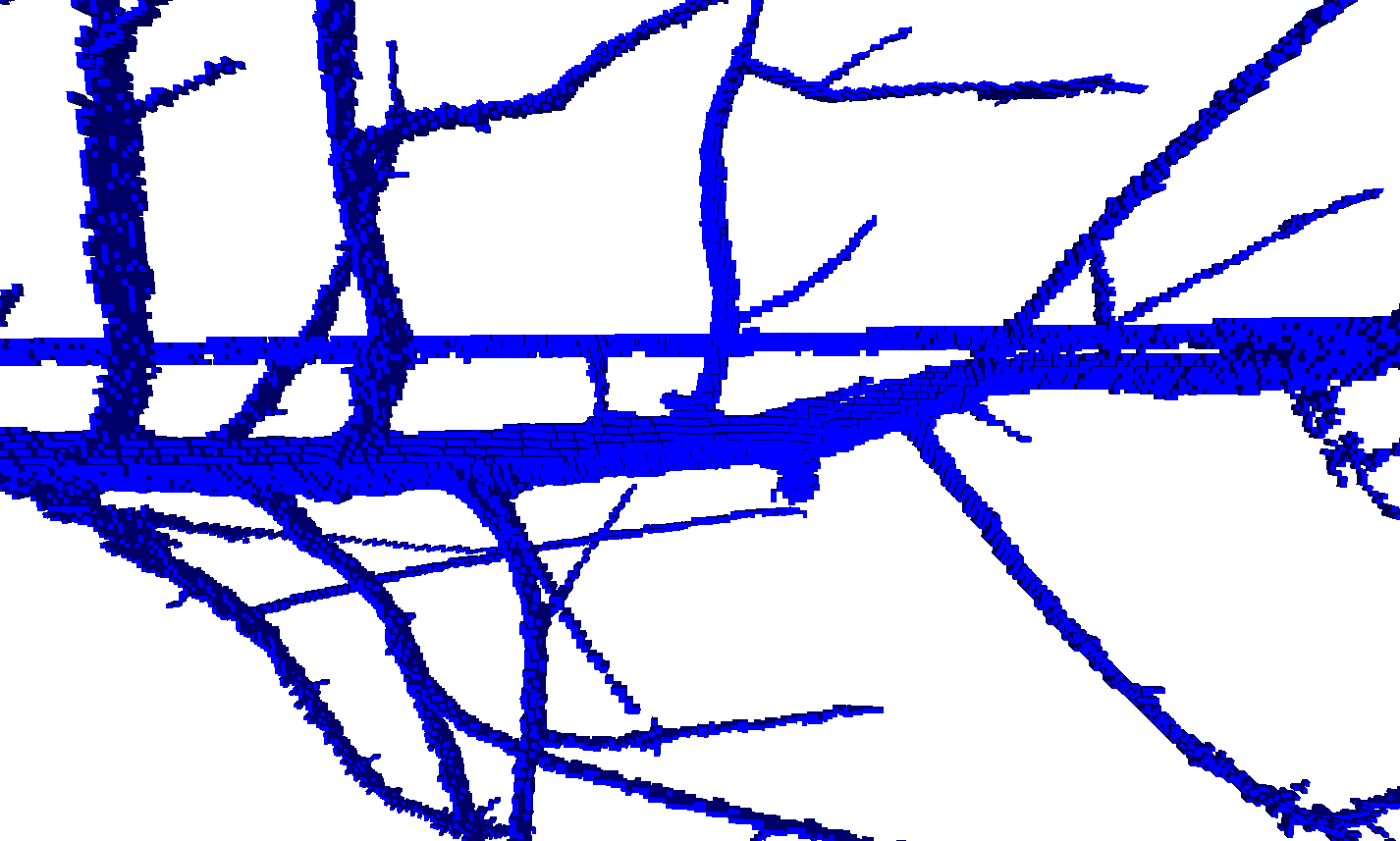}
 \label{sf:loopA} 
} \\
\subfloat[]{  
	\includegraphics[trim  = 2cm 4.5cm 0cm 0cm, clip,width=\wLH\linewidth]{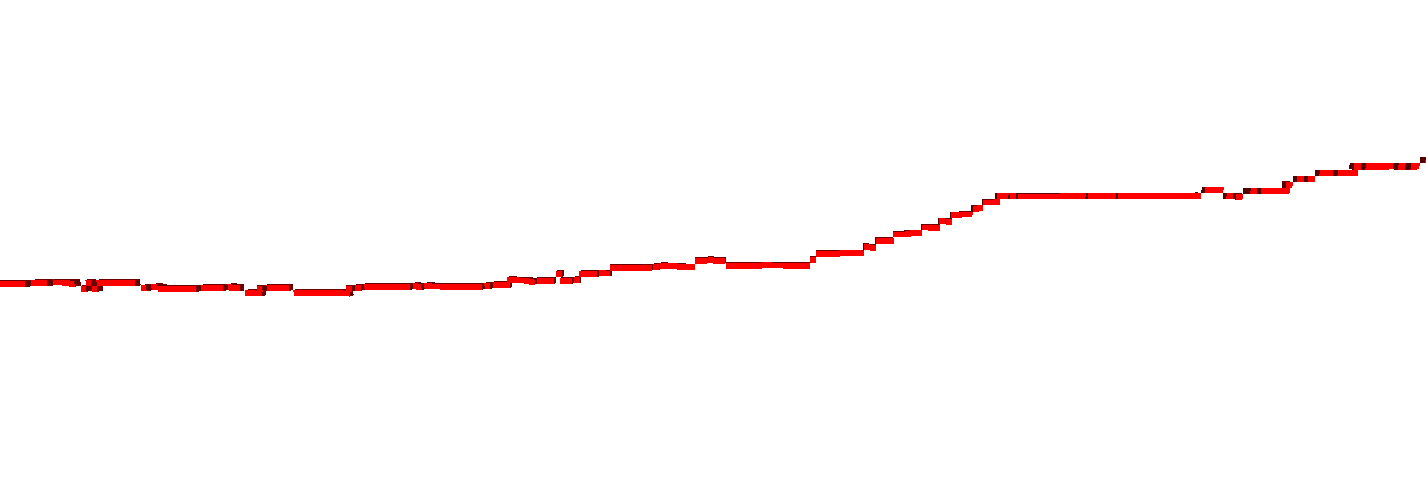}
\label{sf:loopB}
} \\
\subfloat[]{  
	\includegraphics[trim  = 2cm 4.5cm 0cm 0cm, clip,width=\wLH\linewidth]{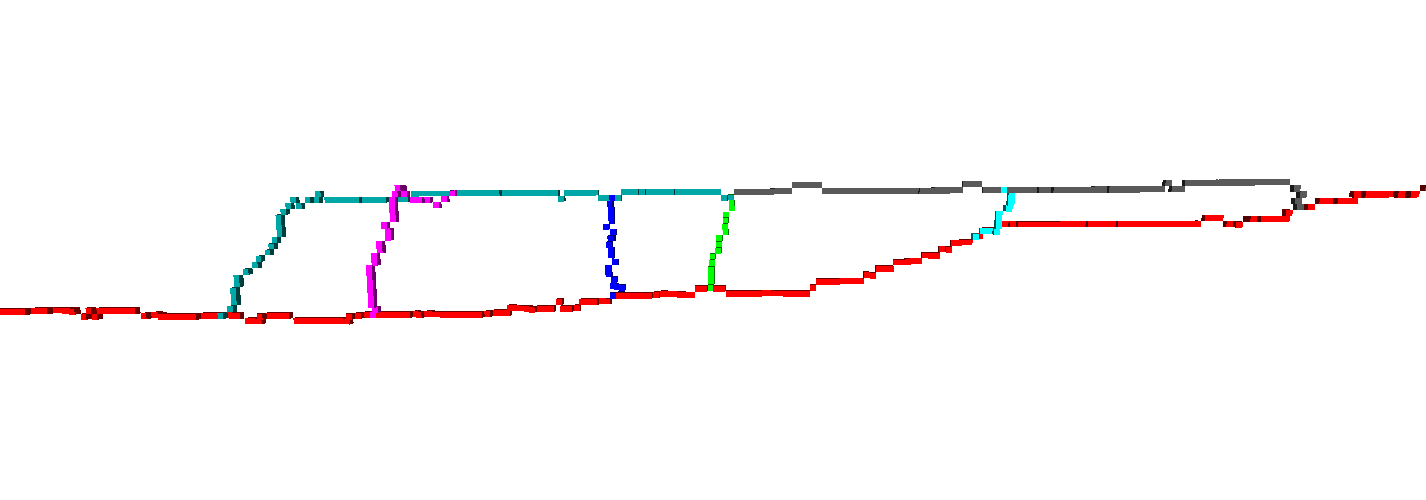}
\label{sf:loopG}
}
\caption{\textbf{Best viewed in color.} An example of the loop handling procedure. \ref{sf:loopA} shows the surface in blue, and \ref{sf:loopB} is the existing curve skeleton in red. Fig. \ref{sf:loopG} shows the result after Algorithm \ref{alg:loop_handling} is performed.}
\label{fig:loop_handling}
\end{figure}

\section{Additional Results}

\subsection{Simulated noise experiment}
\begin{figure*}[!ht]
    \centering
    %\includegraphics[width=1\textwidth]{figs1.eps}
    %\caption{caption1}
    %\label{f1}
\subfloat[Ground truth model]  
	{  
	\includegraphics[width=0.35\linewidth]{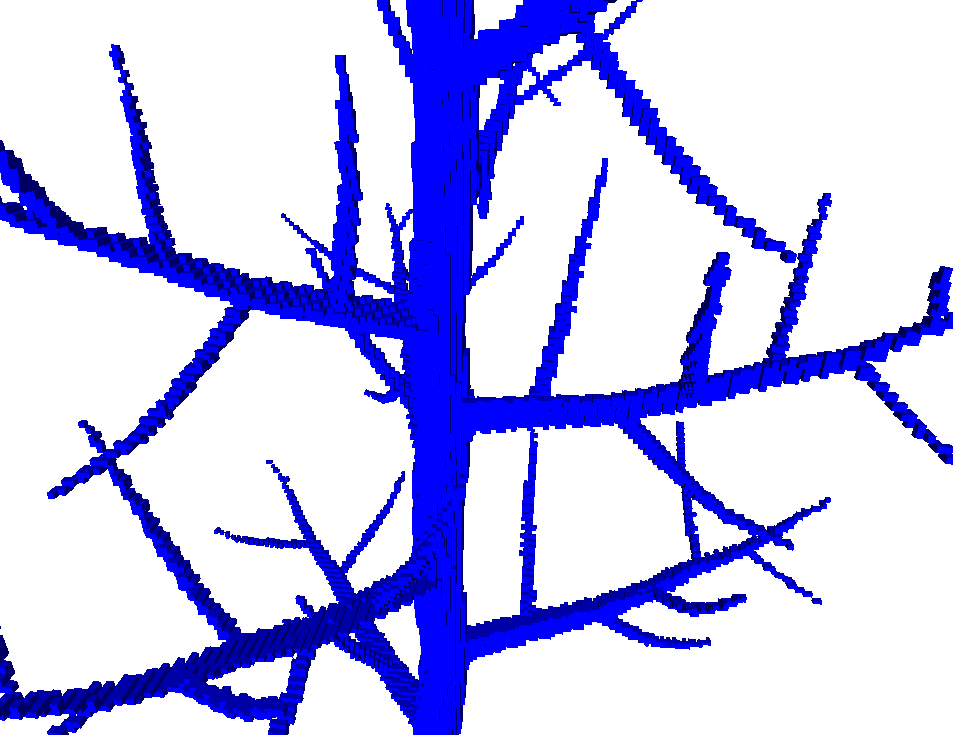}
\label{sf:no_noise_tree}
}% \hspace{0.05 cm}
\subfloat[Curve skeleton of \ref{sf:no_noise_tree}]  
	{  
	\includegraphics[trim=0cm 0cm 0cm 2cm, clip,width=0.35\linewidth]{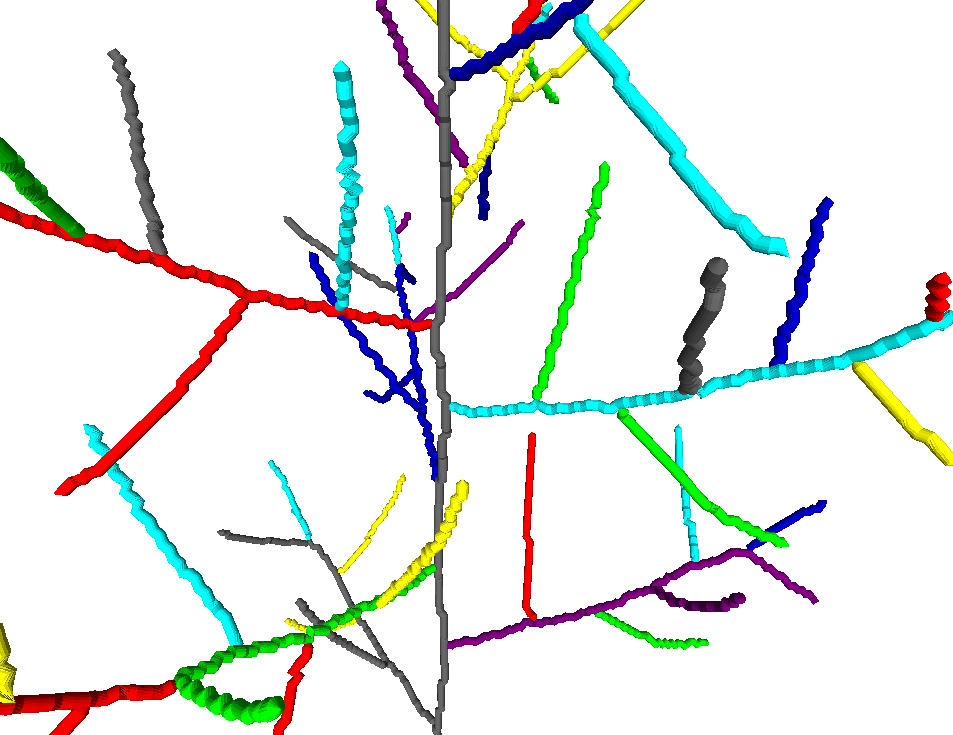}
	\label{sf:cs_gtl2}
} \\
\subfloat[Noise iteration 14]  
	{  
	\includegraphics[trim=0cm 0cm 0cm 2cm, clip,width=0.35\linewidth]{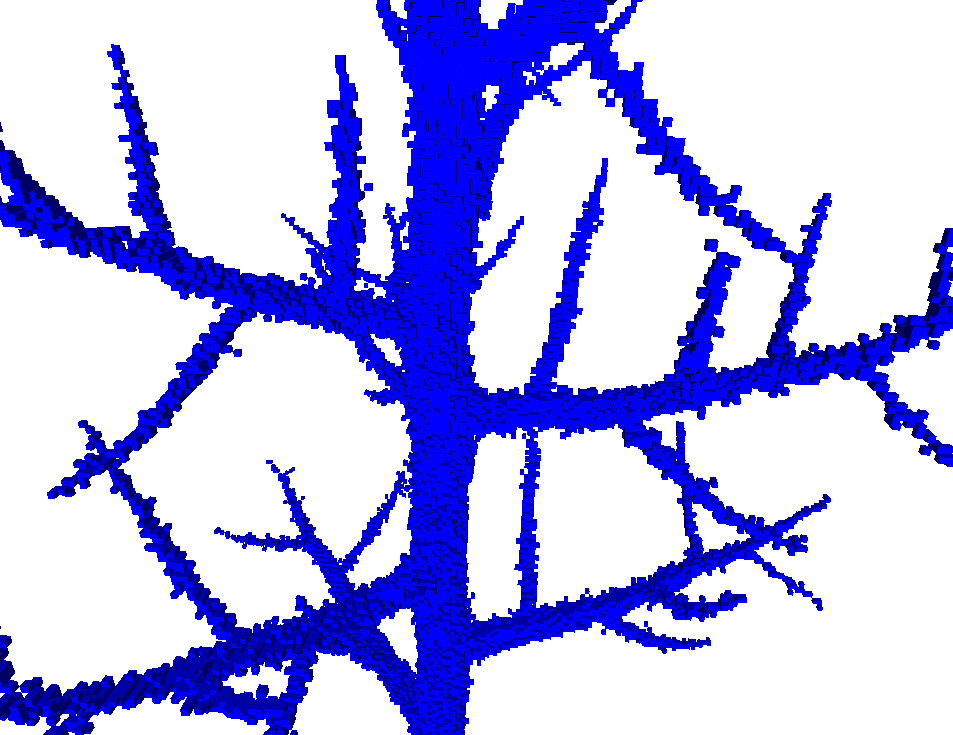}
\label{sf:noise_tree}
} 
\subfloat[Curve skeleton of \ref{sf:noise_tree}]  
	{  
	\includegraphics[trim=0cm 0cm 0cm 2cm, clip,width=0.35\linewidth]{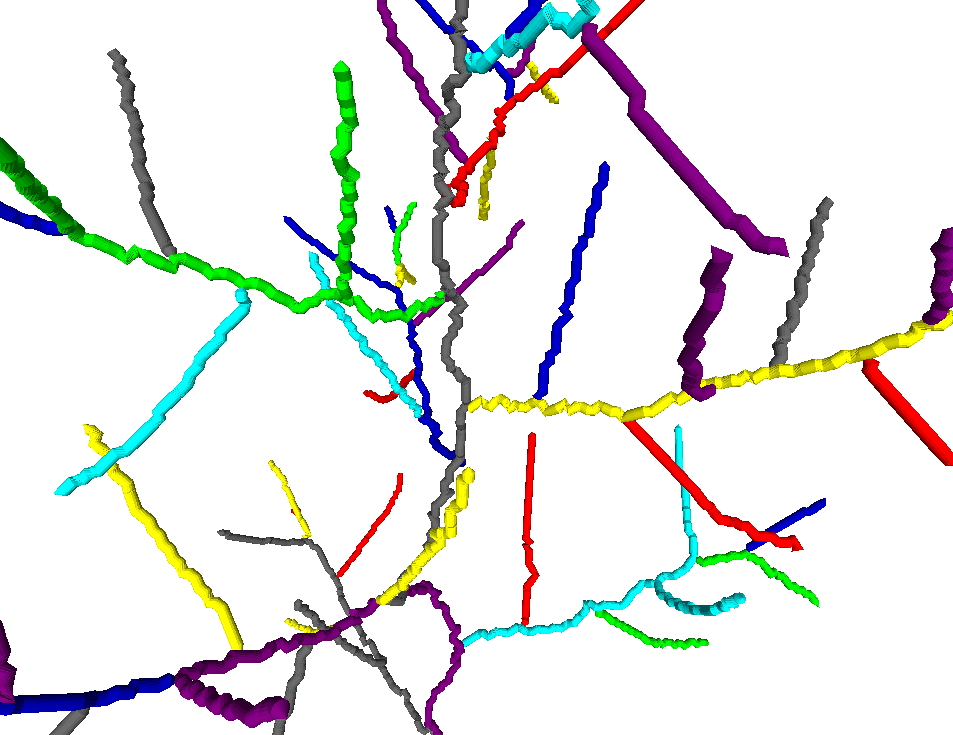}
\label{sf:cs_noisyl2}
}
\caption{\textbf{Best viewed in color.} Detail of a synthetic model of a tree without noise \ref{sf:no_noise_tree} and with noise \ref{sf:noise_tree} at iteration 14.  The curve skeleton of the ground truth is given in the top row, while curve skeleton of the noisy object is given in the second row.} 
\label{fig:synthetic}
\end{figure*}

Figure \ref{fig:synthetic} shows the models used for the simulated noise experiment, and computed curve skeletons using our proposed method.

\renewcommand*{\factorSynthetics}{0.11}
\begin{figure*}[]
\centering
\subfloat[Original surface]  
	{  
	\includegraphics[trim = 1cm 1cm 1cm 1cm,clip, width=\factorSynthetics\linewidth]{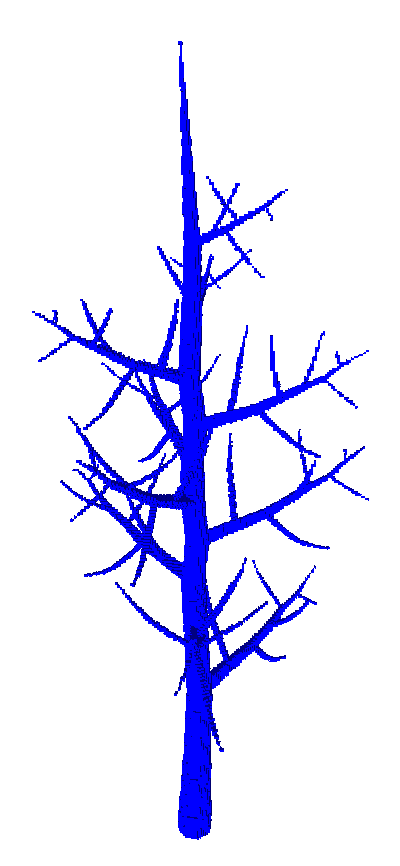}
} $\quad$
\subfloat[Ground truth curve skeleton]  
	{  
	\includegraphics[trim = 1cm 1cm 1cm 1cm,clip, width=\factorSynthetics\linewidth]{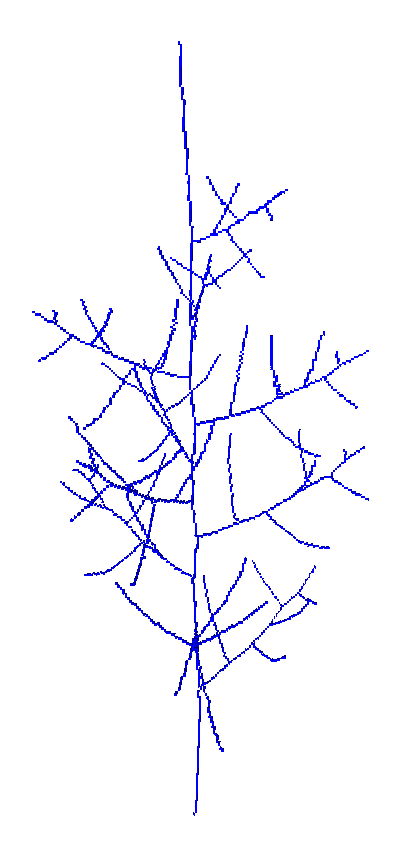}
\label{sf:gt_cs}
} $\quad$
\subfloat[Thinning]  
	{  
	\includegraphics[trim = 1cm 1cm 1cm 1cm,clip, width=\factorSynthetics\linewidth]{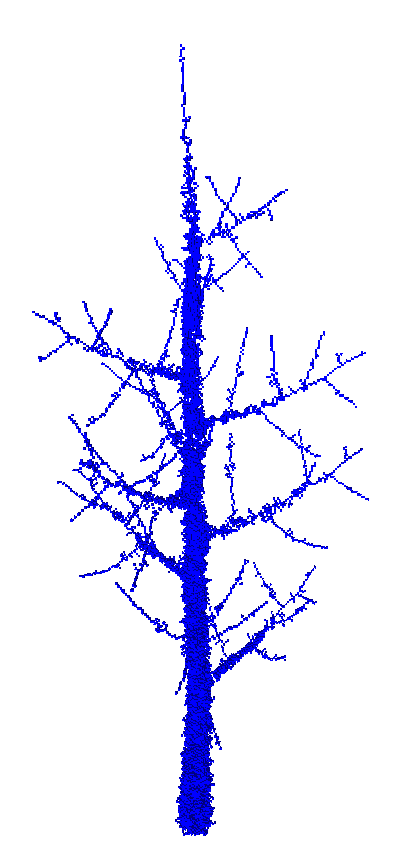}
} $\quad$
\subfloat[PINK skel]  
	{  
	\includegraphics[trim = 1cm 1cm 1cm 1cm,clip, width=\factorSynthetics\linewidth]{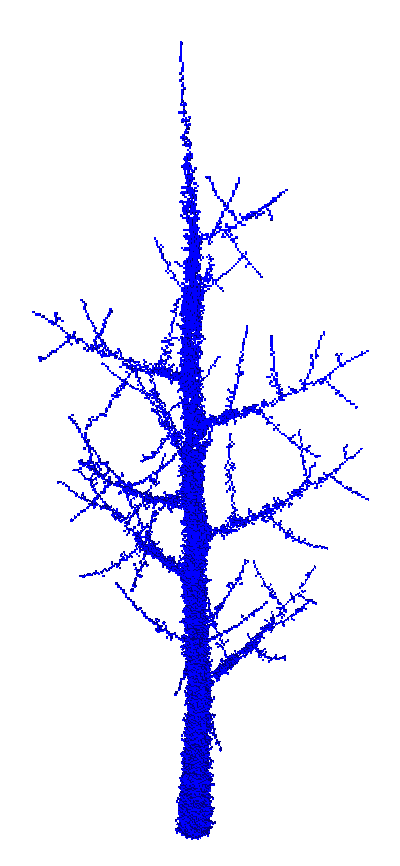}
} $\quad$
\subfloat[PINK filter3d]  
	{  
	\includegraphics[trim = 1cm 1cm 1cm 1cm,clip, width=\factorSynthetics\linewidth]{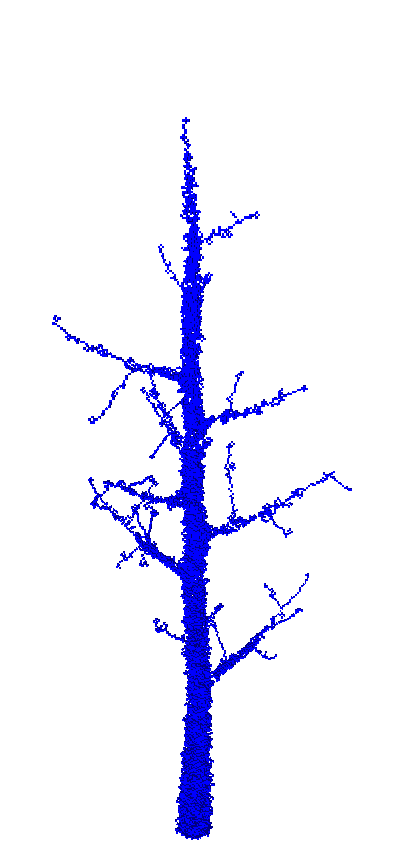}
} $\quad$
\subfloat[Jin \textit{et al.}]  
	{  
	\includegraphics[trim = 1cm 1cm 1cm 1cm,clip, width=\factorSynthetics\linewidth]{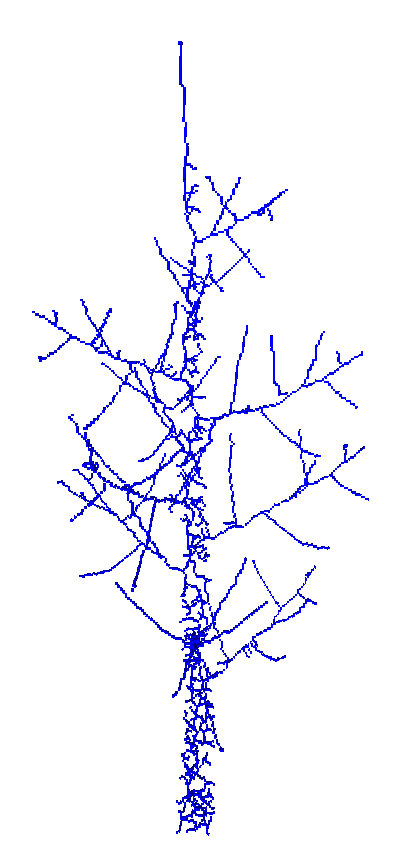}
} $\quad$
\subfloat[Our method]  
	{  
	\includegraphics[trim = 1cm 1cm 1cm 1cm,clip, width=\factorSynthetics\linewidth]{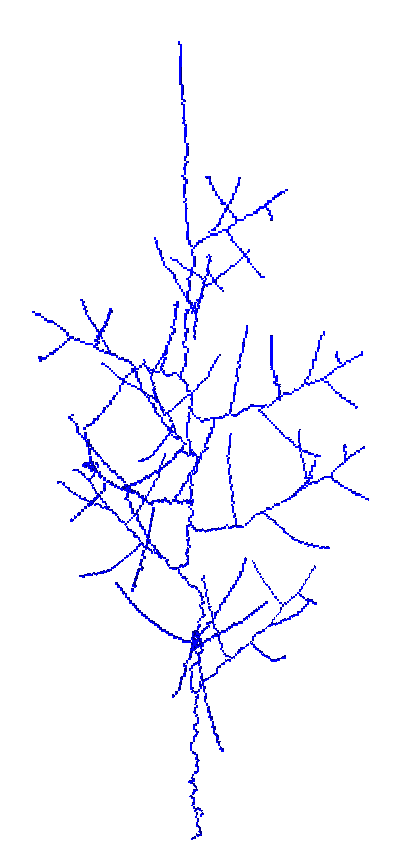}
\label{sf:gt_syn_ours}
} 
\caption{Synthetic model used in the evaluation of the robustness to noise and the corresponding outputs of each algorithm for a noise level of 14. Note that in the presence of noise the skeletons generated by the thinning algorithm as well as the two PINK methods contain a substantially higher number of voxels and no longer consist of thin one-dimensional segments.}
\label{fig:synthetic_results}
\end{figure*}

\begin{figure}[!ht]
    \centering
	\includegraphics[width=1\linewidth]{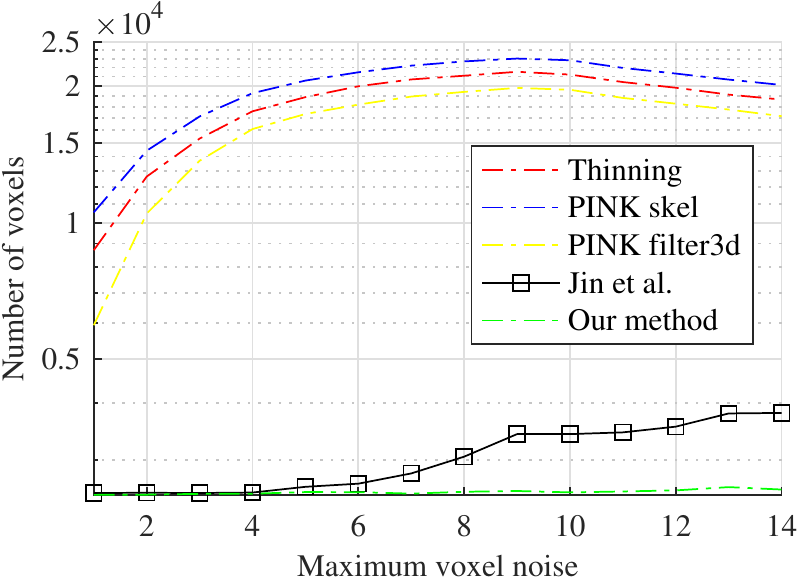}
\caption{\textbf{Best viewed in color.} Number of voxels in the curve skeletons of the synthetic model as a function of the noise level. The vertical axis is shown on a logarithmic scale due to the dramatic differences between our method and most of the comparison approaches.}
\label{fig:voxels}
%%\end{figure*}
\end{figure}

As in the experiments with the real datasets, the comparison methods are characterized by greater numbers of spurious voxels than our method as the noise level increases. We represent this in Figure \ref{fig:voxels}, which shows the number of voxels in the curve skeletons for each of the methods as a function of the noise level.  The figure shows that, as the noise level rises, the number of voxels also increases for all the methods. For the thinning and PINK methods, this increase is dramatic, in some cases up to ~9 times the initial value.  The Jin \etal ~method also shows some additional voxels as the noise rises, particularly for noise levels higher than 6, showing a total increase of approximately 50\% from zero noise to noise level 14. Our method has the least additional voxels, showing a total increase of only 4\%.

\subsection{Tree models acquired in field conditions}
 Figures \ref{fig:resultsA} to \ref{fig:resultsI} show high resolution images of the nine real trees used to evaluate our method as described in Table 1 of the main paper. See the discussion in the main paper in the Experiments section for a more detailed description of the figures below.

\clearpage
\begin{landscape}
%Cropping: “trim=1cm 2cm 3cm 4cm” trims (crops) from left, bottom, right and top by 1, 2, 3 and 4cm respectively. It must be accompanied by “clip=true”.
\renewcommand*{\factorTreeB}{0.30}

%%%%%%%%%%%%%%%%%%%%%%%%%%%% A %%%%%%%%%%%%%%%%%%%%%%%%%%%%%%%
\begin{figure}
\centering
\subfloat[Surface]  
	{  
	\includegraphics[trim = 0cm 0cm 0cm 0cm,clip, width=\factorTreeB\linewidth]{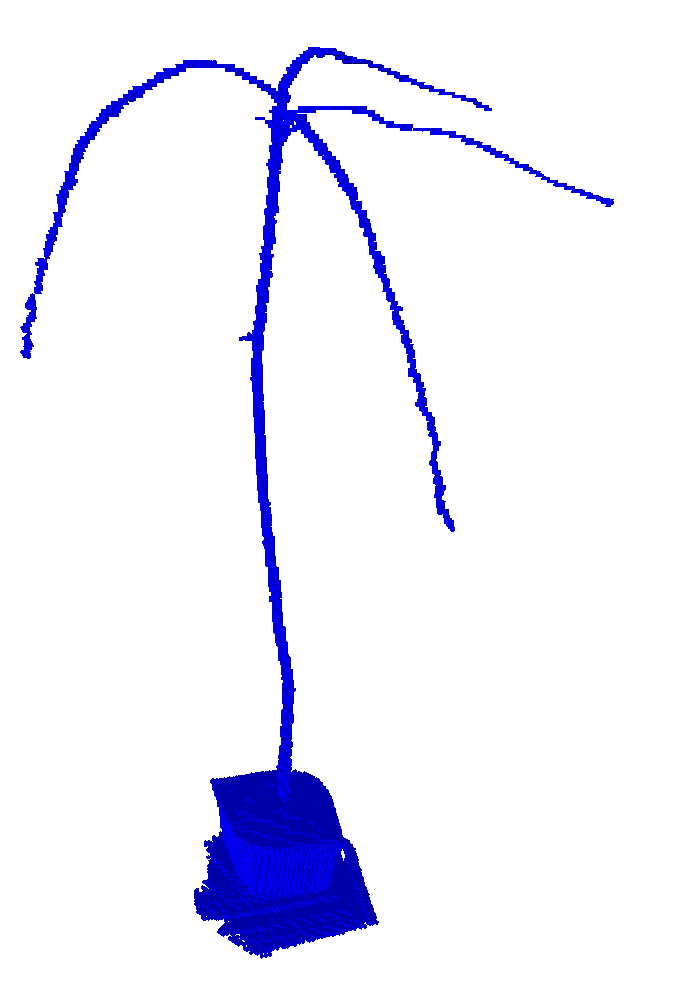}
}
\subfloat[Thinning]  
	{  
	\includegraphics[trim = 0cm 0cm 0cm 0cm,clip, width=\factorTreeB\linewidth]{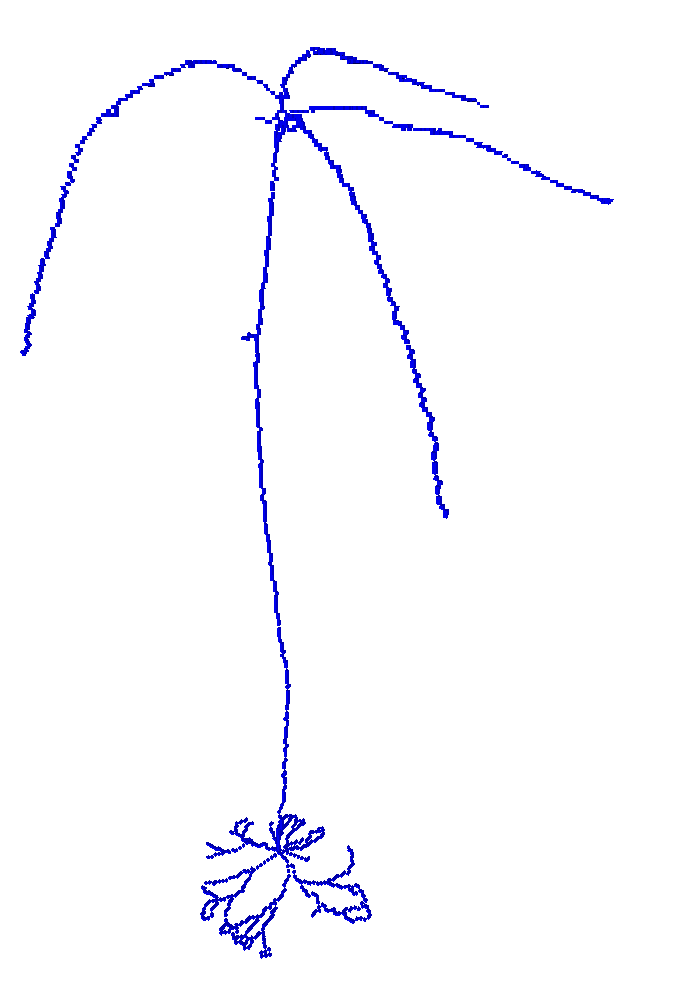}
}
\subfloat[PINK skel]  
	{  
	\includegraphics[trim = 0cm 0cm 0cm 0cm,clip, width=\factorTreeB\linewidth]{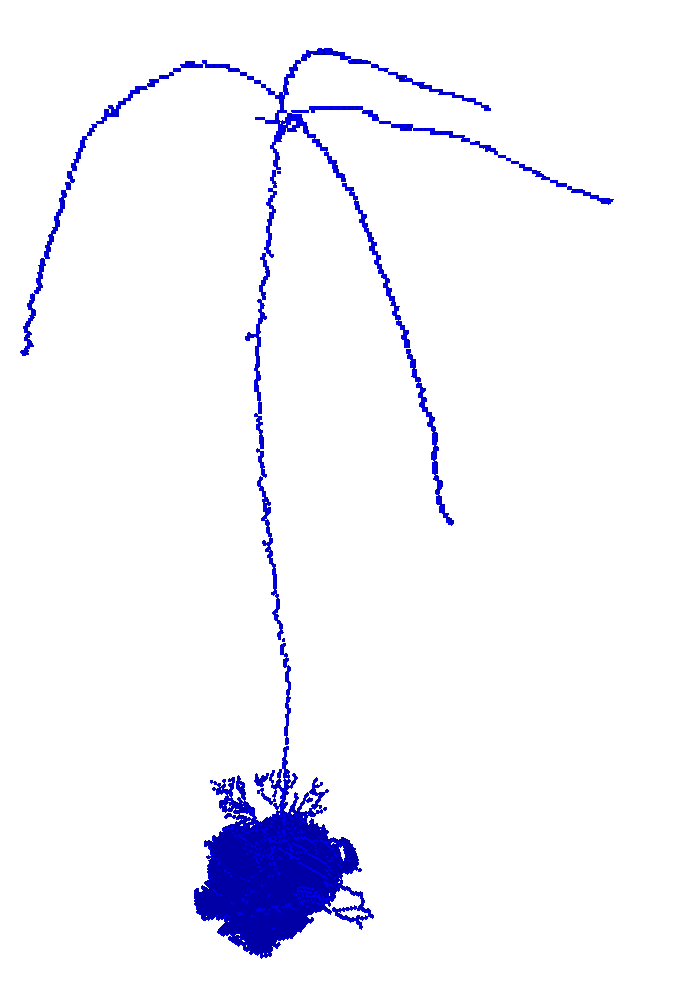}
}
\caption{\textbf{Best viewed in color.} Original surface, comparison curve skeletons, and curve skeleton computed with our method,  for Dataset A (part 1 of 2).}\label{fig:resultsA}
\end{figure}

\clearpage

\begin{figure}
\subfloat[PINK filter3d]  
	{  
	\includegraphics[trim = 0cm 0cm 0cm 0cm,clip, width=\factorTreeB\linewidth]{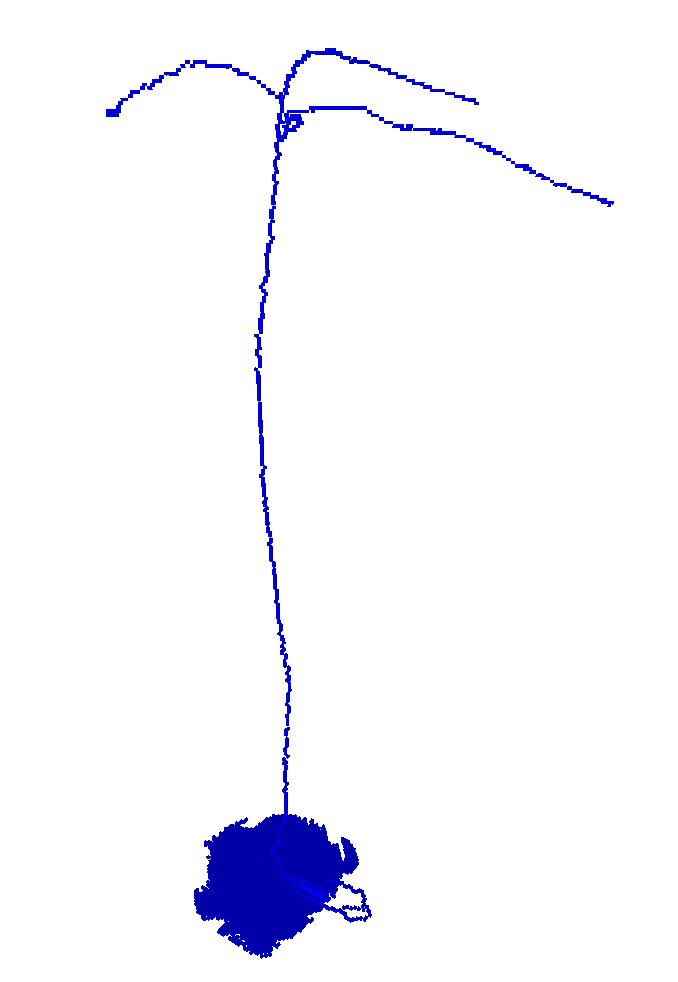}
}
\subfloat[Jin \etal]  
	{  
	\includegraphics[trim = 0cm 0cm 0cm 0cm,clip, width=\factorTreeB\linewidth]{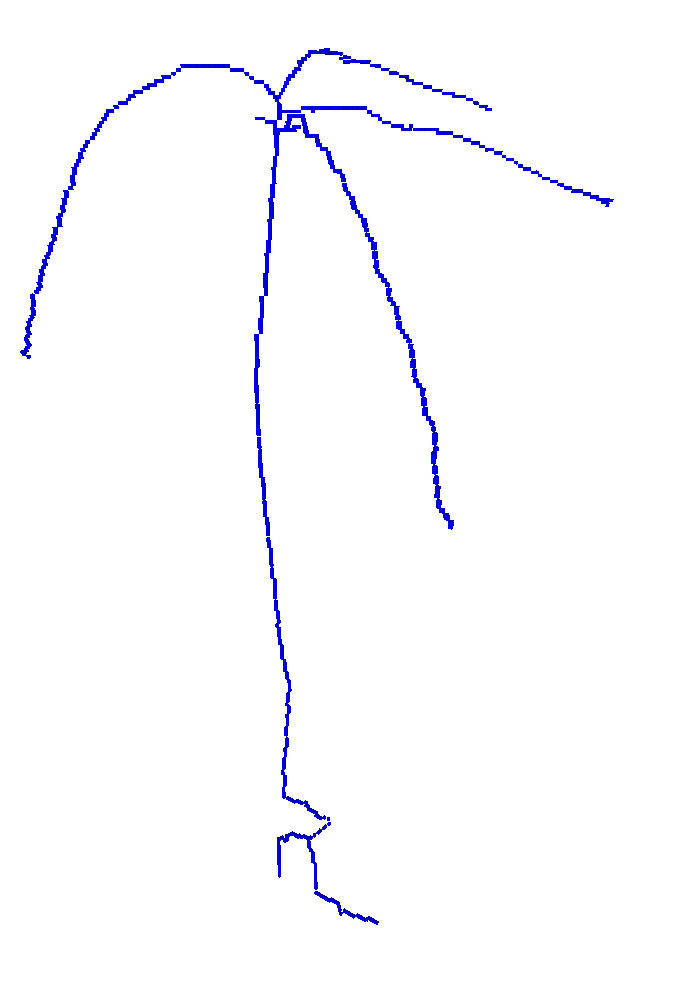}
}
\subfloat[Our method]  
	{  
	\includegraphics[trim = 0cm 0cm 0cm 0cm,clip, width=\factorTreeB\linewidth]{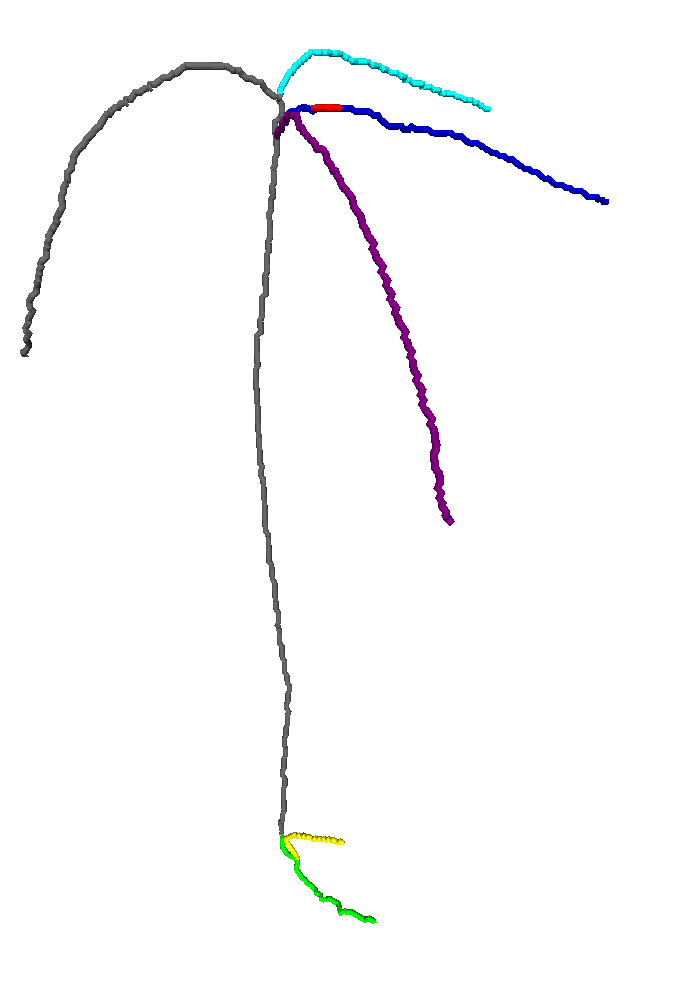}
}
\caption{\textbf{Best viewed in color.} Original surface, comparison curve skeletons, and curve skeleton computed with our method,  for Dataset A (part 2 of 2).}
\label{fig:resultsII}
\end{figure}
\clearpage

%%%%%%%%%%%%%%%%%%%%%%%%%% B %%%%%%%%%%%%%%%%%%%%%%%%%%%%%%%%%
\begin{figure}
\centering
\subfloat[Surface]  
	{  
	\includegraphics[trim = 0cm 2cm 1cm 0cm,clip, width=\factorTreeB\linewidth]{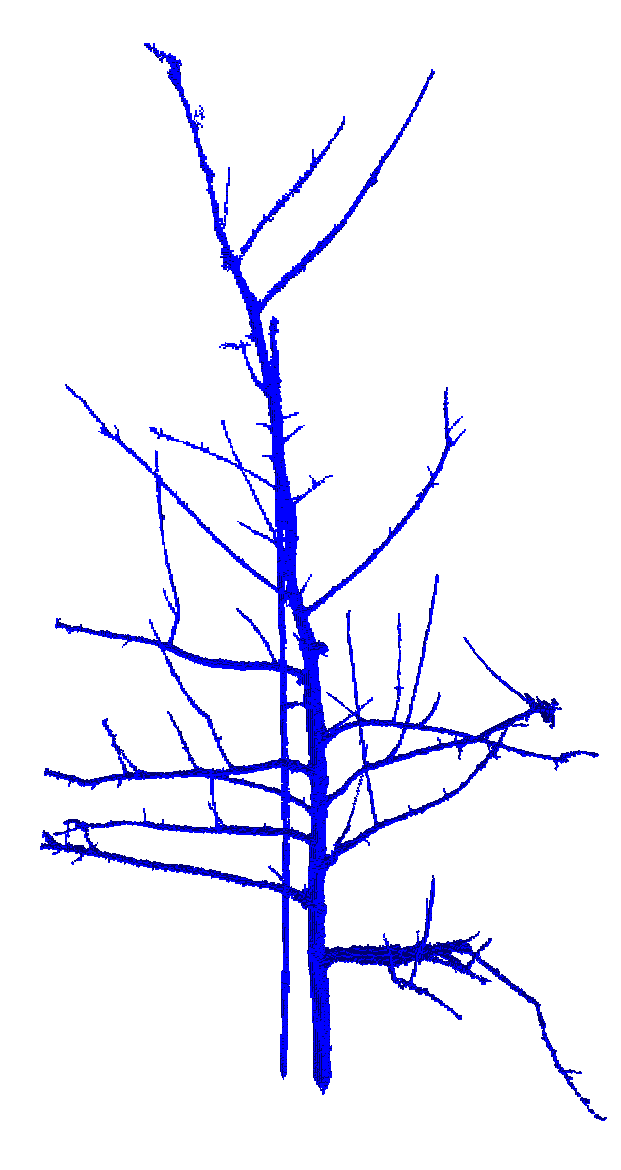}
}
\subfloat[Thinning]  
	{  
	\includegraphics[trim = 0cm 2cm 1cm 0cm,clip, width=\factorTreeB\linewidth]{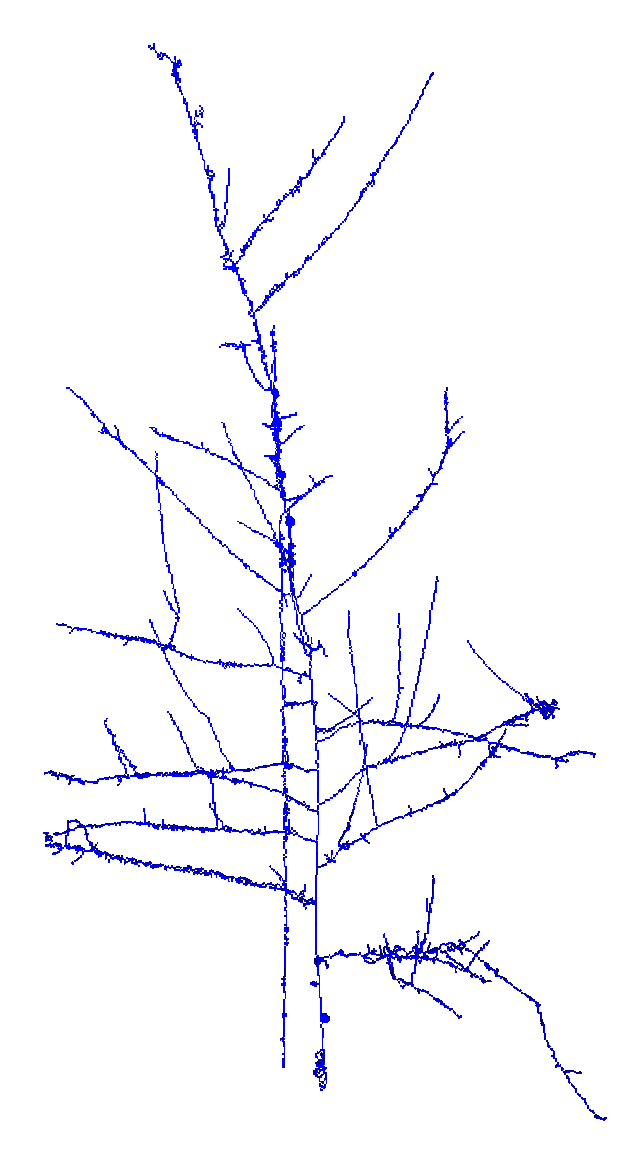}
}
\subfloat[PINK skel]  
	{  
	\includegraphics[trim = 0cm 2cm 1cm 0cm,clip, width=\factorTreeB\linewidth]{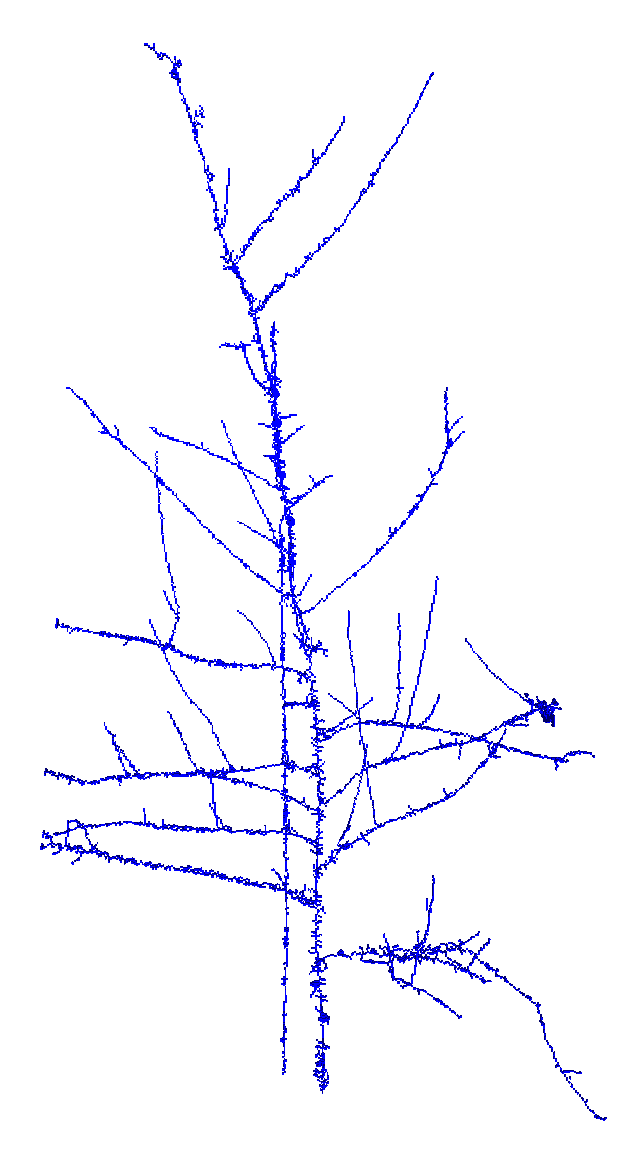}
}
\caption{\textbf{Best viewed in color.} Original surface, comparison curve skeletons, and curve skeleton computed with our method,  for Dataset B (part 1 of 2).}
\end{figure}

\clearpage

\begin{figure}
\subfloat[PINK filter3d]  
	{  
	\includegraphics[trim = 0cm 2cm 1cm 0cm,clip, width=\factorTreeB\linewidth]{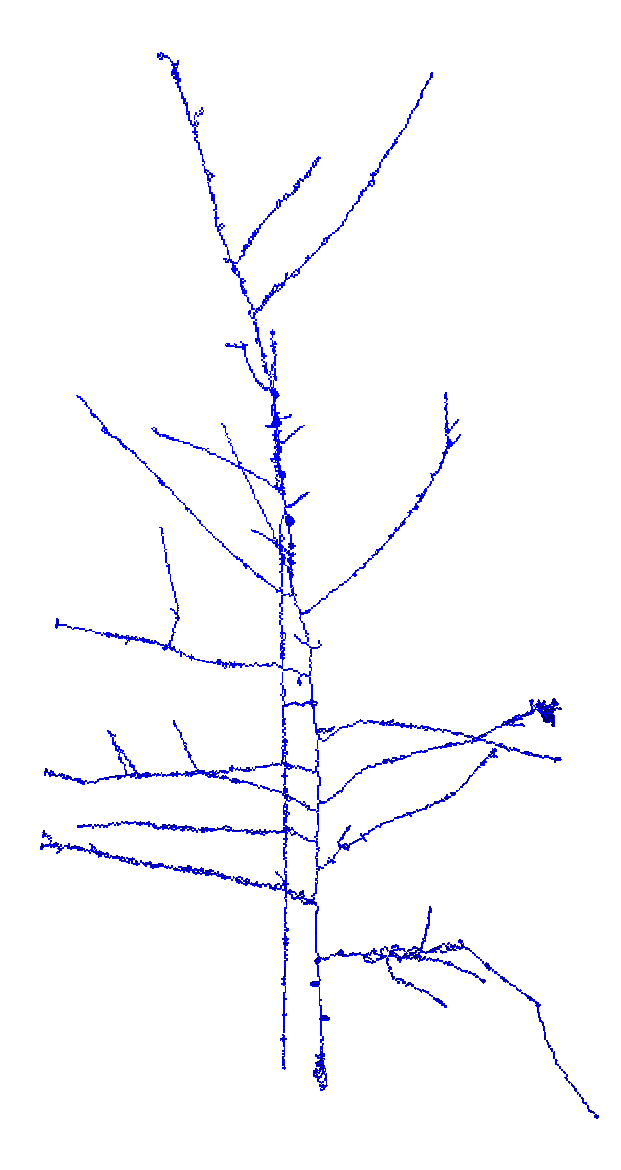}
}
\subfloat[Jin \etal]  
	{  
	\includegraphics[trim = 0cm 2cm 1cm 0cm,clip, width=\factorTreeB\linewidth]{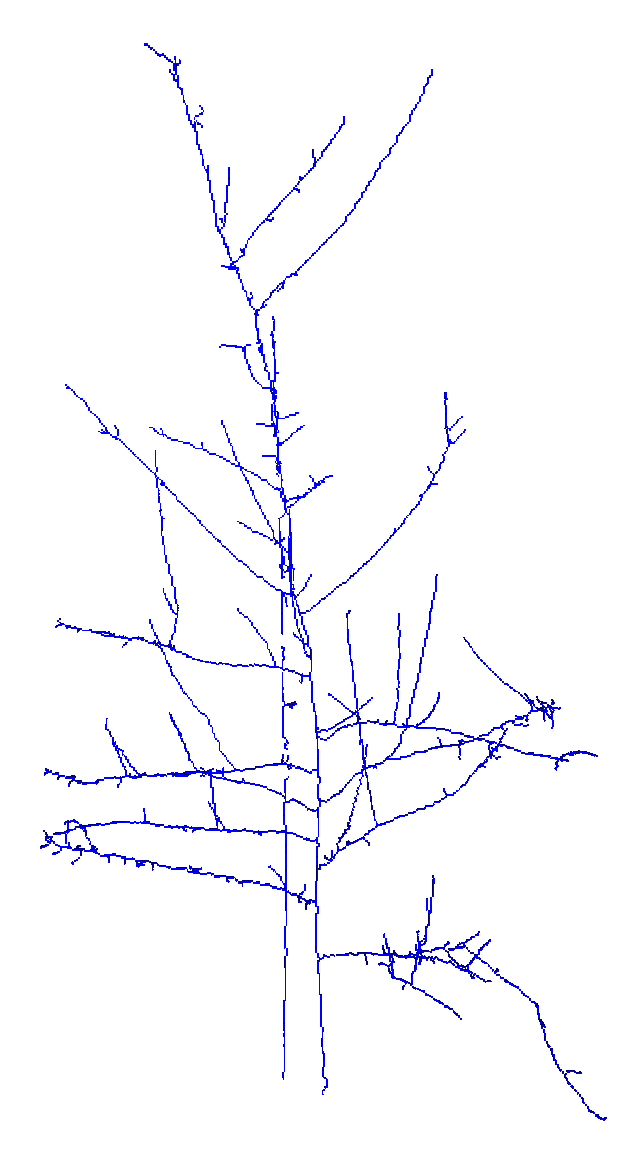}
}
\subfloat[Our method]  
	{  
	\includegraphics[trim = 0cm 2cm 1cm 0cm,clip, width=\factorTreeB\linewidth]{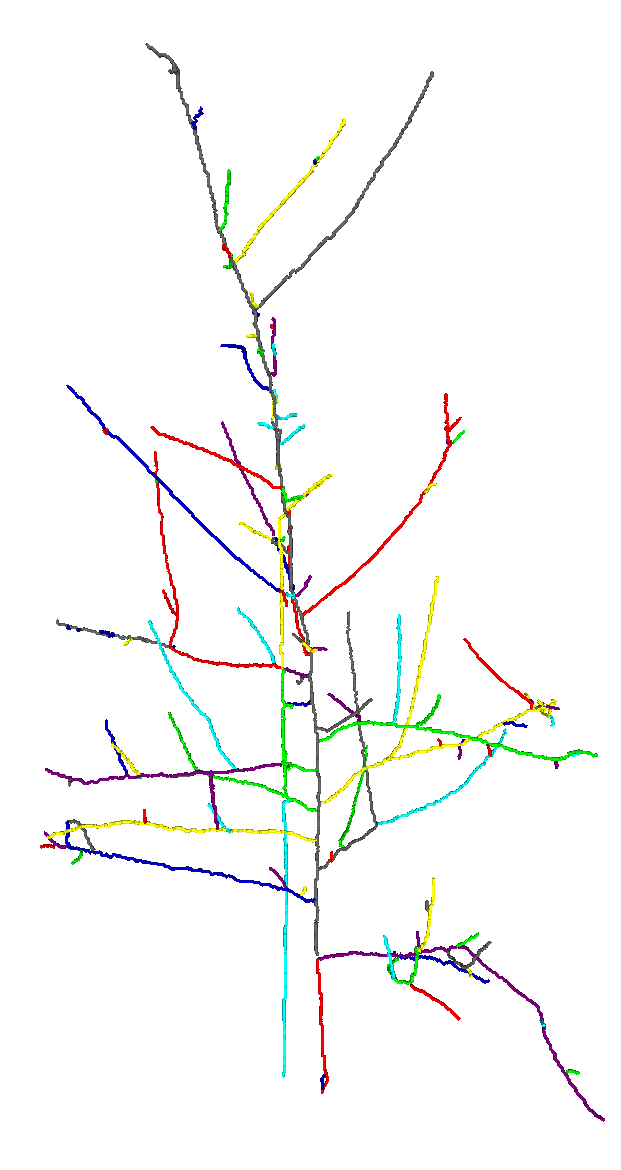}
}
\caption{\textbf{Best viewed in color.} Original surface, comparison curve skeletons, and curve skeleton computed with our method,  for Dataset B (part 2 of 2).}
\label{fig:resultsII}
\end{figure}
\clearpage

%%%%%%%%%%%%%%%%%%%%%%%  C %%%%%%%%%%%%%%%%%%%%%%%%%
\begin{figure}
\centering
\subfloat[Surface]  
	{  
	\includegraphics[trim = 1cm 2.5cm 3.5cm 3cm,clip, width=\factorTreeB\linewidth]{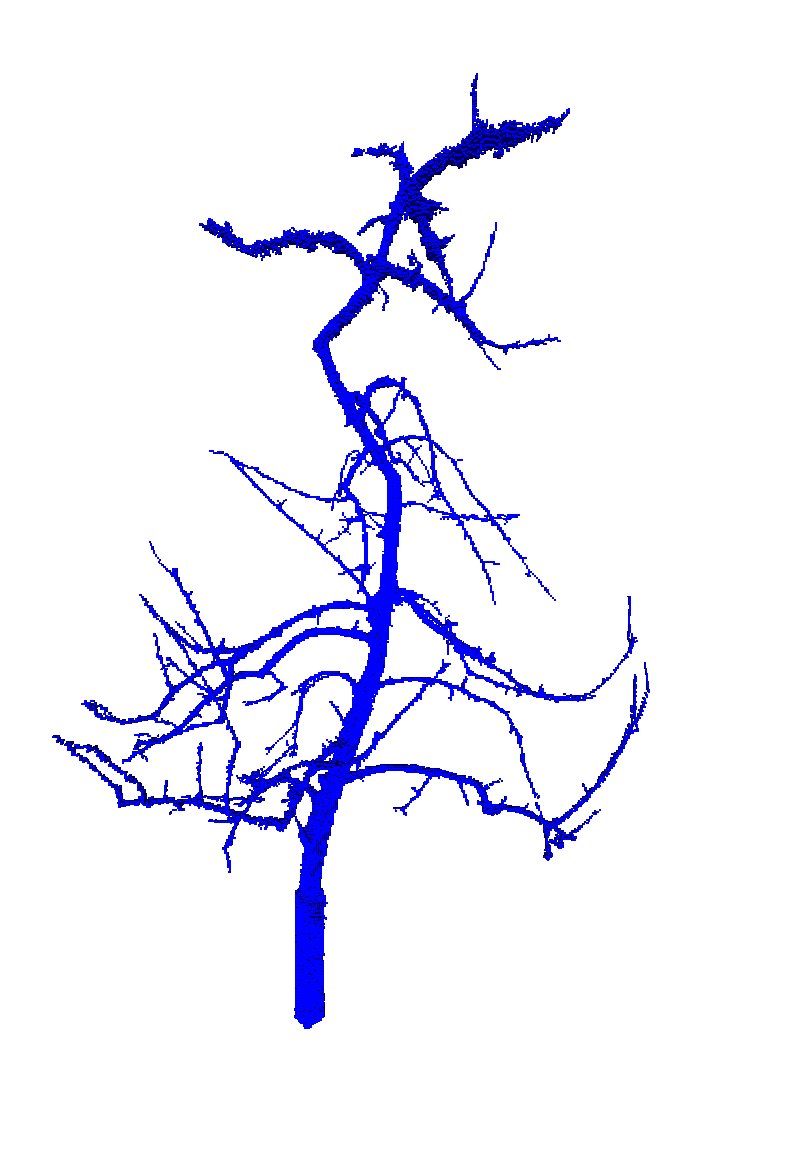}
}
\subfloat[Thinning]  
	{  
	\includegraphics[trim =1cm 2.5cm 3.5cm 3cm,clip, width=\factorTreeB\linewidth]{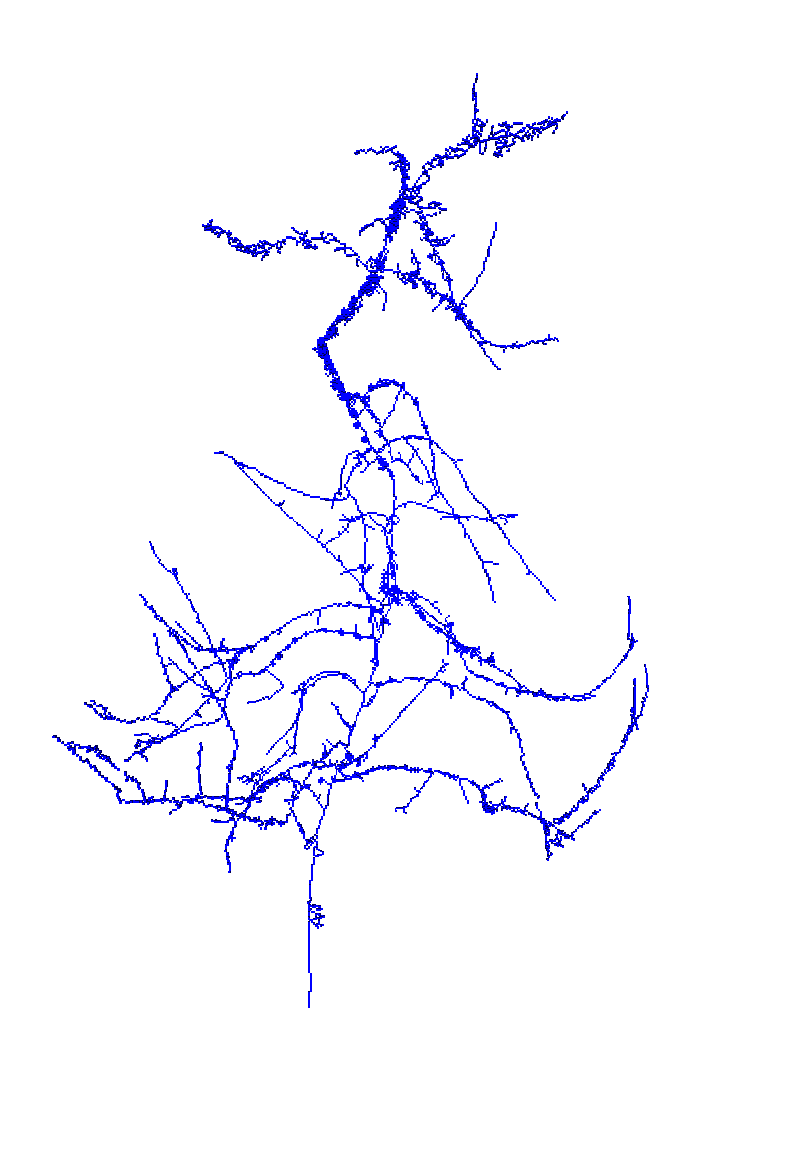}
}
\subfloat[PINK skel]  
	{  
	\includegraphics[trim = 1cm 2.5cm 3.5cm 3cm,clip, width=\factorTreeB\linewidth]{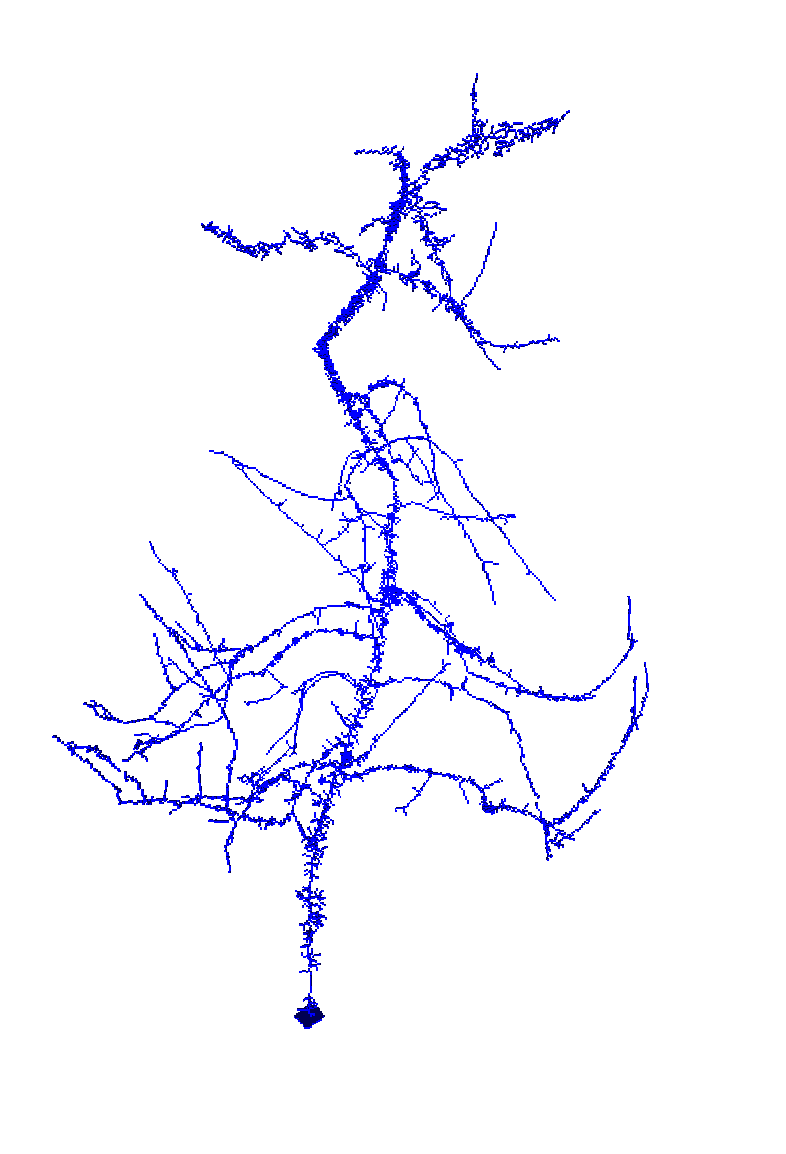}
}
\caption{\textbf{Best viewed in color.} Original surface, comparison curve skeletons, and curve skeleton computed with our method,  for Dataset C (part 1 of 2).}
\end{figure}

\clearpage

\begin{figure}
\subfloat[PINK filter3d]  
	{  
	\includegraphics[trim = 1cm 2.5cm 3.5cm 3cm,clip, width=\factorTreeB\linewidth]{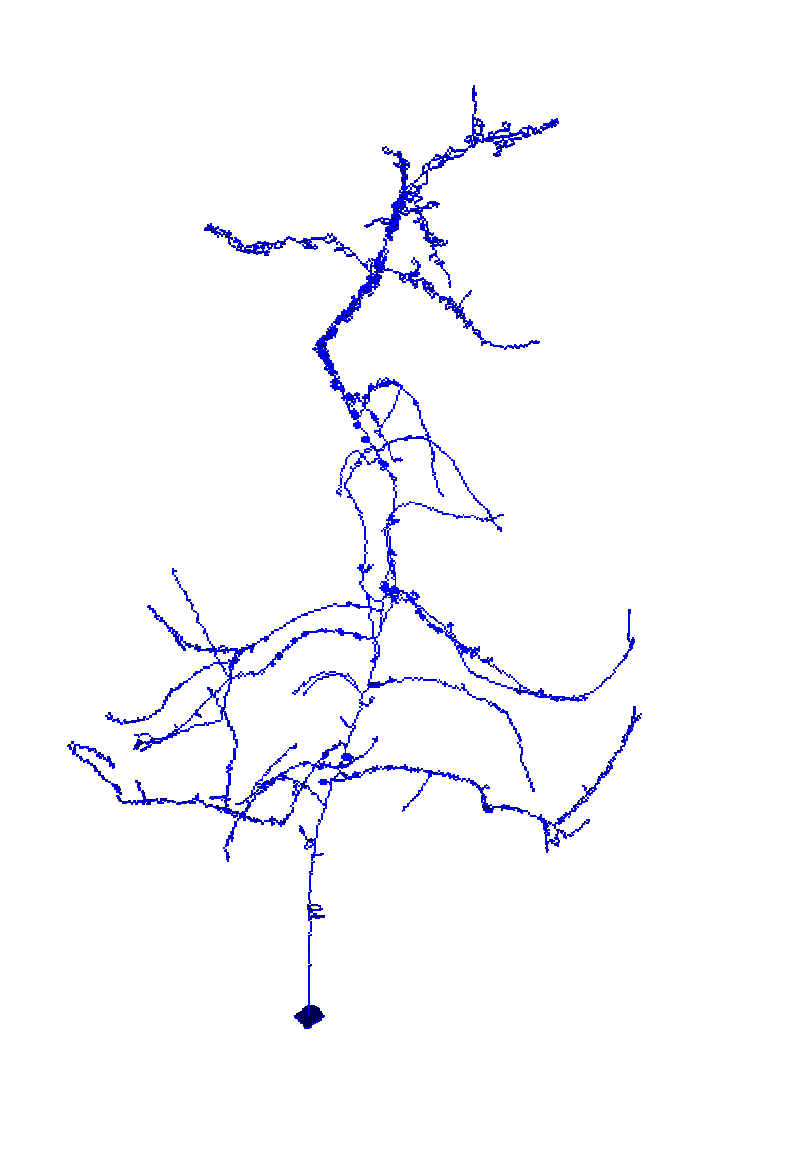}
}
\subfloat[Jin \etal]  
	{  
	\includegraphics[trim = 1cm 2.5cm 3.5cm 3cm,clip, width=\factorTreeB\linewidth]{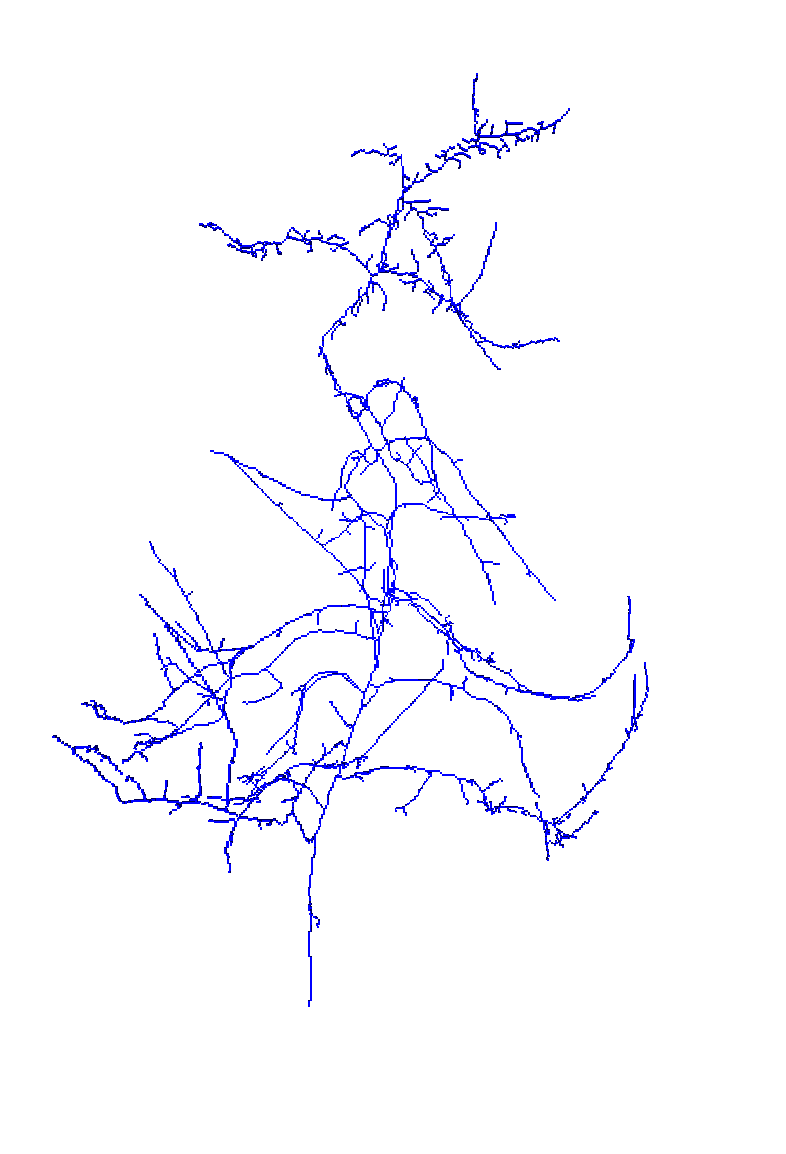}
}
\subfloat[Our method]  
	{  
	\includegraphics[trim = 1cm 2.5cm 3.5cm 3cm,clip, width=\factorTreeB\linewidth]{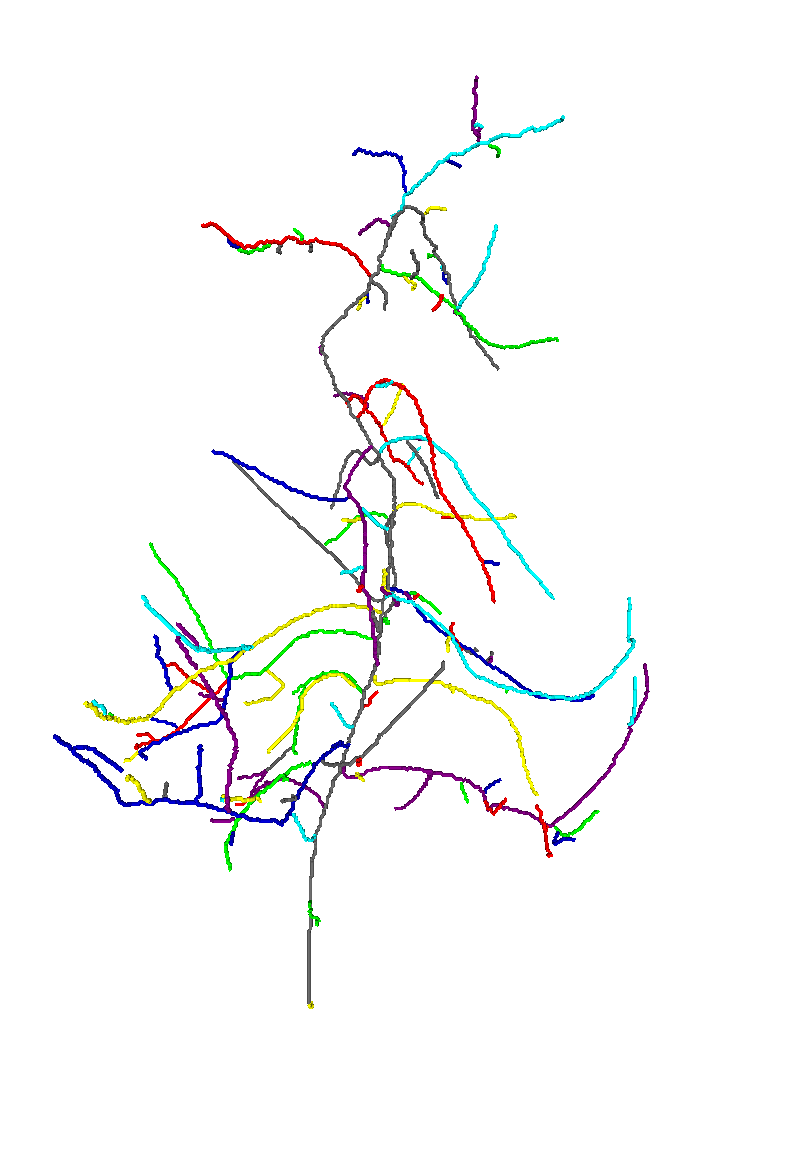}
}
\caption{\textbf{Best viewed in color.} Original surface, comparison curve skeletons, and curve skeleton computed with our method,  for Dataset C (part 2 of 2).}
\label{fig:resultsIII}
\end{figure}
\clearpage

%%%%%%%%%%%%%%%%%%%%%%%%%%%%%%%% D %%%%%%%%%%%%%%%%%%%%%%%%%%%%%%%%%%%%
\begin{figure}
\centering
\subfloat[Surface]  
	{  
	\includegraphics[trim = 0cm 0cm 0cm 0cm,clip, width=\factorTreeB\linewidth]{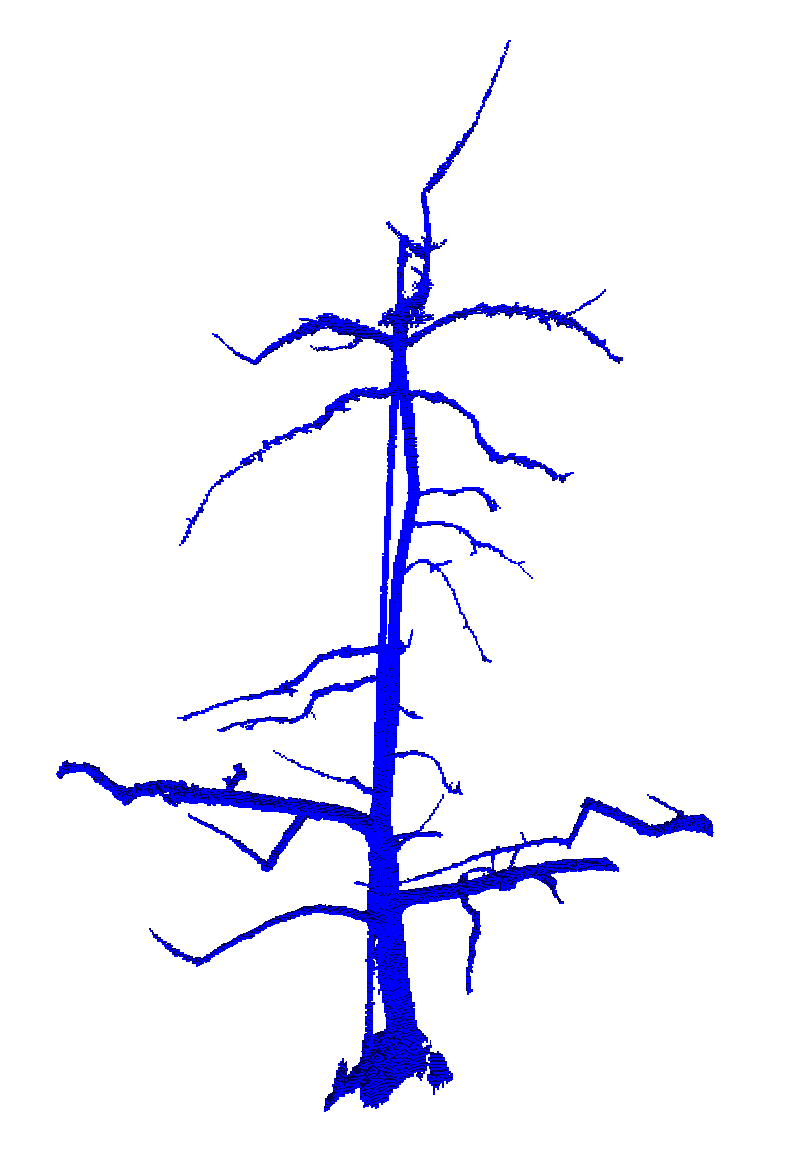}
}
\subfloat[Thinning]  
	{  
	\includegraphics[trim = 0cm 0cm 0cm 0cm,clip, width=\factorTreeB\linewidth]{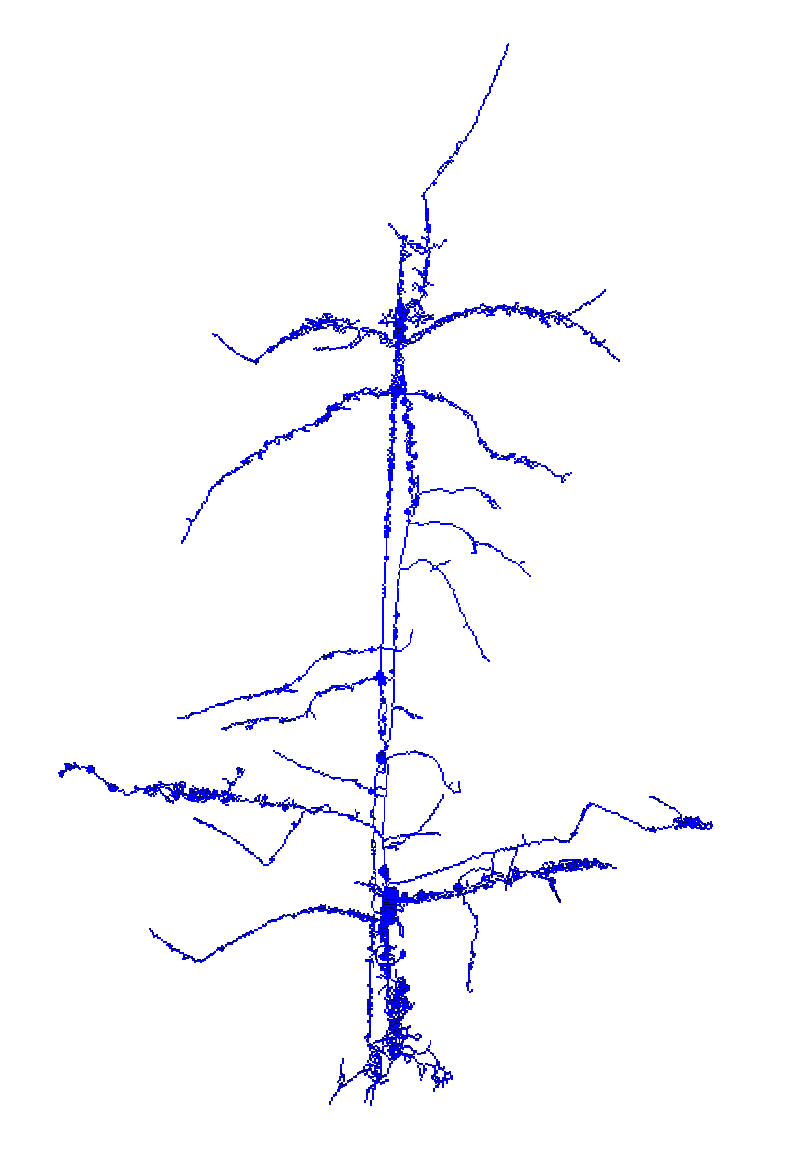}
}
\subfloat[PINK skel]  
	{  
	\includegraphics[trim = 0cm 0cm 0cm 0cm,clip, width=\factorTreeB\linewidth]{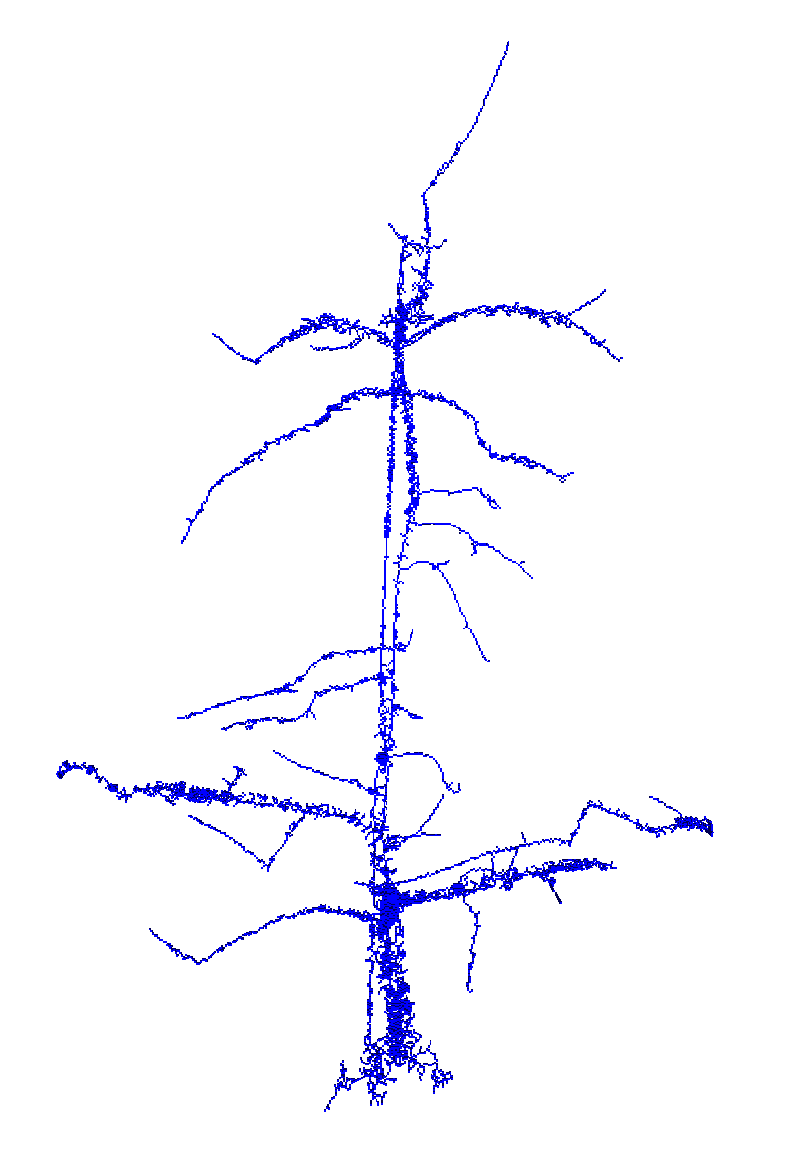}
}
\caption{\textbf{Best viewed in color.} Original surface, comparison curve skeletons, and curve skeleton computed with our method,  for Dataset D (part 1 of 2).}
\end{figure}

\clearpage

\begin{figure}
\subfloat[PINK filter3d]  
	{  
	\includegraphics[trim = 0cm 0cm 0cm 0cm,clip, width=\factorTreeB\linewidth]{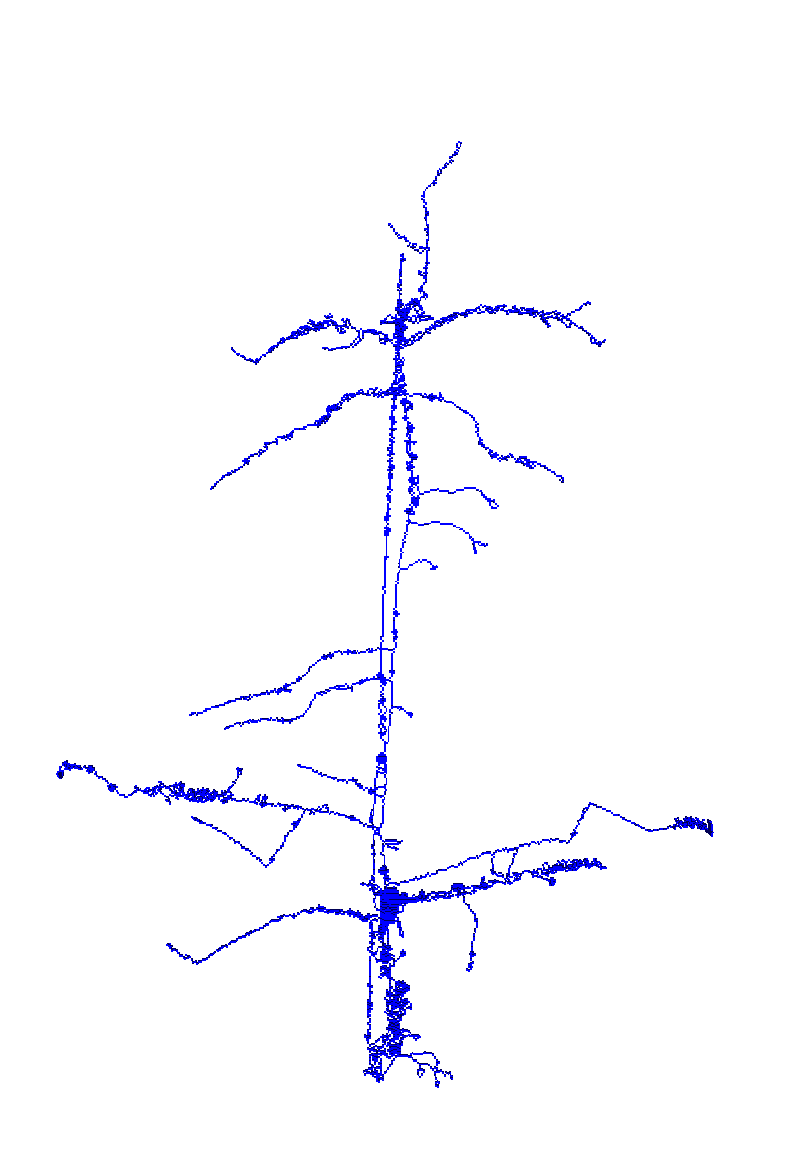}
}
\subfloat[Jin \etal]  
	{  
	\includegraphics[trim = 0cm 0cm 0cm 0cm,clip, width=\factorTreeB\linewidth]{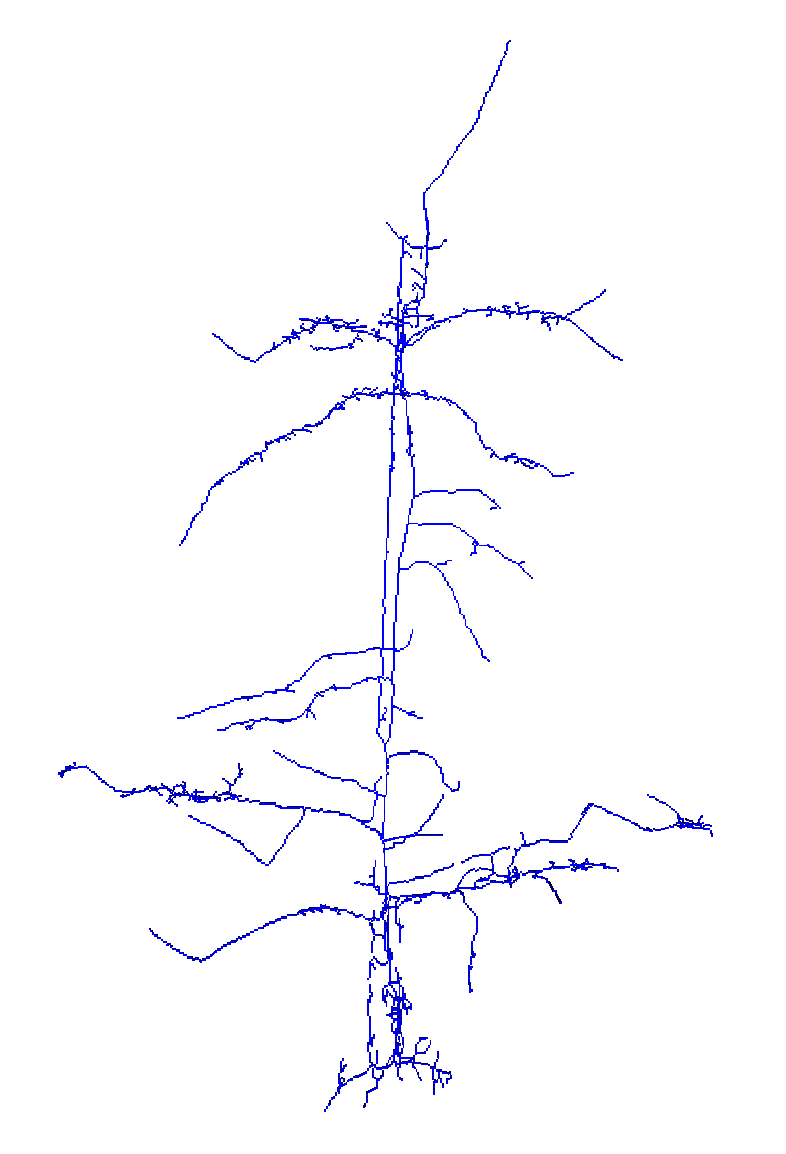}
}
\subfloat[Our method]  
	{  
	\includegraphics[trim = 0cm 0cm 0cm 0cm,clip, width=\factorTreeB\linewidth]{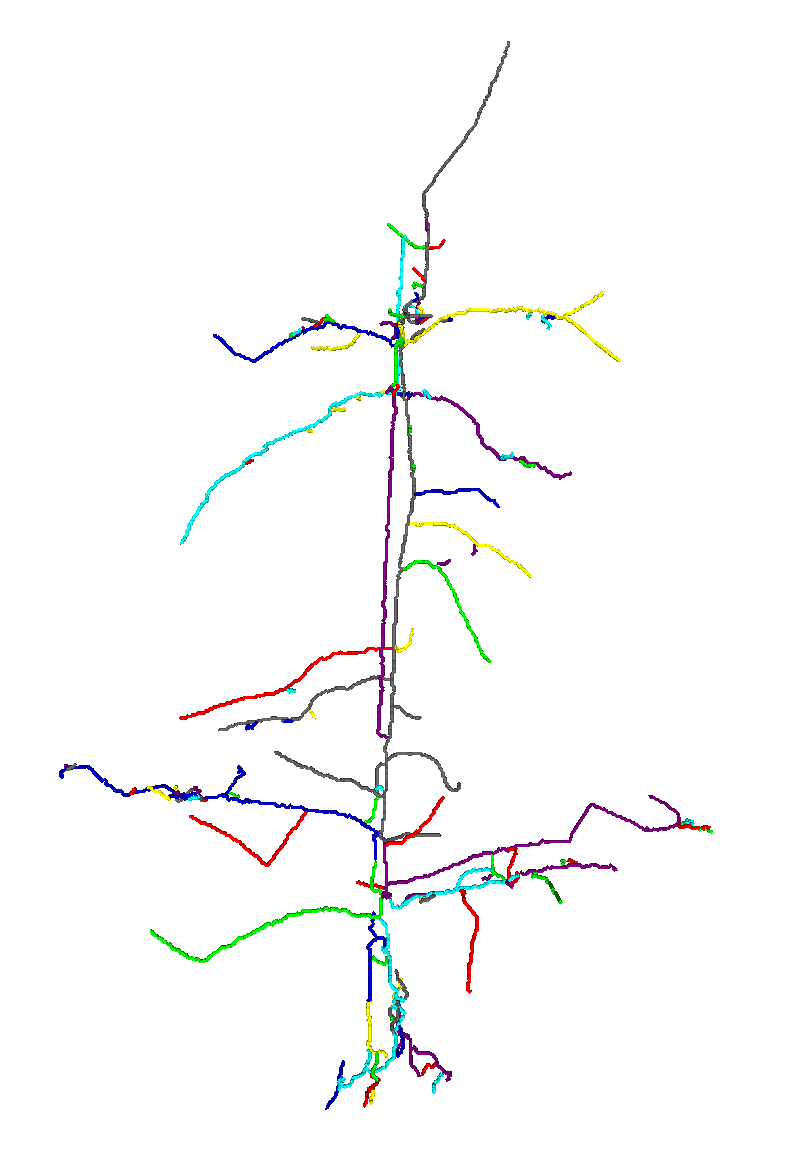}
}
\caption{\textbf{Best viewed in color.} Original surface, comparison curve skeletons, and curve skeleton computed with our method,  for Dataset D (part 2 of 2).}
\label{fig:resultsIV}
\end{figure}
\clearpage

%%%%%%%%%%%%%%%%%%%%%%% E %%%%%%%%%%%%%%%%%%%%%%%%%
\begin{figure}
\centering
\subfloat[Surface]  
	{  
	\includegraphics[trim = 0cm 0cm 0cm 0cm,clip, width=\factorTreeB\linewidth]{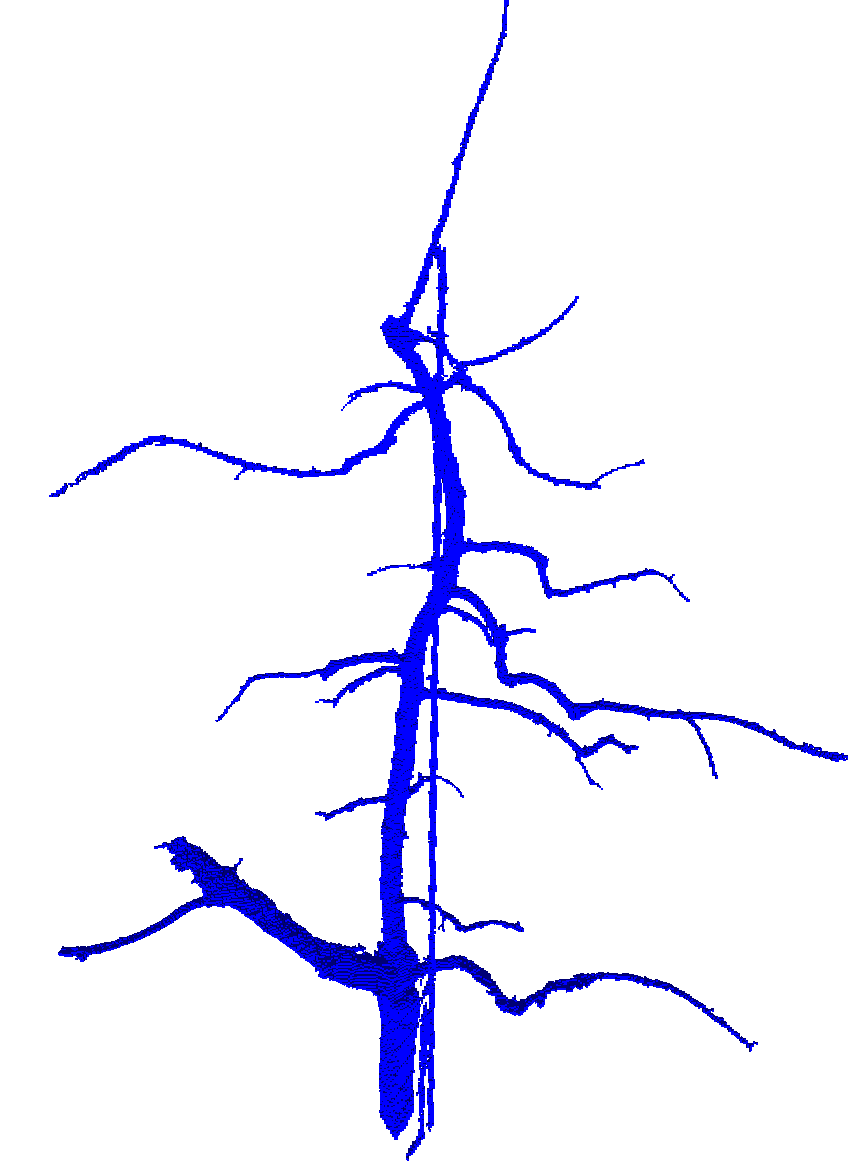}
}
\subfloat[Thinning]  
	{  
	\includegraphics[trim =0cm 0cm 0cm 0cm,clip, width=\factorTreeB\linewidth]{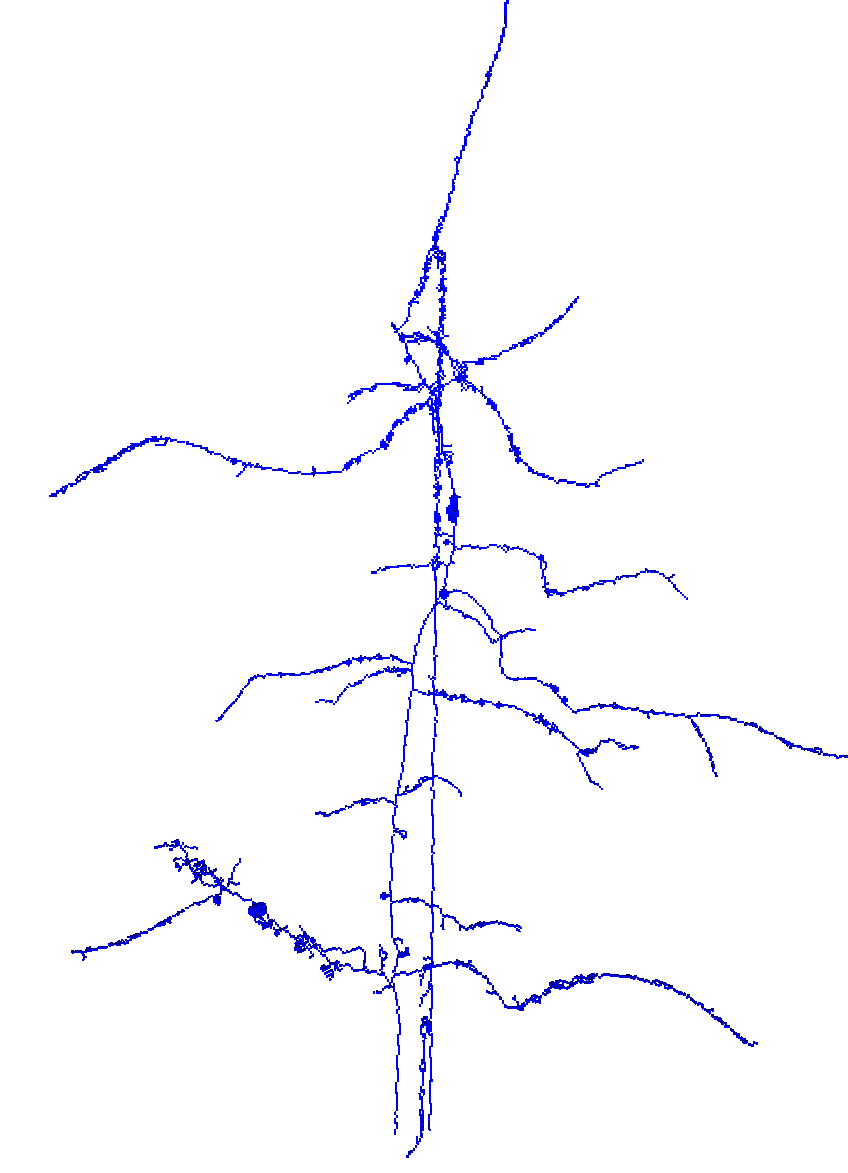}
}
\subfloat[PINK skel]  
	{  
	\includegraphics[trim = 0cm 0cm 0cm 0cm,clip, width=\factorTreeB\linewidth]{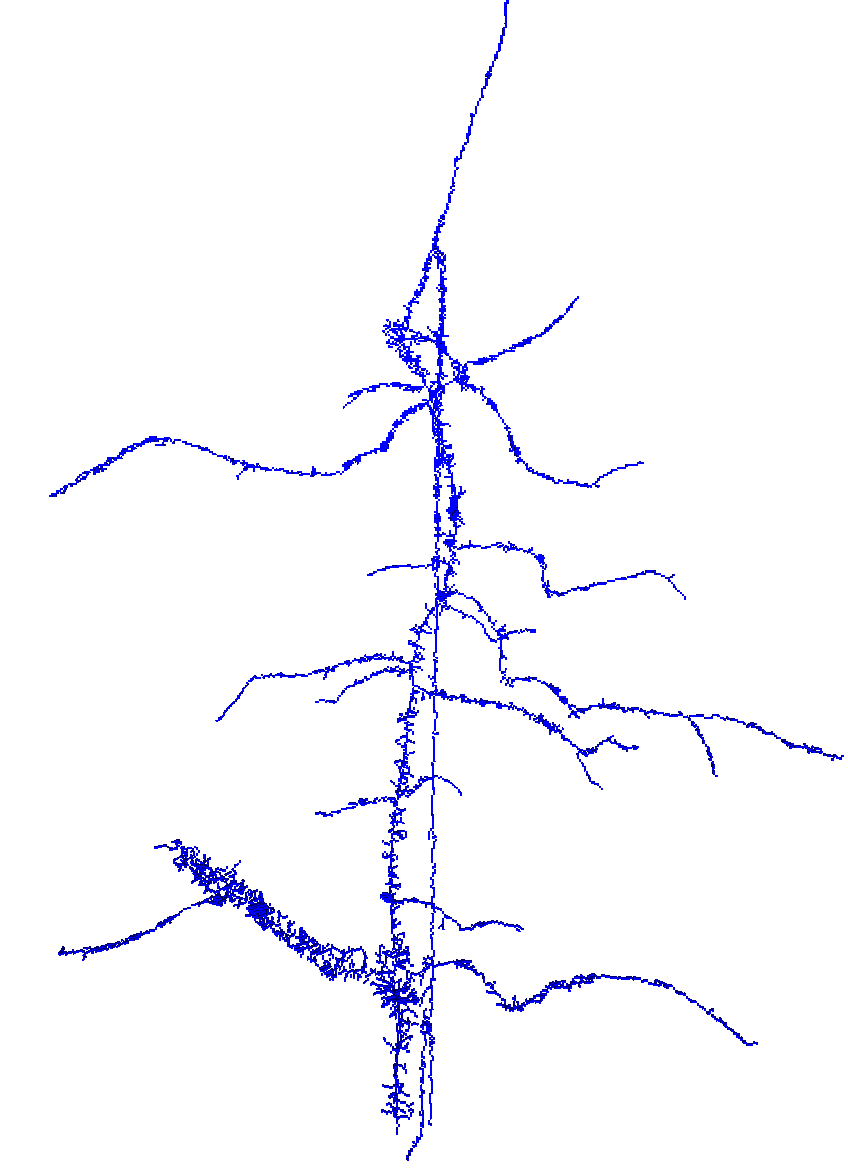}
}
\caption{\textbf{Best viewed in color.} Original surface, comparison curve skeletons, and curve skeleton computed with our method,  for Dataset E (part 1 of 2).}
\end{figure}

\clearpage

\begin{figure}
\subfloat[PINK filter3d]  
	{  
	\includegraphics[trim = 0cm 0cm 0cm 0cm,clip, width=\factorTreeB\linewidth]{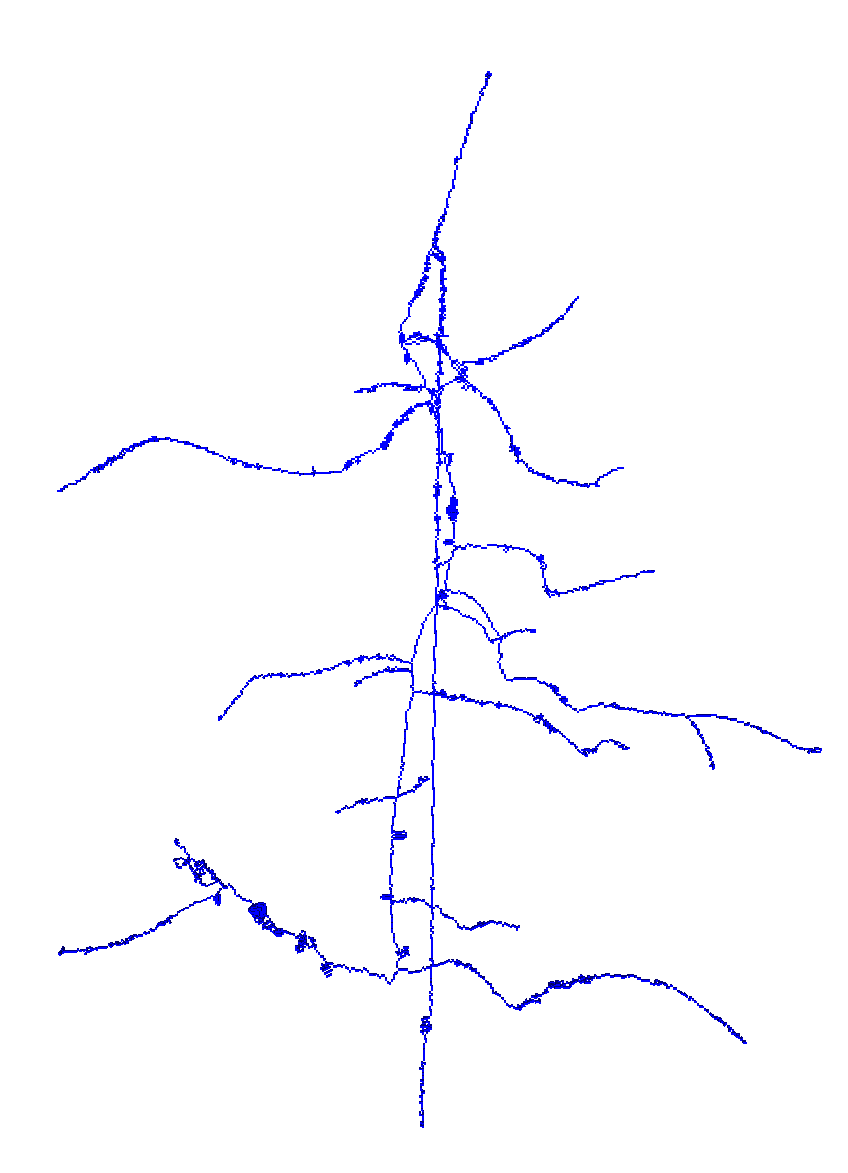}
}
\subfloat[Jin \etal]  
	{  
	\includegraphics[trim = 0cm 0cm 0cm 0cm,clip, width=\factorTreeB\linewidth]{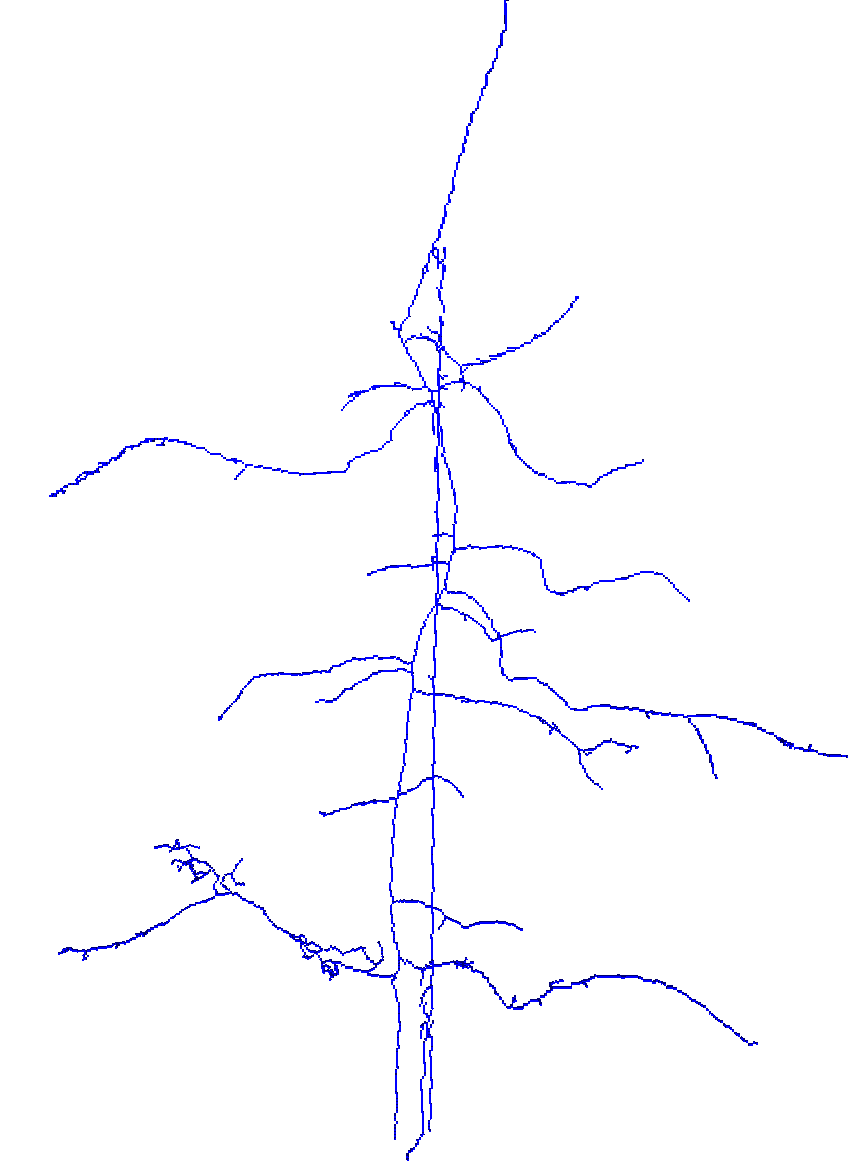}
}
\subfloat[Our method]  
	{  
	\includegraphics[trim = 0cm 0cm 0cm 0cm,clip, width=\factorTreeB\linewidth]{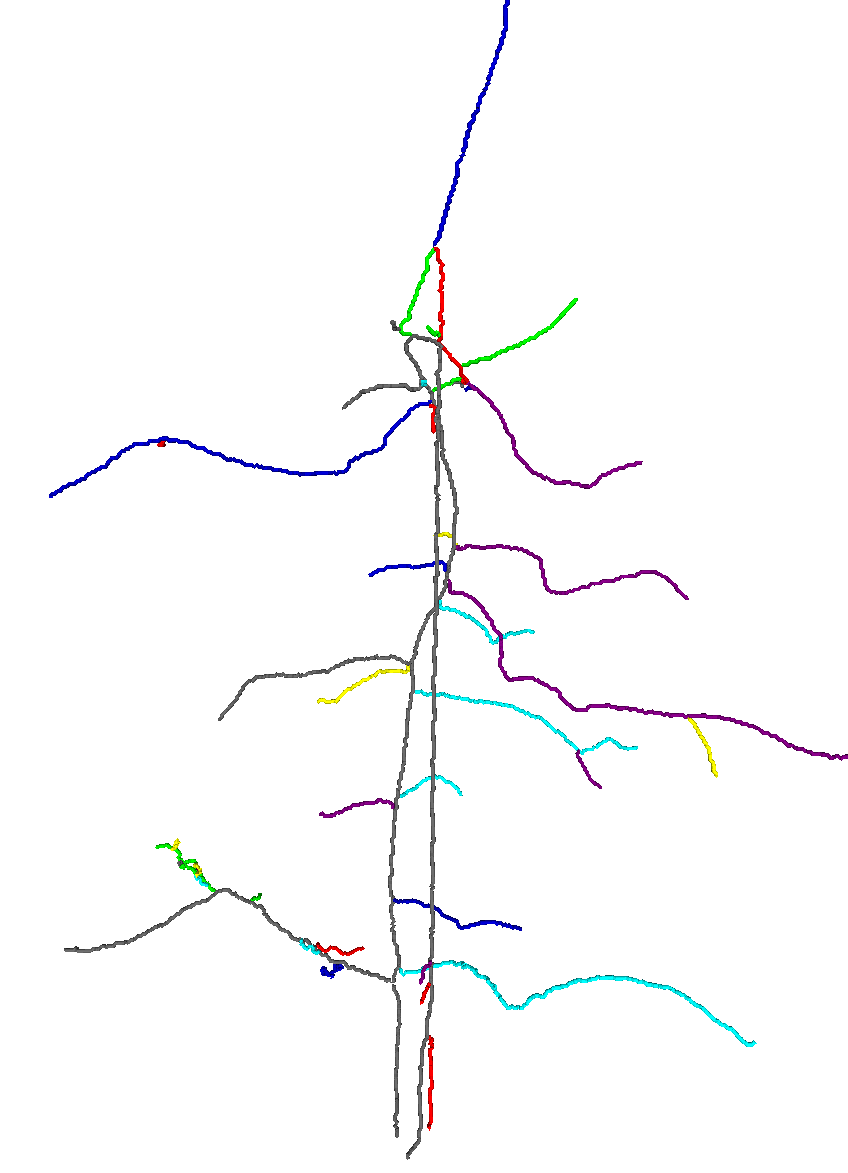}
}
\caption{\textbf{Best viewed in color.} Original surface, comparison curve skeletons, and curve skeleton computed with our method,  for Dataset E (part 2 of 2).}
\label{fig:resultsV}
\end{figure}
\clearpage

%%%%%%%%%%%%%%%%%%%%%%%  F %%%%%%%%%%%%%%%%%%%%%%%%%
\begin{figure}
\centering
\subfloat[Surface]  
	{  
	\includegraphics[trim = 0cm 0cm 0cm 0cm,clip, width=\factorTreeB\linewidth]{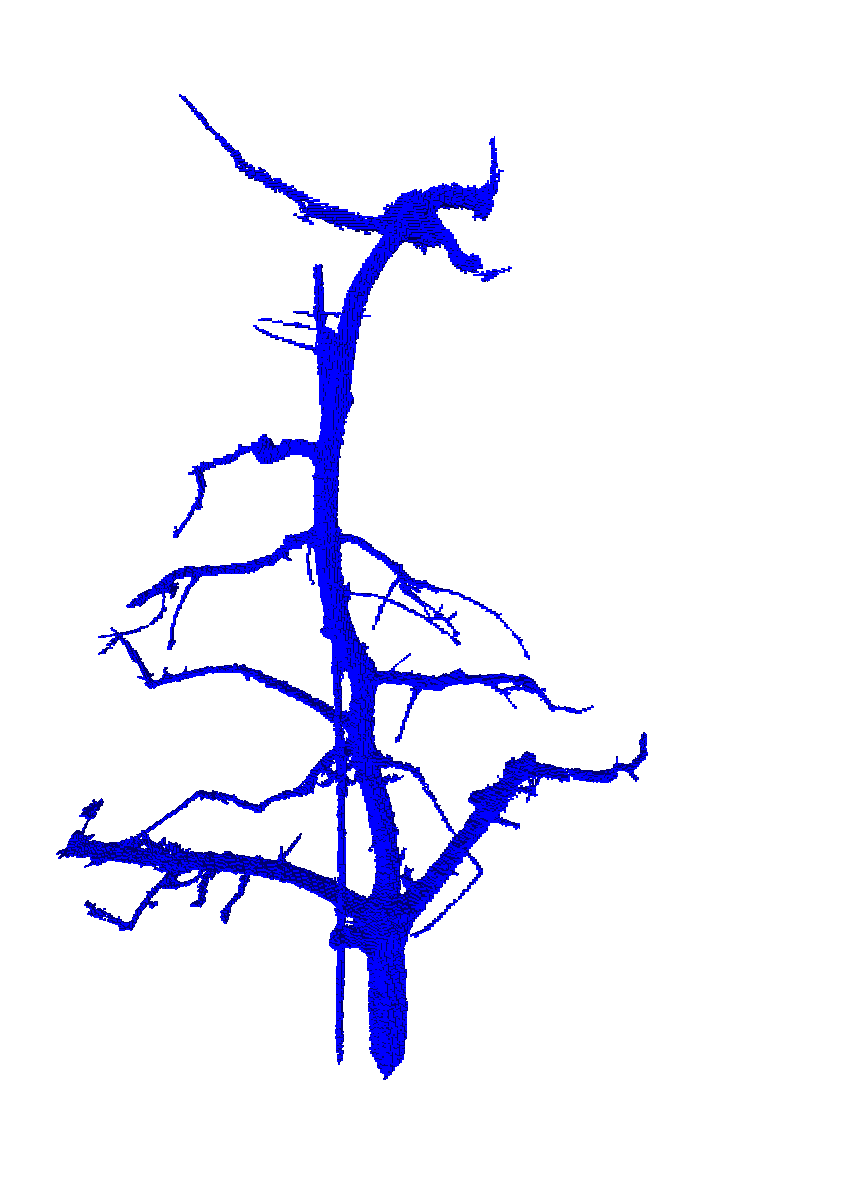}
}
\subfloat[Thinning]  
	{  
	\includegraphics[trim =0cm 0cm 0cm 0cm,clip, width=\factorTreeB\linewidth]{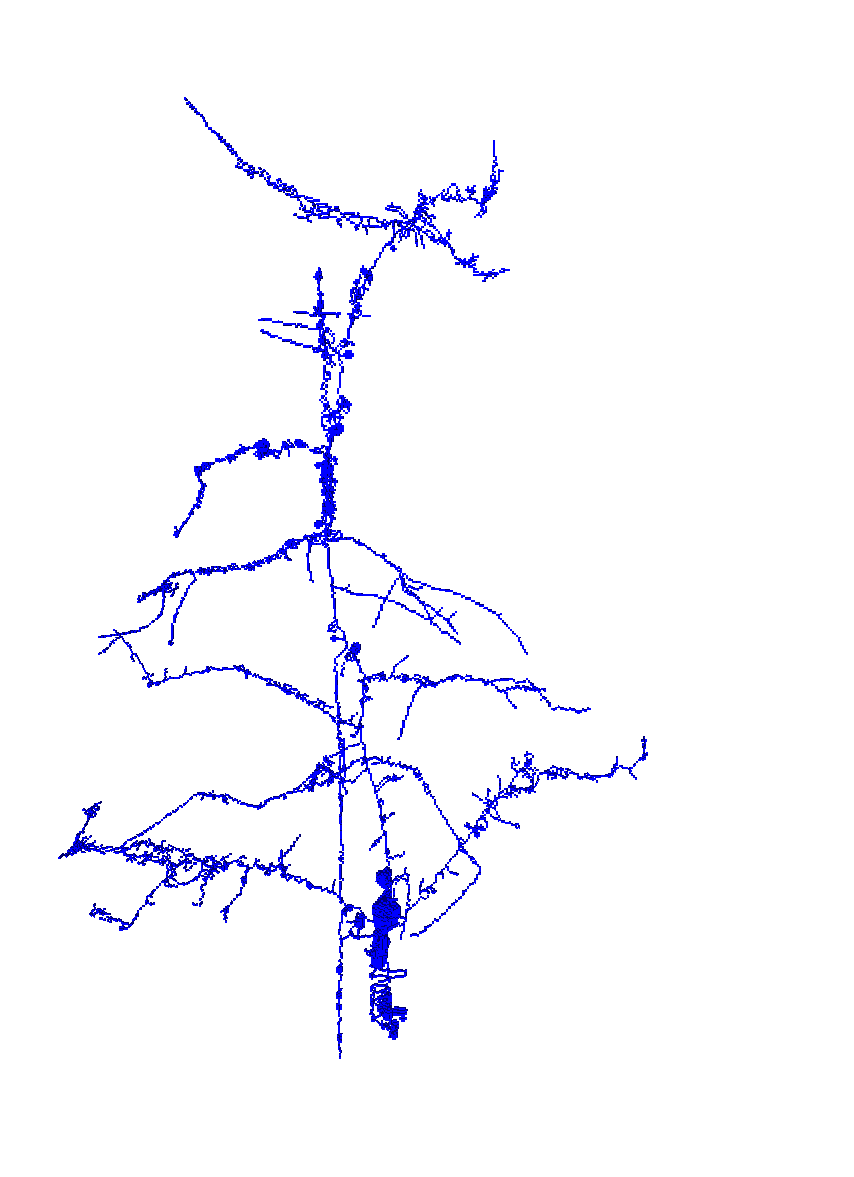}
}
\subfloat[PINK skel]  
	{  
	\includegraphics[trim = 0cm 0cm 0cm 0cm,clip, width=\factorTreeB\linewidth]{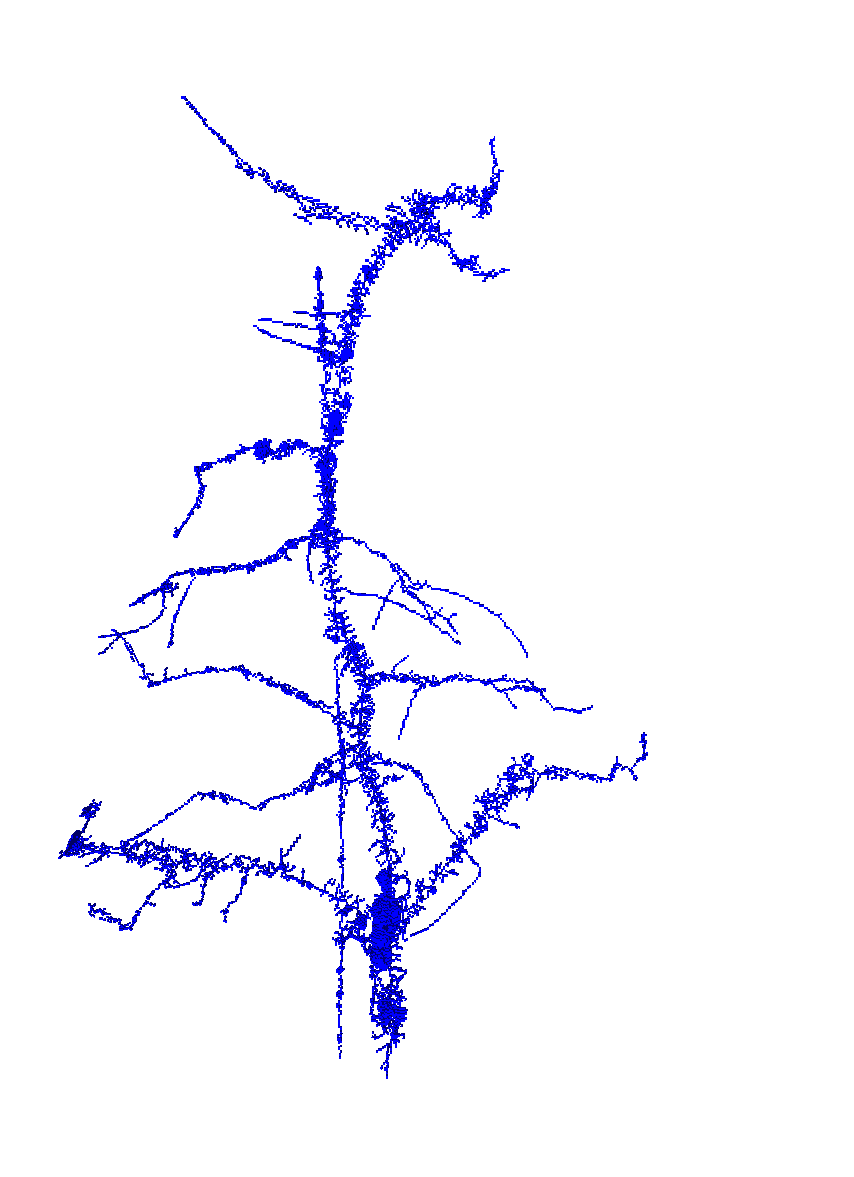}
}
\caption{\textbf{Best viewed in color.} Original surface, comparison curve skeletons, and curve skeleton computed with our method,  for Dataset F (part 1 of 2).}
\end{figure}

\clearpage
\begin{figure}
\subfloat[PINK filter3d]  
	{  
	\includegraphics[trim = 0cm 0cm 0cm 0cm,clip, width=\factorTreeB\linewidth]{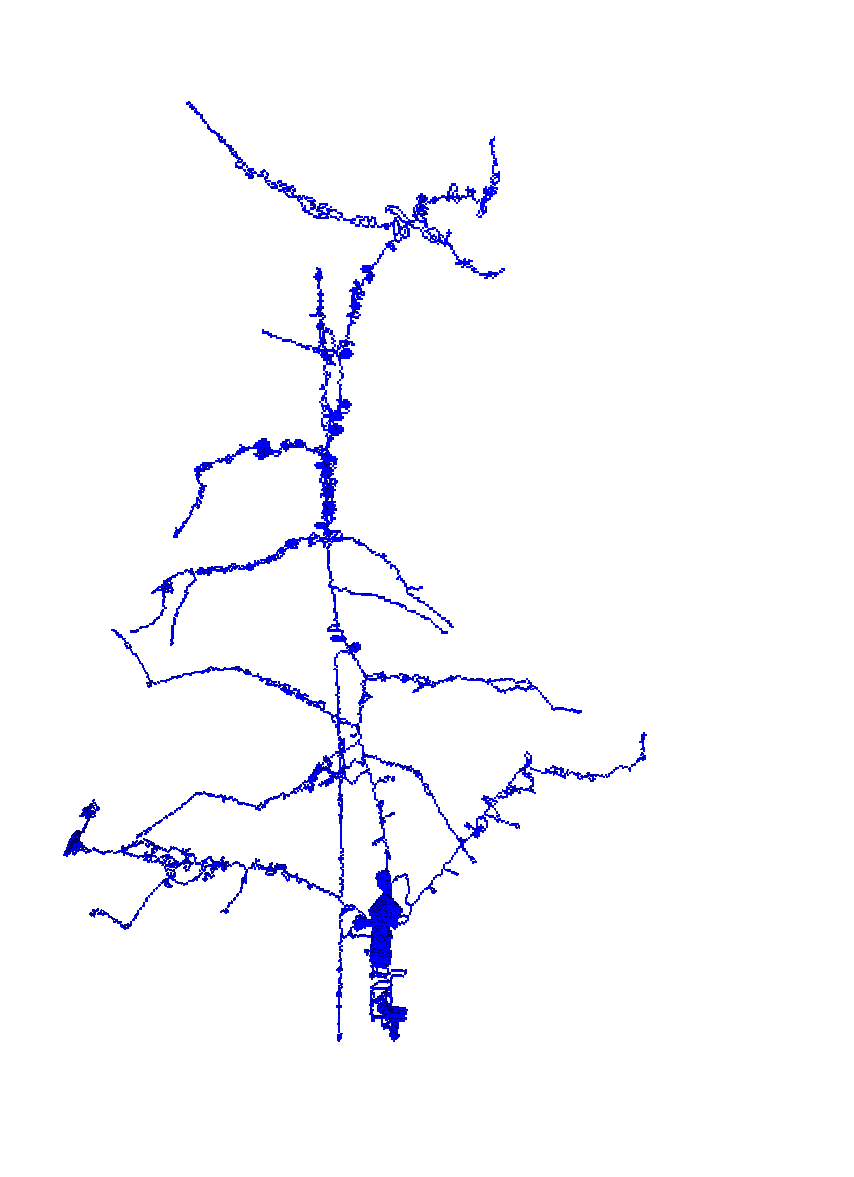}
}
\subfloat[Jin \etal]  
	{  
	\includegraphics[trim = 0cm 0cm 0cm 0cm,clip, width=\factorTreeB\linewidth]{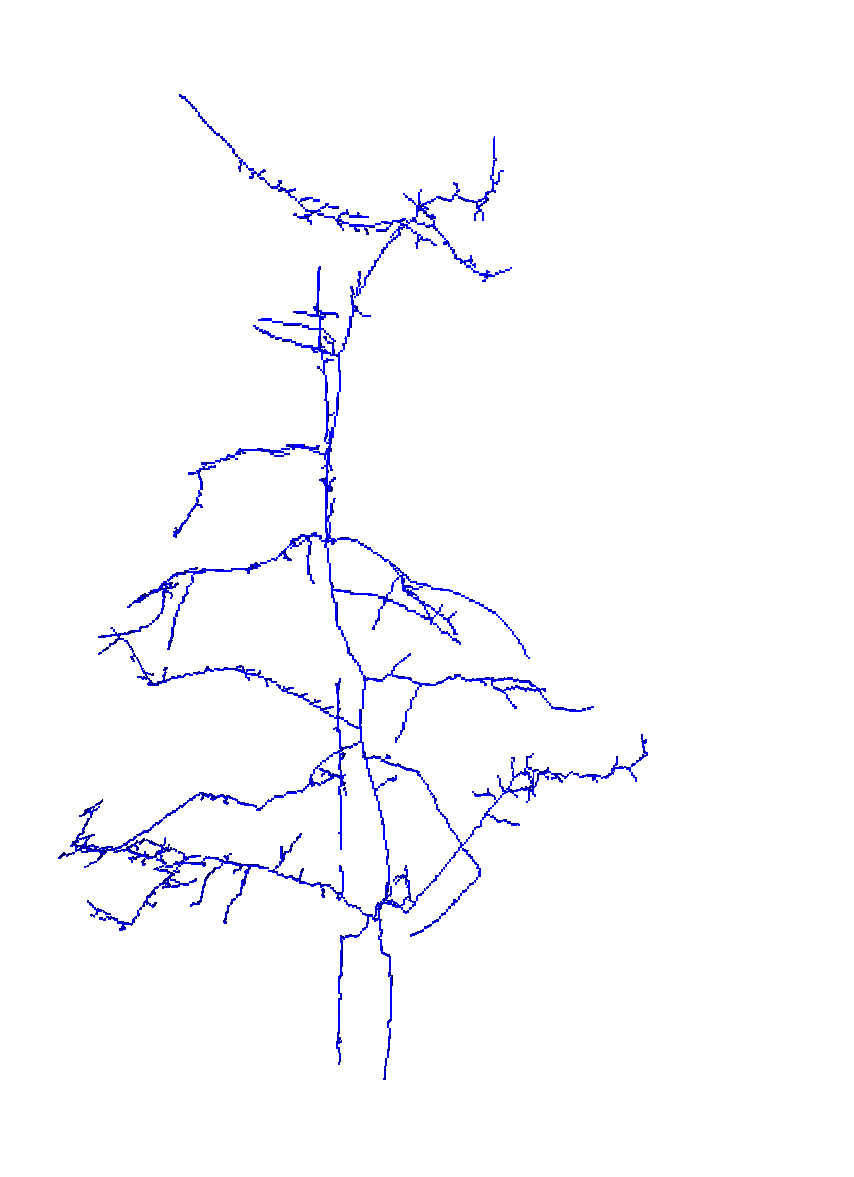}
}
\subfloat[Our method]  
	{  
	\includegraphics[trim = 0cm 0cm 0cm 0cm,clip, width=\factorTreeB\linewidth]{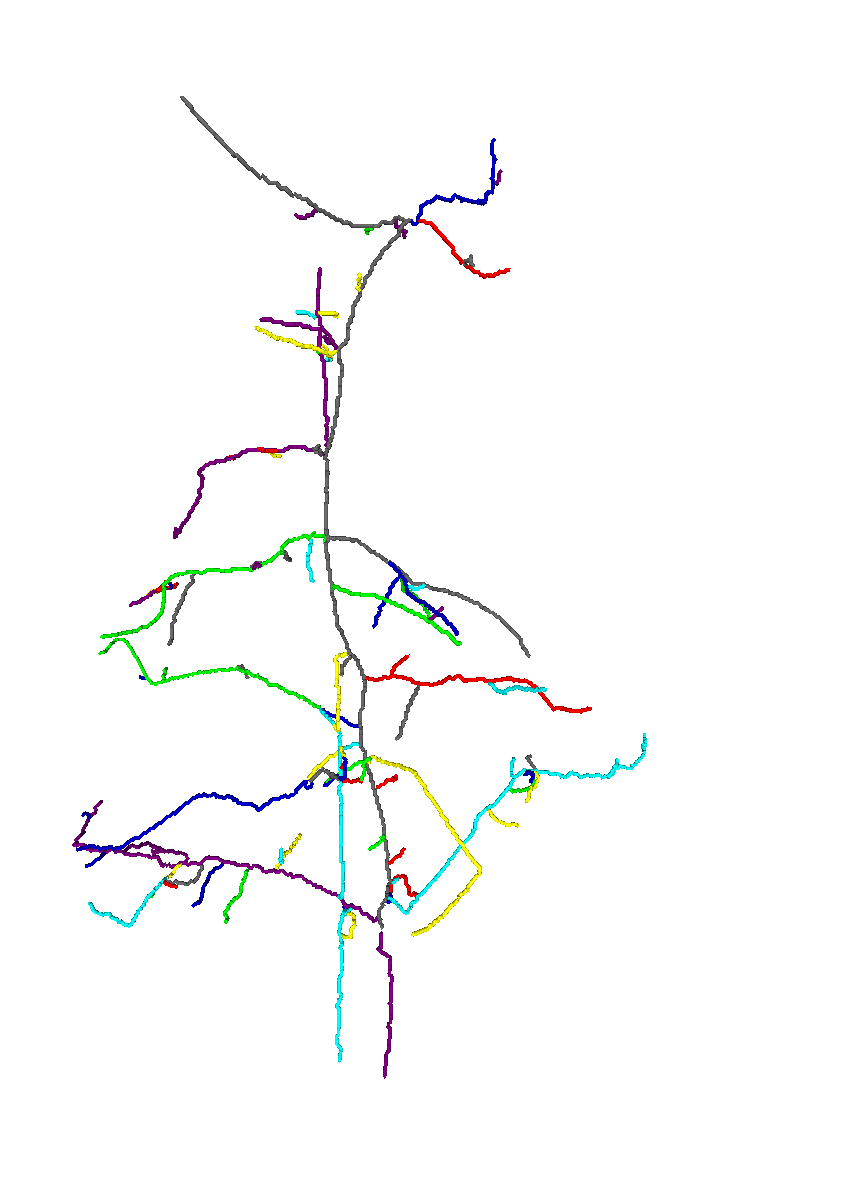}
}
\caption{\textbf{Best viewed in color.} Original surface, comparison curve skeletons, and curve skeleton computed with our method,  for Dataset F (part 2 of 2).}
\label{fig:resultsVI}
\end{figure}
\clearpage

%%%%%%%%%%%%%%%%%%%%%%%  G %%%%%%%%%%%%%%%%%%%%%%%%%
\begin{figure}
\centering
\subfloat[Surface]  
	{  
	\includegraphics[trim = 0cm 0cm 0cm 0cm,clip, width=\factorTreeB\linewidth]{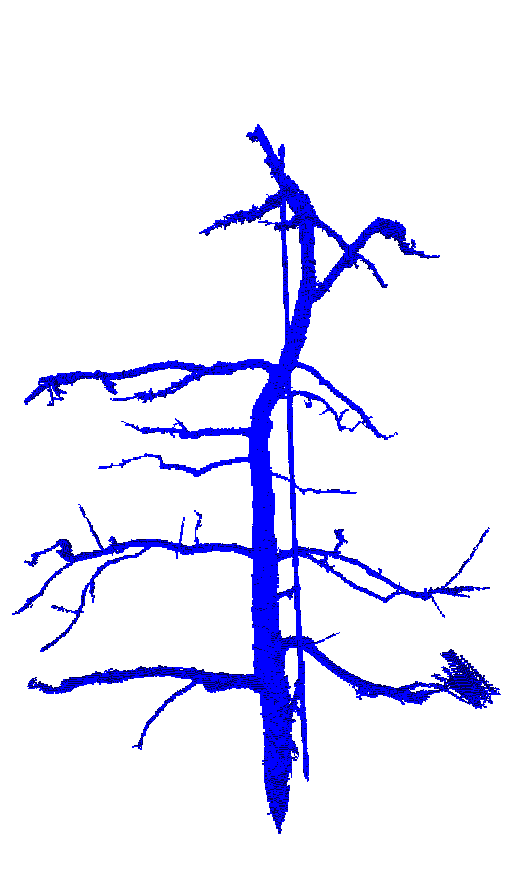}
}
\subfloat[Thinning]  
	{  
	\includegraphics[trim =0cm 0cm 0cm 0cm,clip, width=\factorTreeB\linewidth]{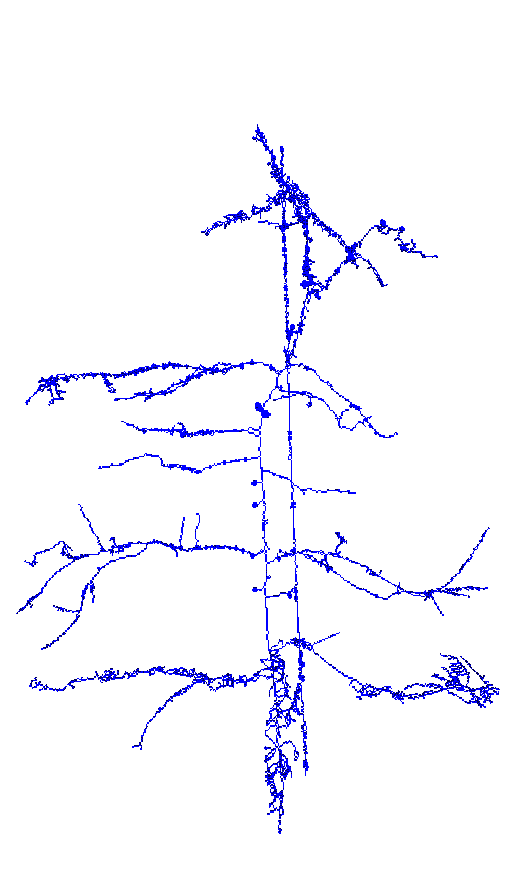}
}
\subfloat[PINK skel]  
	{  
	\includegraphics[trim = 0cm 0cm 0cm 0cm,clip, width=\factorTreeB\linewidth]{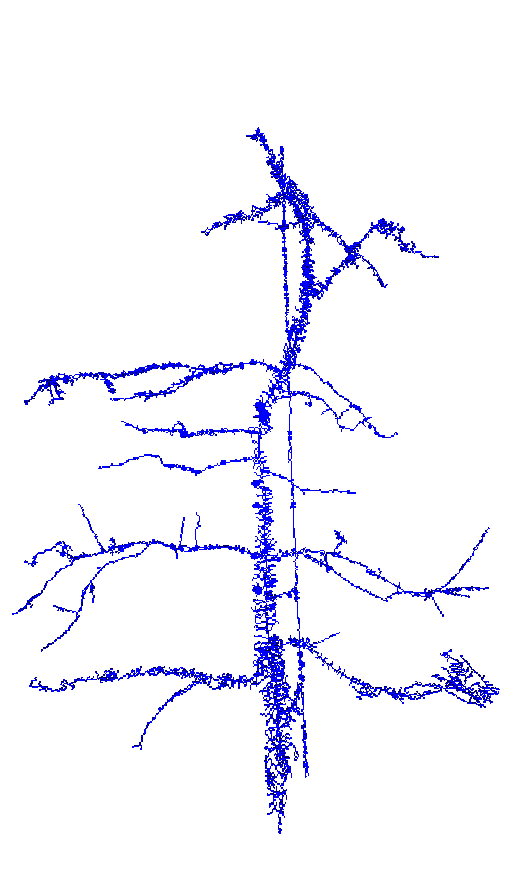}
}
\caption{\textbf{Best viewed in color.} Original surface, comparison curve skeletons, and curve skeleton computed with our method,  for Dataset G (part 1 of 2).}
\end{figure}

\clearpage
\begin{figure}
\subfloat[PINK filter3d]  
	{  
	\includegraphics[trim = 0cm 0cm 0cm 0cm,clip, width=\factorTreeB\linewidth]{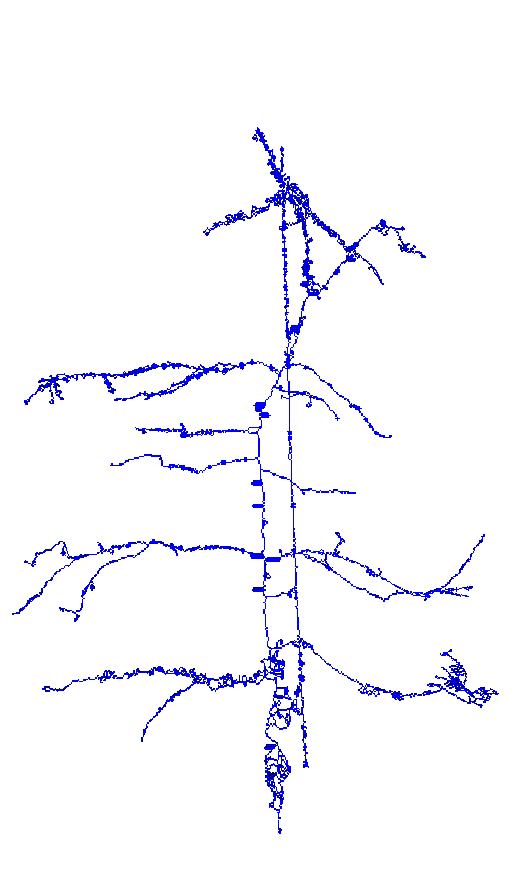}
}
\subfloat[Jin \etal]  
	{  
	\includegraphics[trim = 0cm 0cm 0cm 0cm,clip, width=\factorTreeB\linewidth]{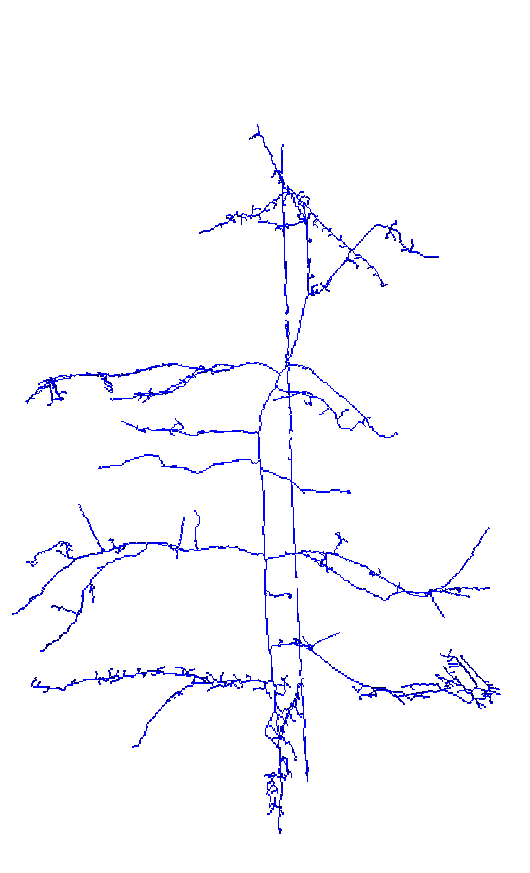}
}
\subfloat[Our method]  
	{  
	\includegraphics[trim = 0cm 0cm 0cm 0cm,clip, width=\factorTreeB\linewidth]{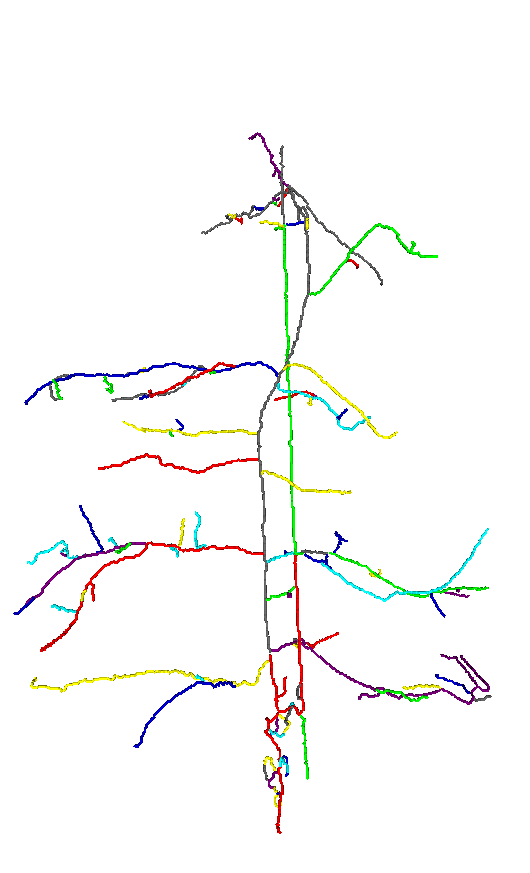}
}
\caption{\textbf{Best viewed in color.} Original surface, comparison curve skeletons, and curve skeleton computed with our method,  for Dataset G (part 2 of 2).}
\label{fig:resultsVII}
\end{figure}

\clearpage
%%%%%%%%%%%%%%%%%%%%%%%  H %%%%%%%%%%%%%%%%%%%%%%%%%
\begin{figure}
\centering
\subfloat[Surface]  
	{  
	\includegraphics[trim = 0cm 0cm 0cm 0cm,clip, width=\factorTreeB\linewidth]{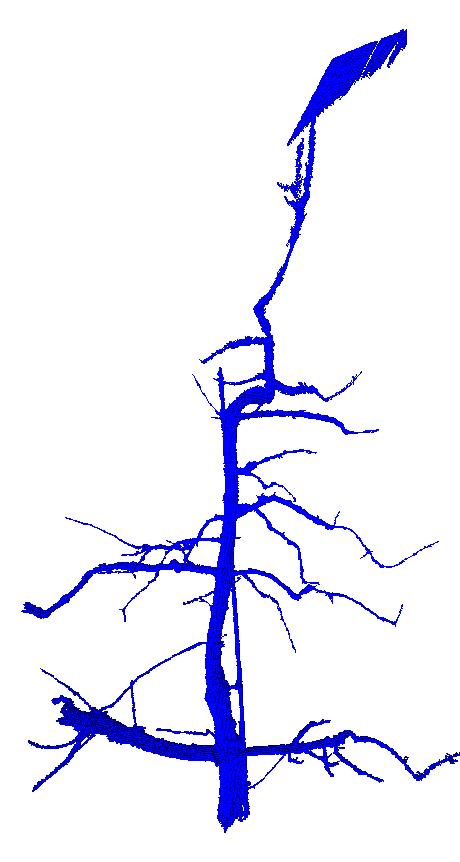}
}
\subfloat[Thinning]  
	{  
	\includegraphics[trim =0cm 0cm 0cm 0cm,clip, width=\factorTreeB\linewidth]{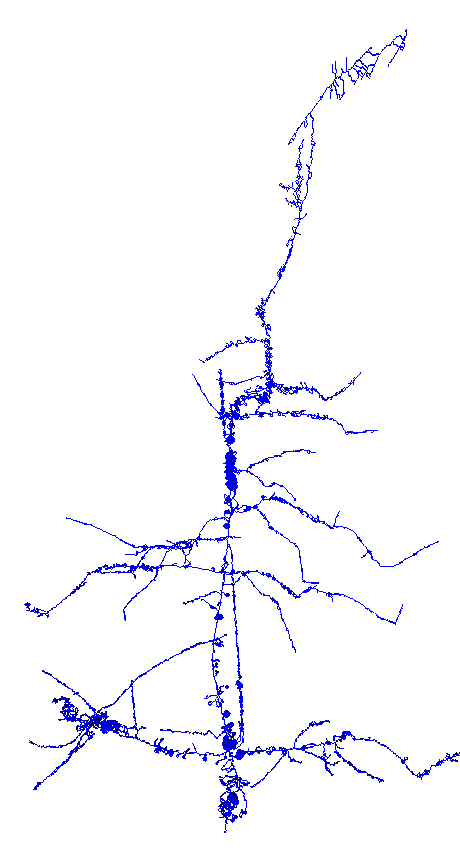}
}
\subfloat[PINK skel]  
	{  
	\includegraphics[trim = 0cm 0cm 0cm 0cm,clip, width=\factorTreeB\linewidth]{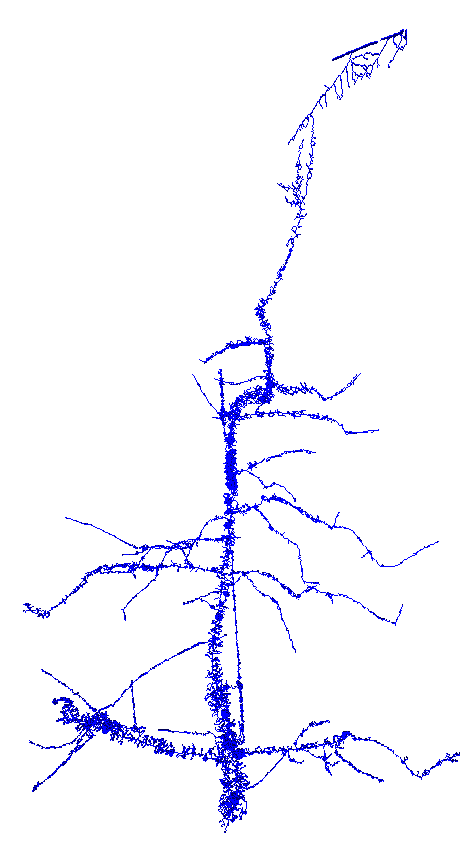}
}
\caption{\textbf{Best viewed in color.} Original surface, comparison curve skeletons, and curve skeleton computed with our method,  for Dataset H (part 1 of 2).}
\end{figure}

\clearpage
\begin{figure}
\subfloat[PINK filter3d]  
	{  
	\includegraphics[trim = 0cm 0cm 0cm 0cm,clip, width=\factorTreeB\linewidth]{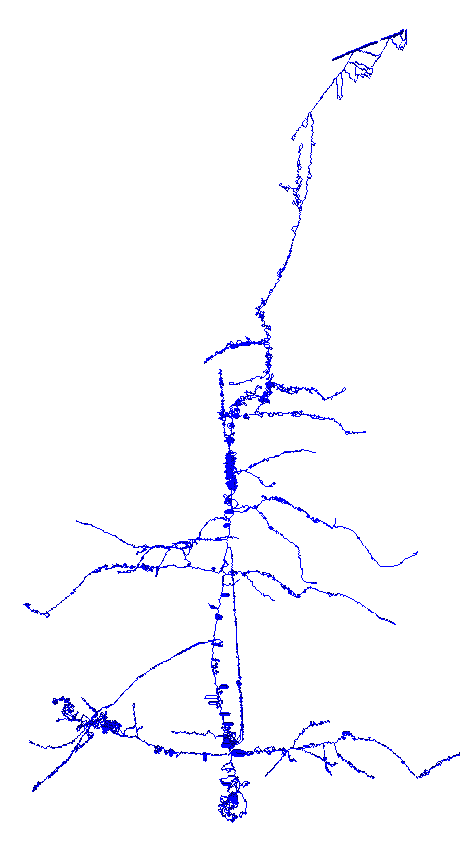}
}
\subfloat[Jin \etal]  
	{  
	\includegraphics[trim = 0cm 0cm 0cm 0cm,clip, width=\factorTreeB\linewidth]{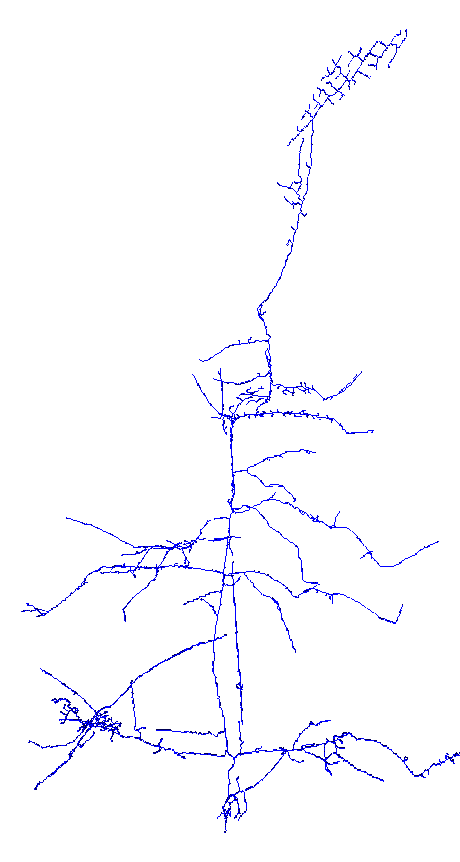}
}
\subfloat[Our method]  
	{  
	\includegraphics[trim = 0cm 0cm 0cm 0cm,clip, width=\factorTreeB\linewidth]{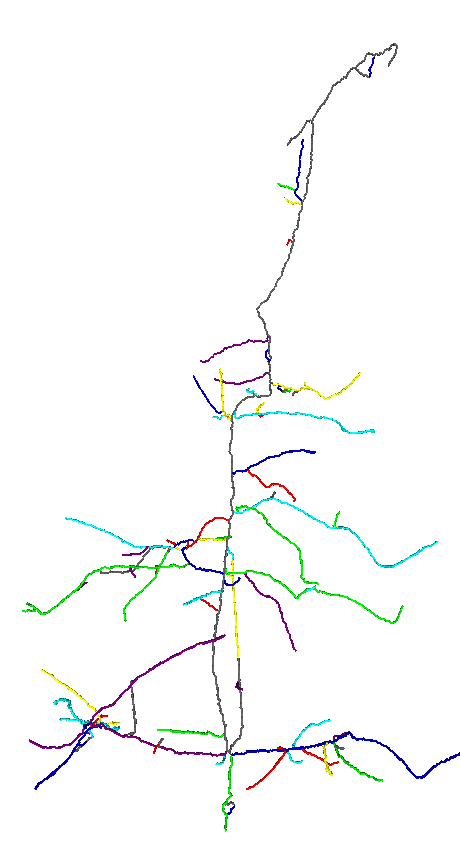}
}
\caption{\textbf{Best viewed in color.} Original surface, comparison curve skeletons, and curve skeleton computed with our method,  for Dataset H (part 2 of 2).}
\label{fig:resultsVII}
\end{figure}

\clearpage
%%%%%%%%%%%%%%%%%%%%%%%  I %%%%%%%%%%%%%%%%%%%%%%%%%
\begin{figure}
\centering
\subfloat[Surface]  
	{  
	\includegraphics[trim = 0cm 0cm 0cm 0cm,clip, width=\factorTreeB\linewidth]{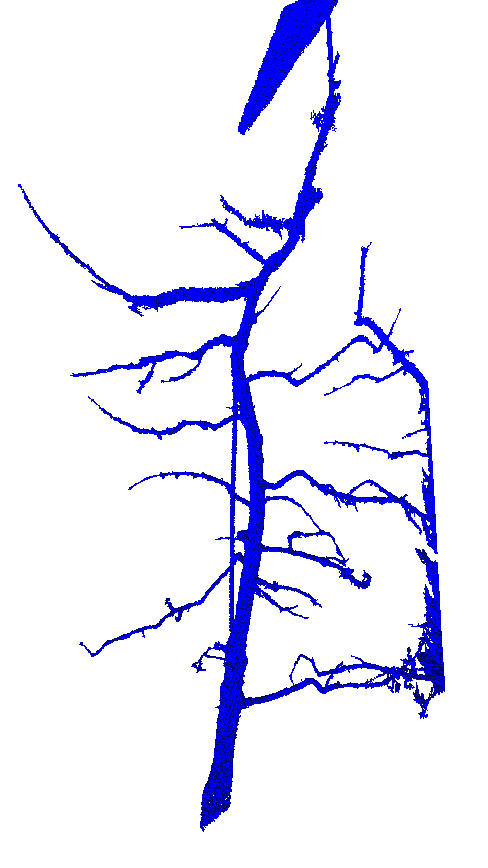}
}
\subfloat[Thinning]  
	{  
	\includegraphics[trim =0cm 0cm 0cm 0cm,clip, width=\factorTreeB\linewidth]{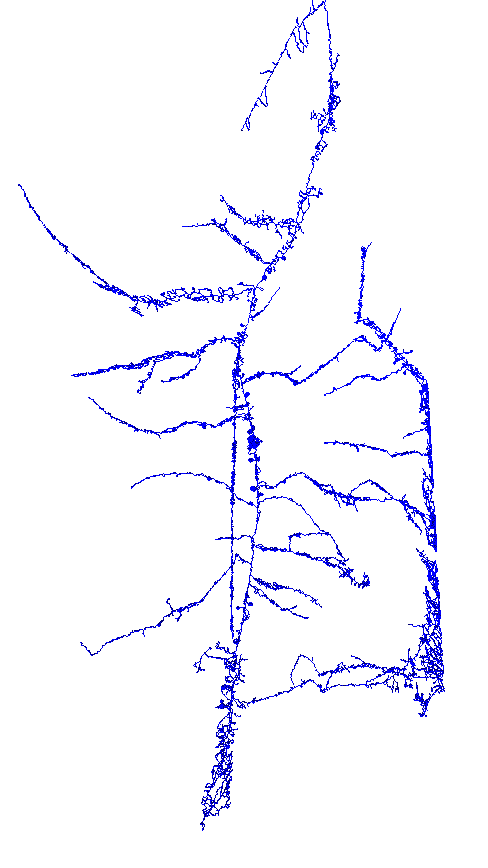}
}
\subfloat[PINK skel]  
	{  
	\includegraphics[trim = 0cm 0cm 0cm 0cm,clip, width=\factorTreeB\linewidth]{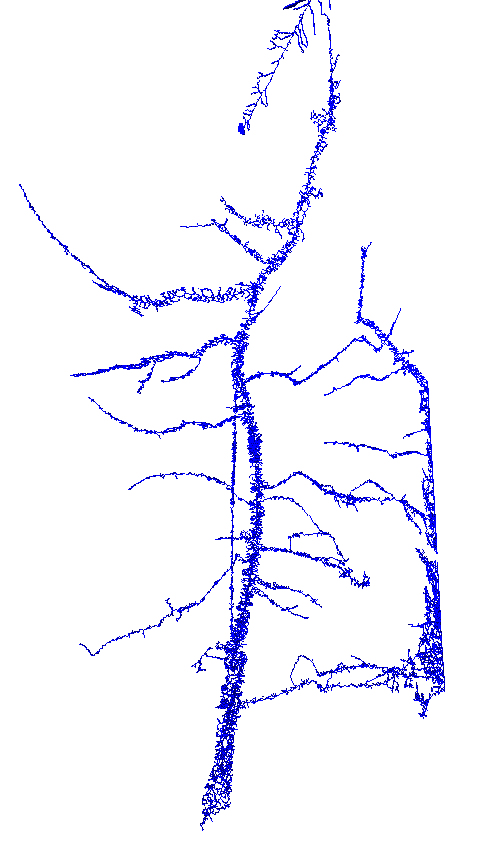}
}
\caption{\textbf{Best viewed in color.} Original surface, comparison curve skeletons, and curve skeleton computed with our method,  for Dataset I (part 1 of 2).}
\end{figure}

\clearpage
\begin{figure}
\subfloat[PINK filter3d]  
	{  
	\includegraphics[trim = 0cm 0cm 0cm 0cm,clip, width=\factorTreeB\linewidth]{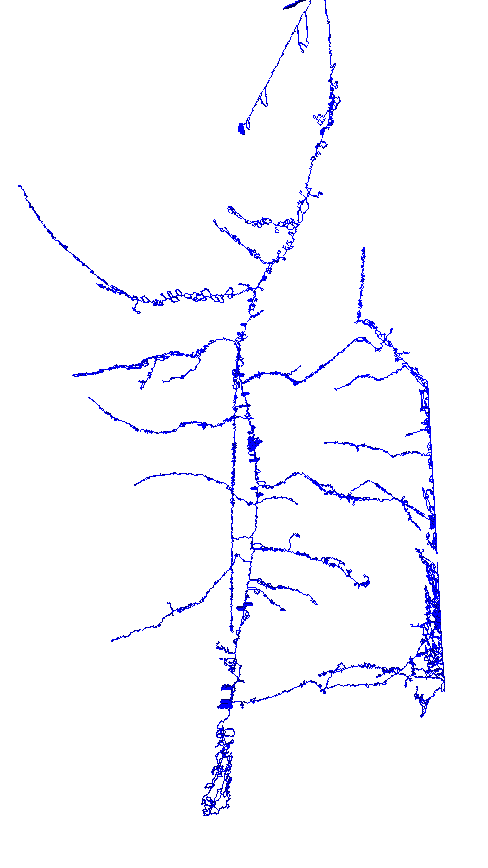}
}
\subfloat[Jin \etal]  
	{  
	\includegraphics[trim = 0cm 0cm 0cm 0cm,clip, width=\factorTreeB\linewidth]{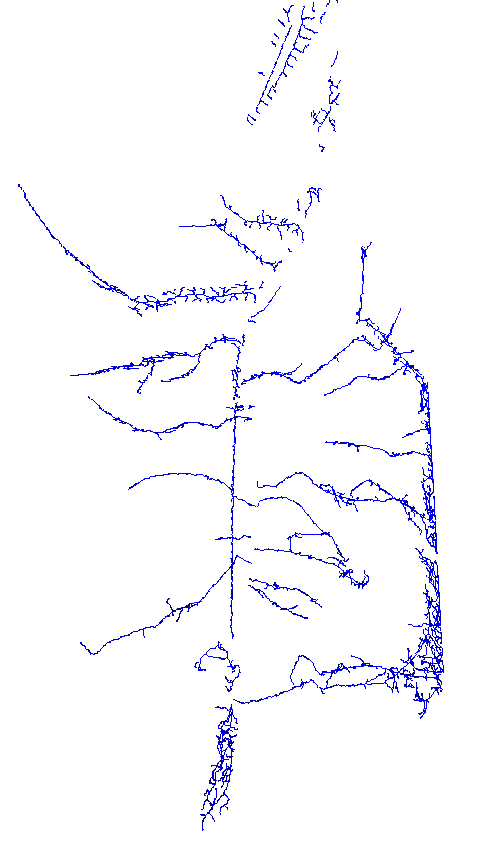}
}
\subfloat[Our method]  
	{  
	\includegraphics[trim = 0cm 0cm 0cm 0cm,clip, width=\factorTreeB\linewidth]{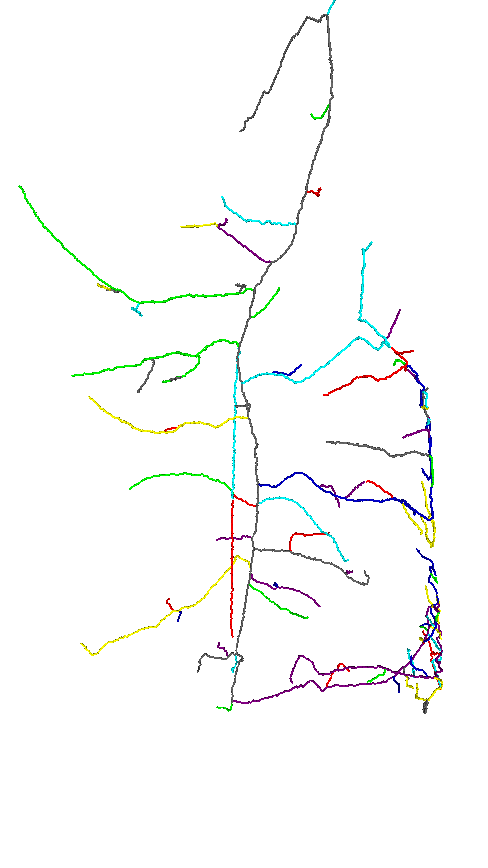}
}
\caption{\textbf{Best viewed in color.} Original surface, comparison curve skeletons, and curve skeleton computed with our method,  for Dataset I (part 2 of 2).}\label{fig:resultsI}
\label{fig:resultsVIII}
\end{figure}

\end{landscape}

\subsection{Computer graphics models}
The curve skeleton algorithm is demonstrated on commonly-used computer graphics models in Figures \ref{fig:model_first}-\ref{fig:model_first}. Surfaces shown in \ref{sf:octopus} to \ref{sf:hand} provided courtesy of INRIA, owner of \ref{sf:seahorse} unknown, all via AIM@SHAPE-VISIONAIR Shape Repository \cite{AIM_Shape}. All models were converted from mesh to voxels using the algorithm of \cite{Nooruddin2003Visualization} as implemented in \cite{binvox}. 

%\begin{landscape}
\renewcommand*{\factorSynthetics}{0.45}
\begin{figure*}[ht!]
\centering
\subfloat[Surface]  
	{  
	\includegraphics[trim = 6cm 2cm 6cm 6cm,clip, width=\factorSynthetics\linewidth]{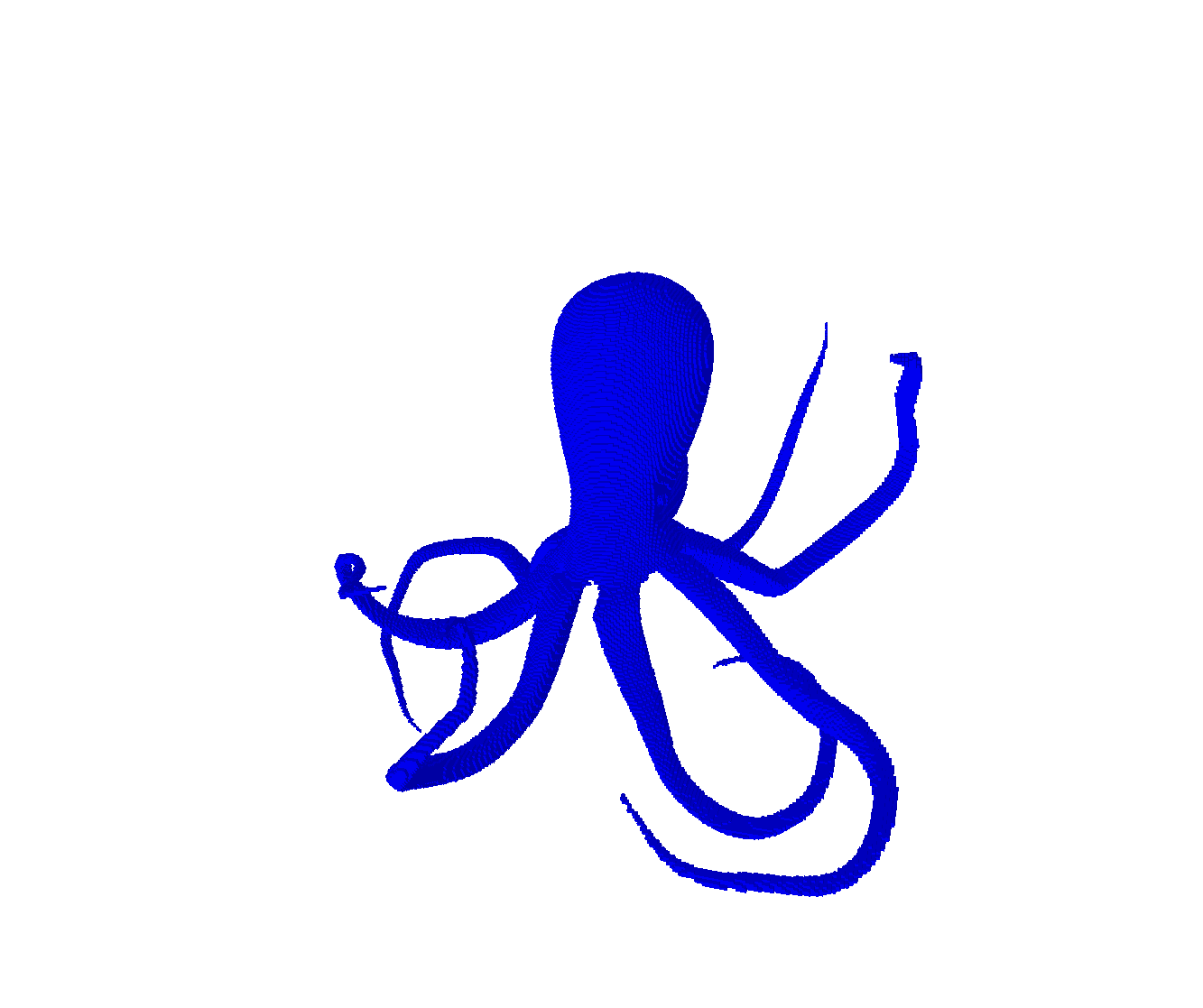}
	\label{sf:octopus}
}
\subfloat[Skeleton]  
	{  
	\includegraphics[trim = 6cm 2cm 6cm 6cm,clip, width=\factorSynthetics\linewidth]{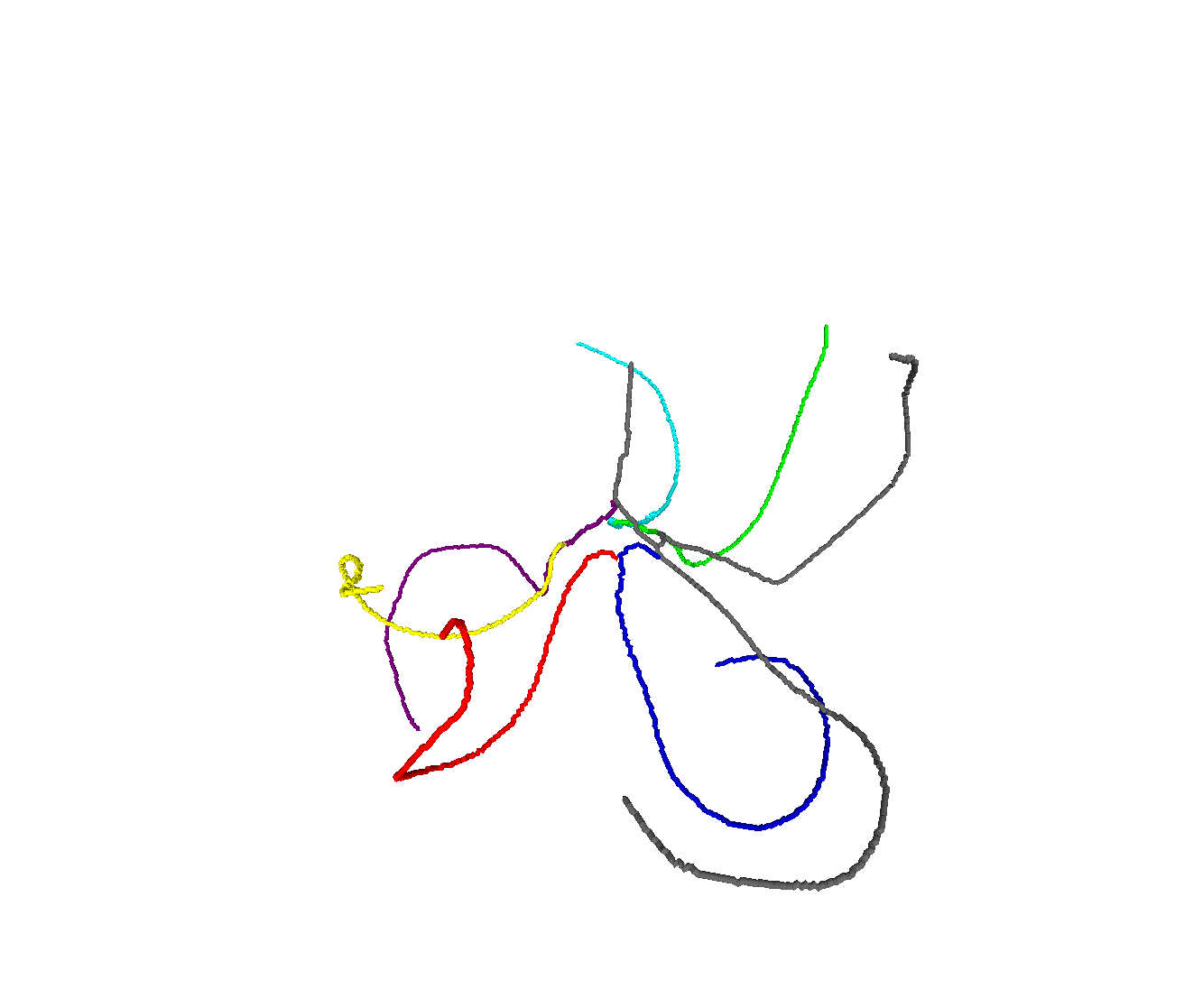}
}

\subfloat[Surface]  
	{  
	\includegraphics[trim = 6cm 2cm 10cm 0cm,clip, width=\factorSynthetics\linewidth]{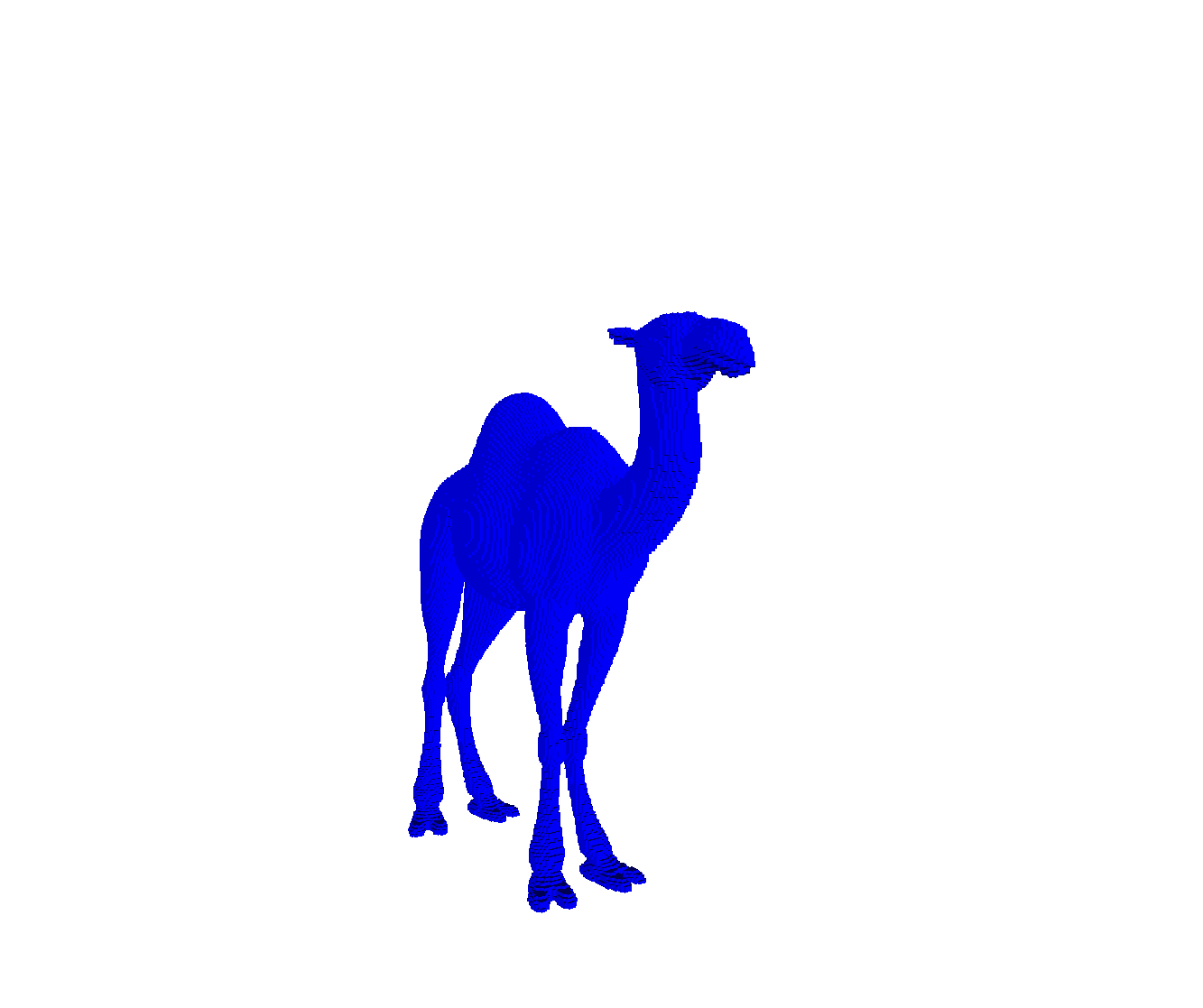}
	\label{sf:camel}
}
\subfloat[Skeleton]  
	{  
	\includegraphics[trim = 6cm 2cm 10cm 0cm,clip, width=\factorSynthetics\linewidth]{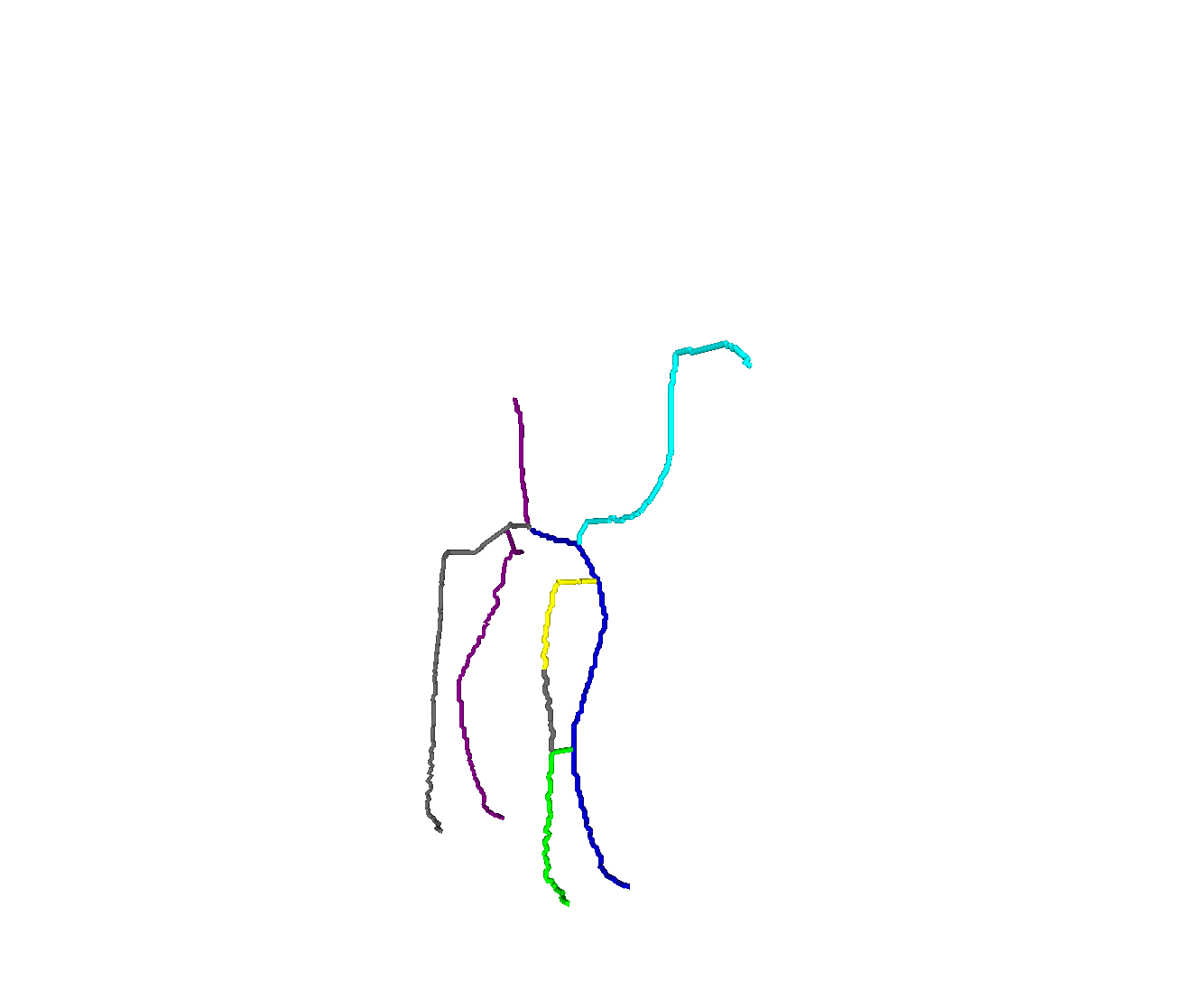}
	\label{sf:camels}
}
\caption{Original surfaces and skeletons computed with our method.}
\label{fig:model_first}
\end{figure*}

\begin{figure*}
\subfloat[Surface]  
	{  
	\includegraphics[trim = 6cm 7cm 9cm 6cm,clip, width=\factorSynthetics\linewidth]{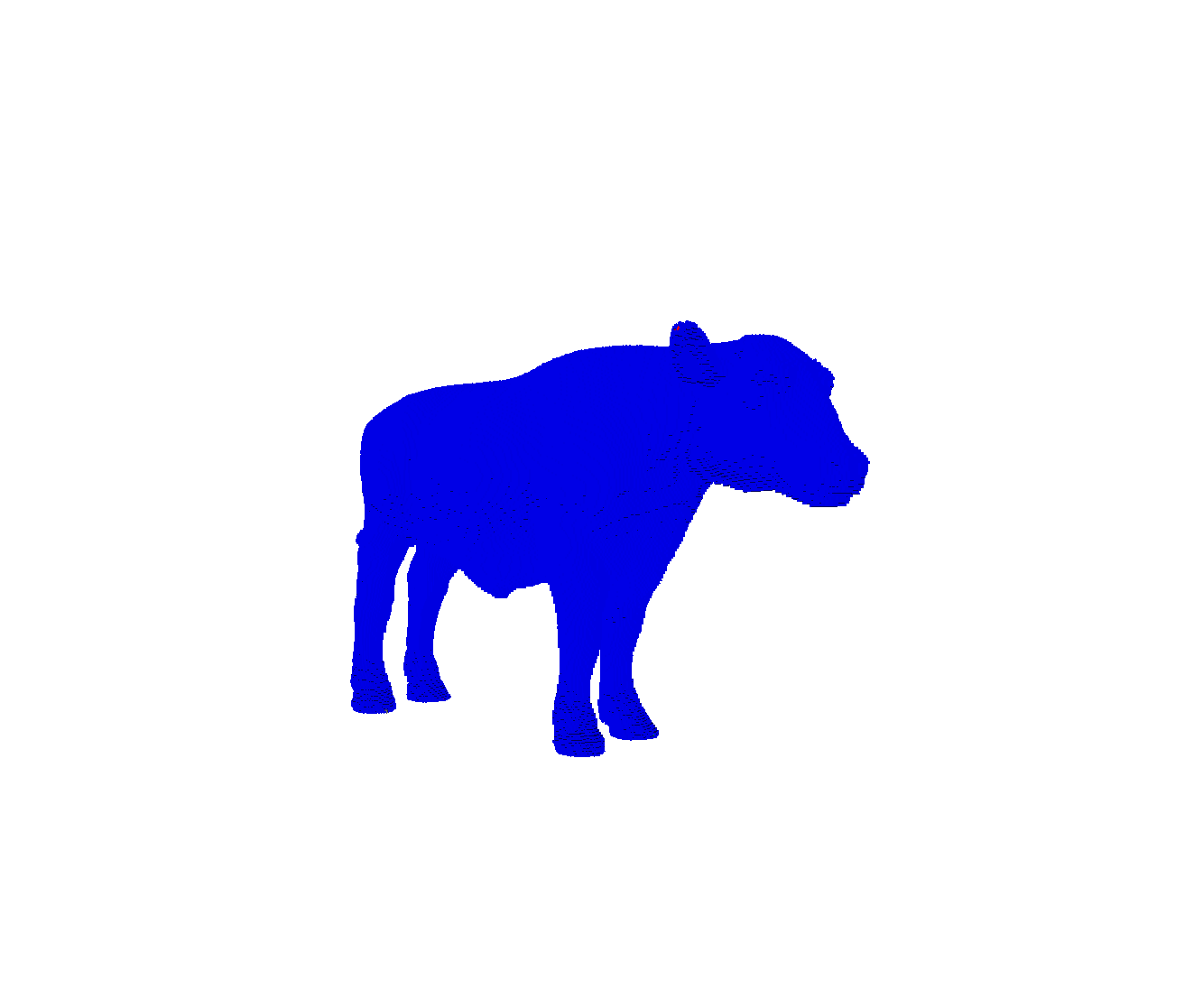}
\label{sf:cow}	
}
\subfloat[Skeleton]  
	{  
	\includegraphics[trim = 6cm 7cm 9cm 6cm,clip, width=\factorSynthetics\linewidth]{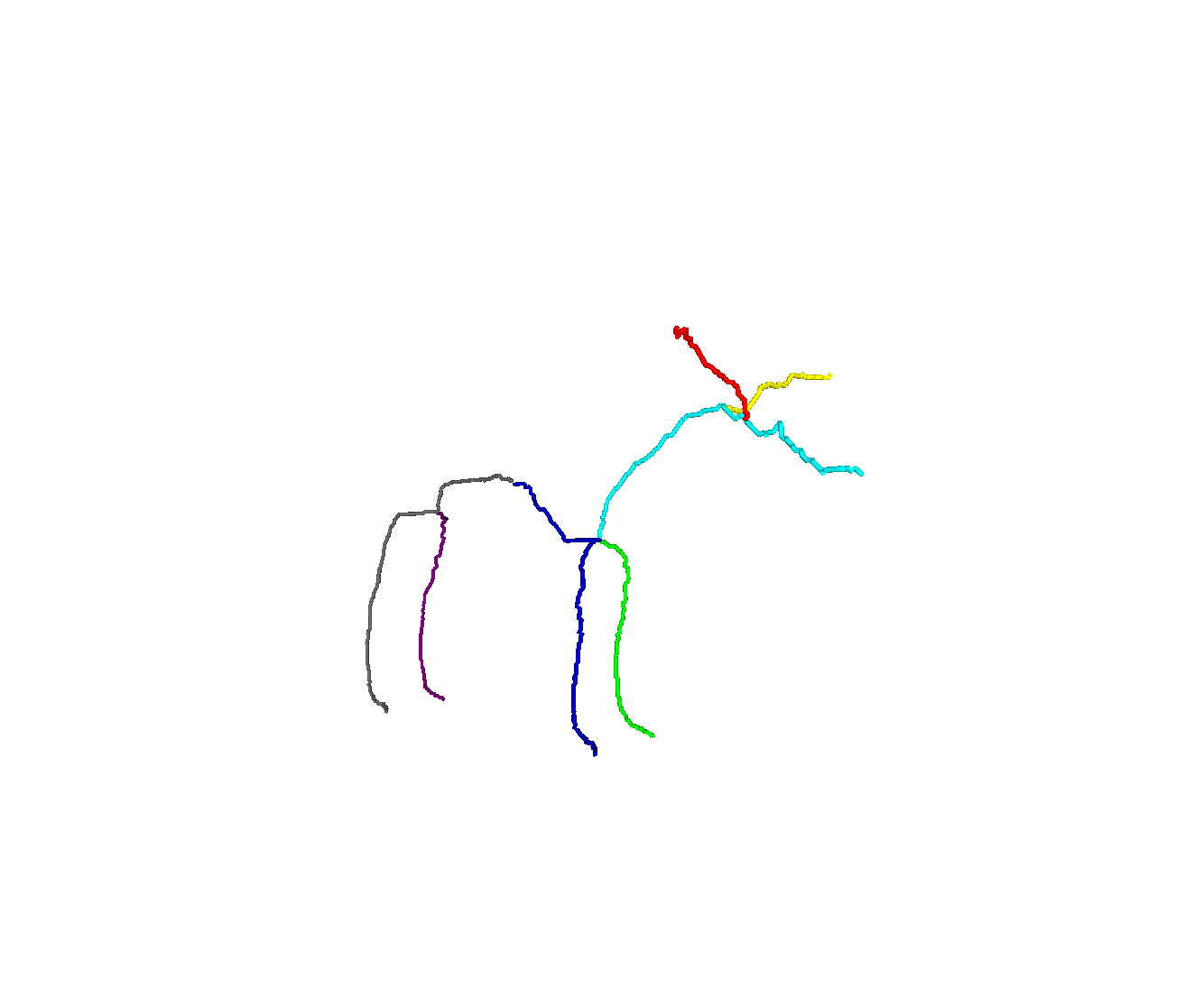}
}

\subfloat[Surface]  
	{  
	\includegraphics[trim = 5cm 4cm 6cm 2cm,clip, width=\factorSynthetics\linewidth]{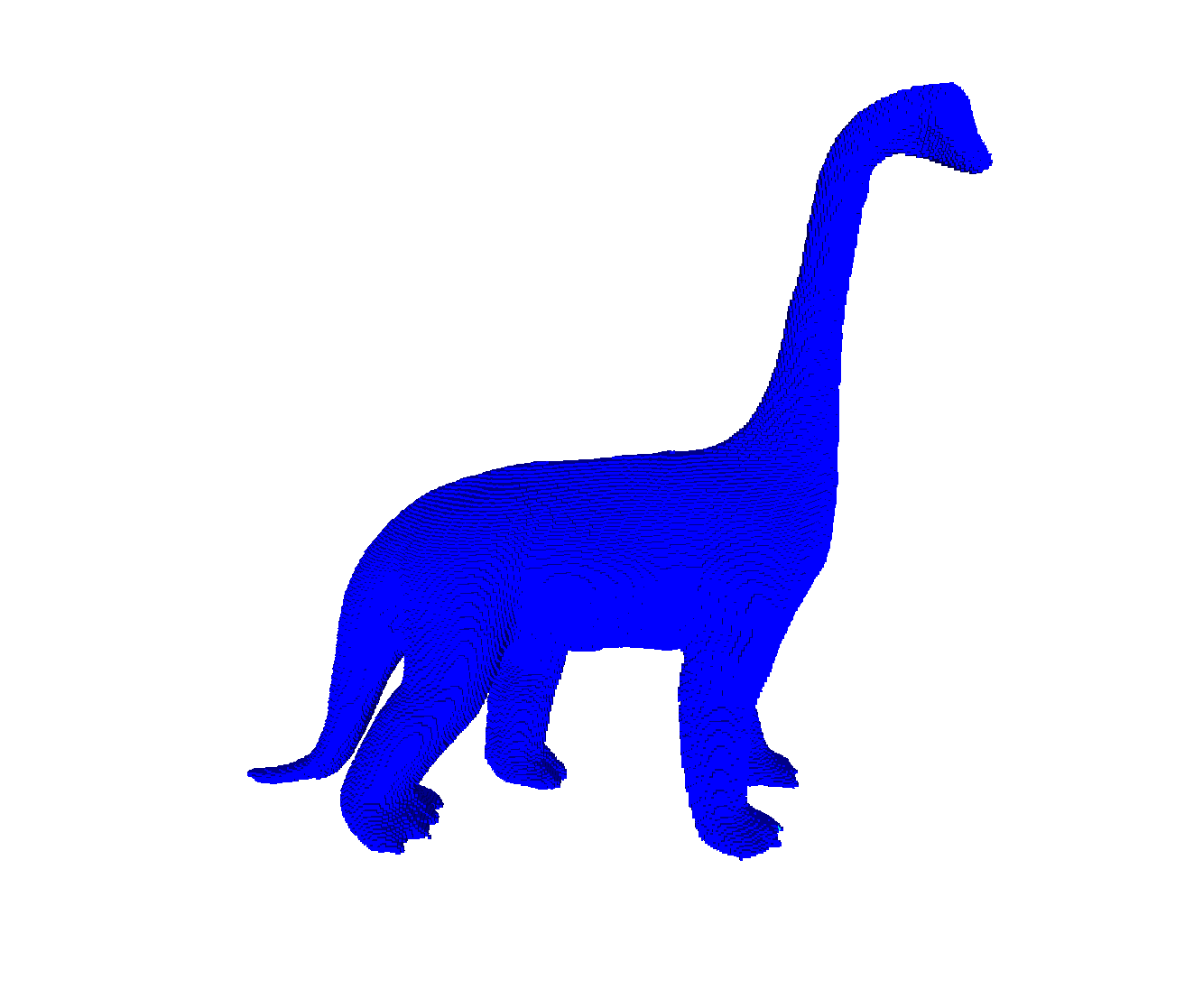}
	\label{sf:dinosaur}
}
\subfloat[Skeleton]  
	{  
	\includegraphics[trim = 5cm 4cm 6cm 2cm,clip, width=\factorSynthetics\linewidth]{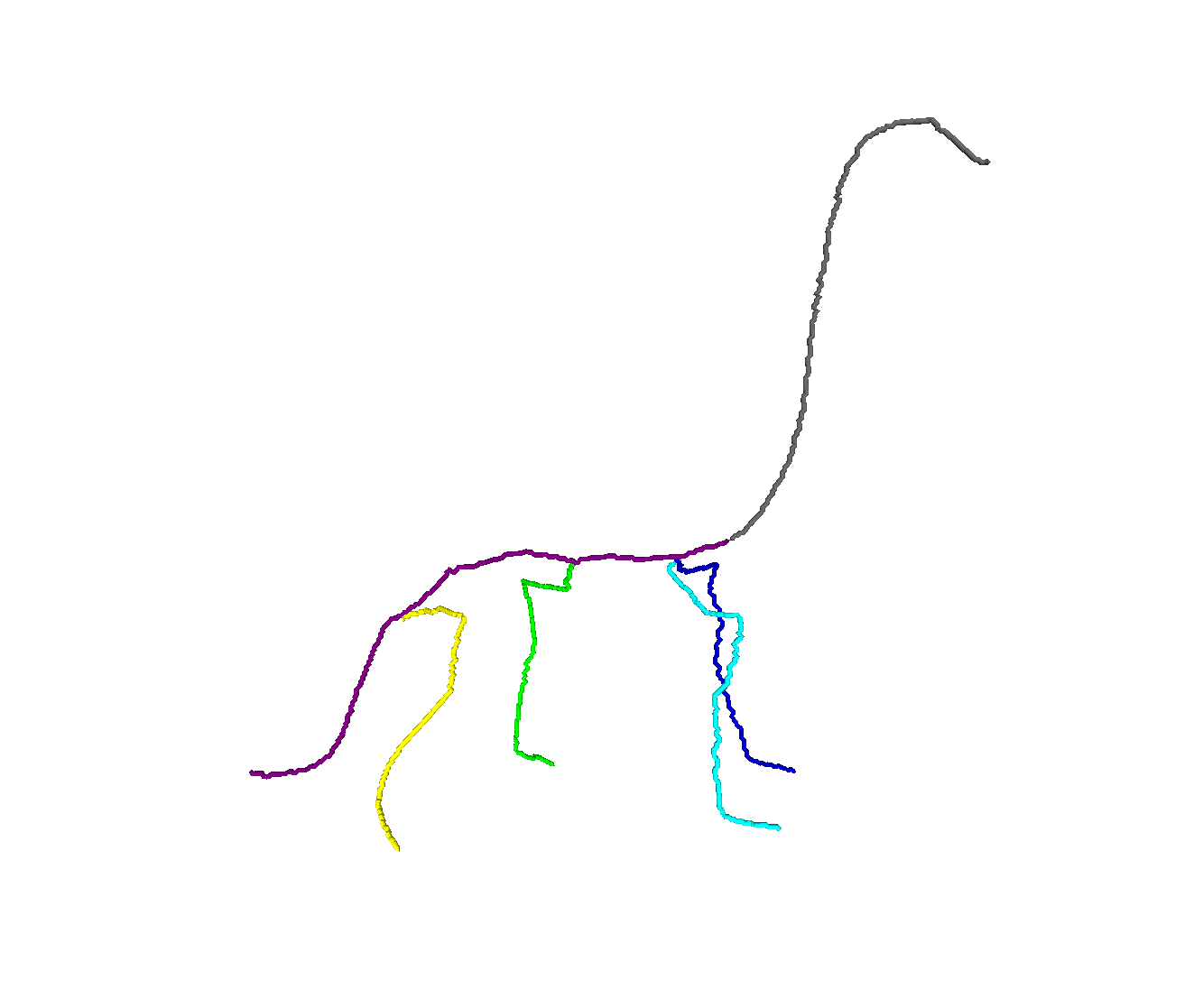}
}
\caption{Original surfaces and skeletons computed with our method.}
\label{fig:model_second}
\end{figure*}

\begin{figure*}
\subfloat[Surface]  
	{  
	\includegraphics[trim = 5cm 3cm 1cm 2cm,clip, width=\factorSynthetics\linewidth]{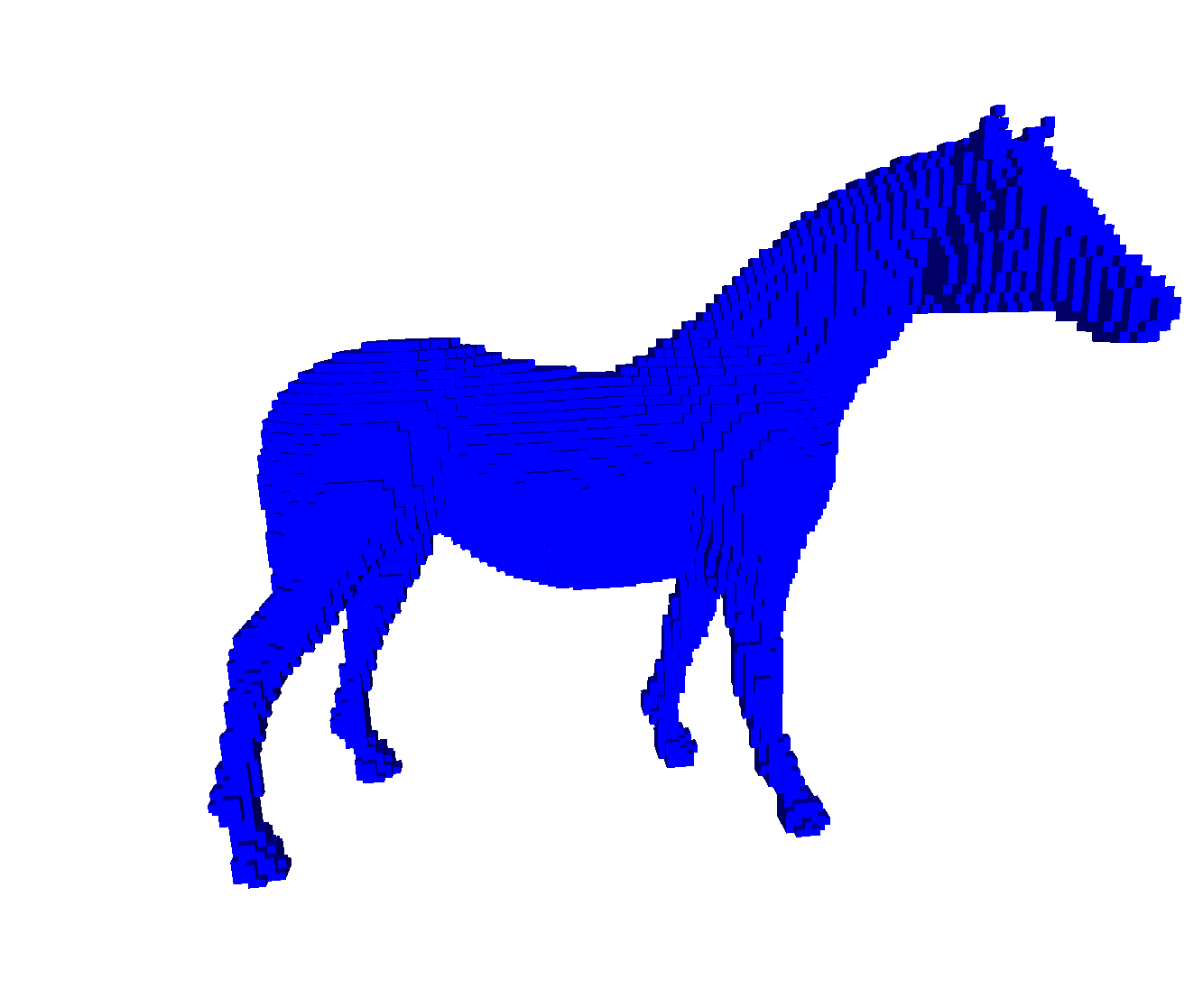}
	\label{sf:horse}
}
\subfloat[Skeleton]  
	{  
	\includegraphics[trim = 5cm 3cm 1cm 2cm,clip, width=\factorSynthetics\linewidth]{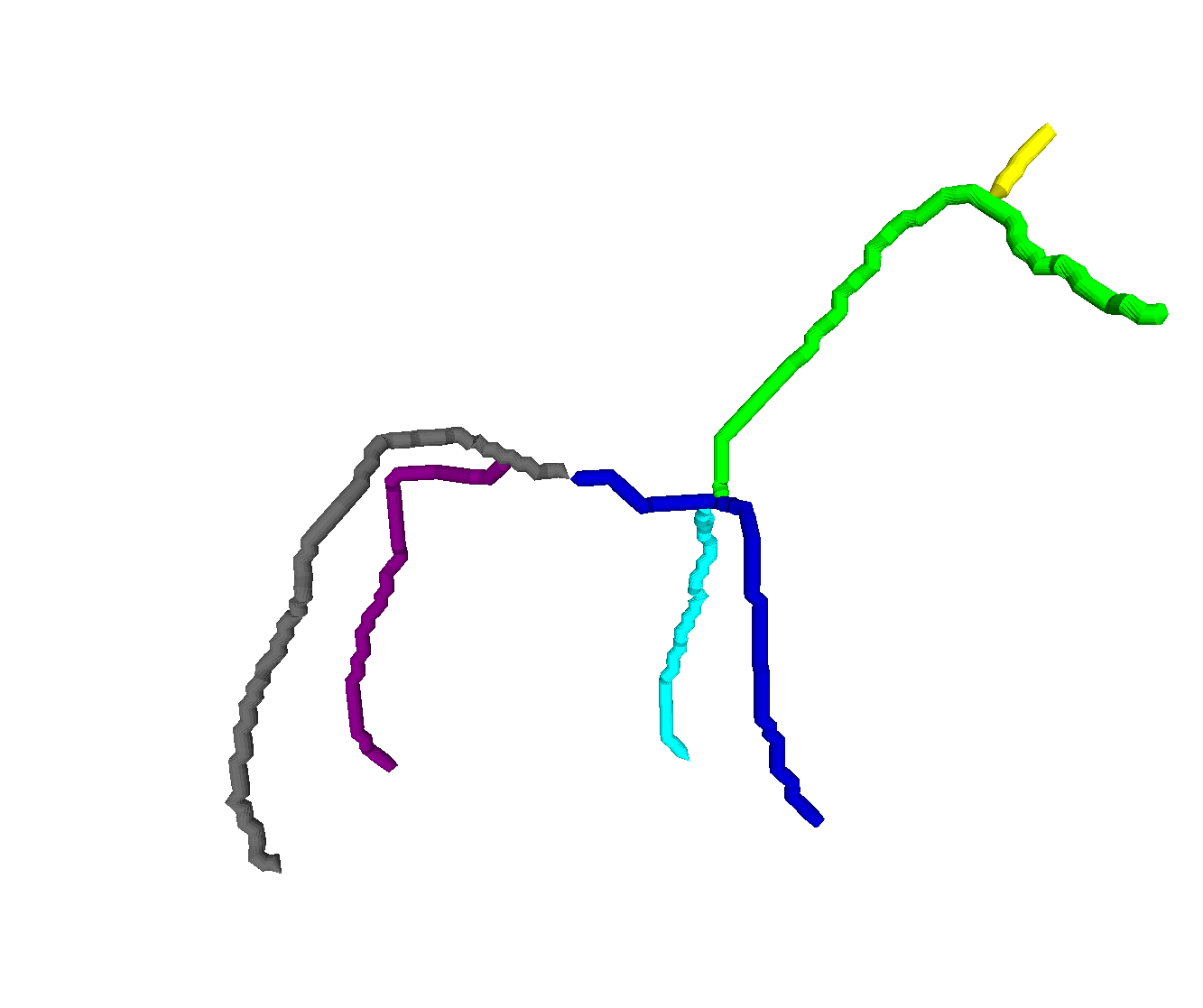}
}

\subfloat[Surface]  
	{  
	\includegraphics[trim = 5cm 3cm 0cm 2cm,clip, width=\factorSynthetics\linewidth]{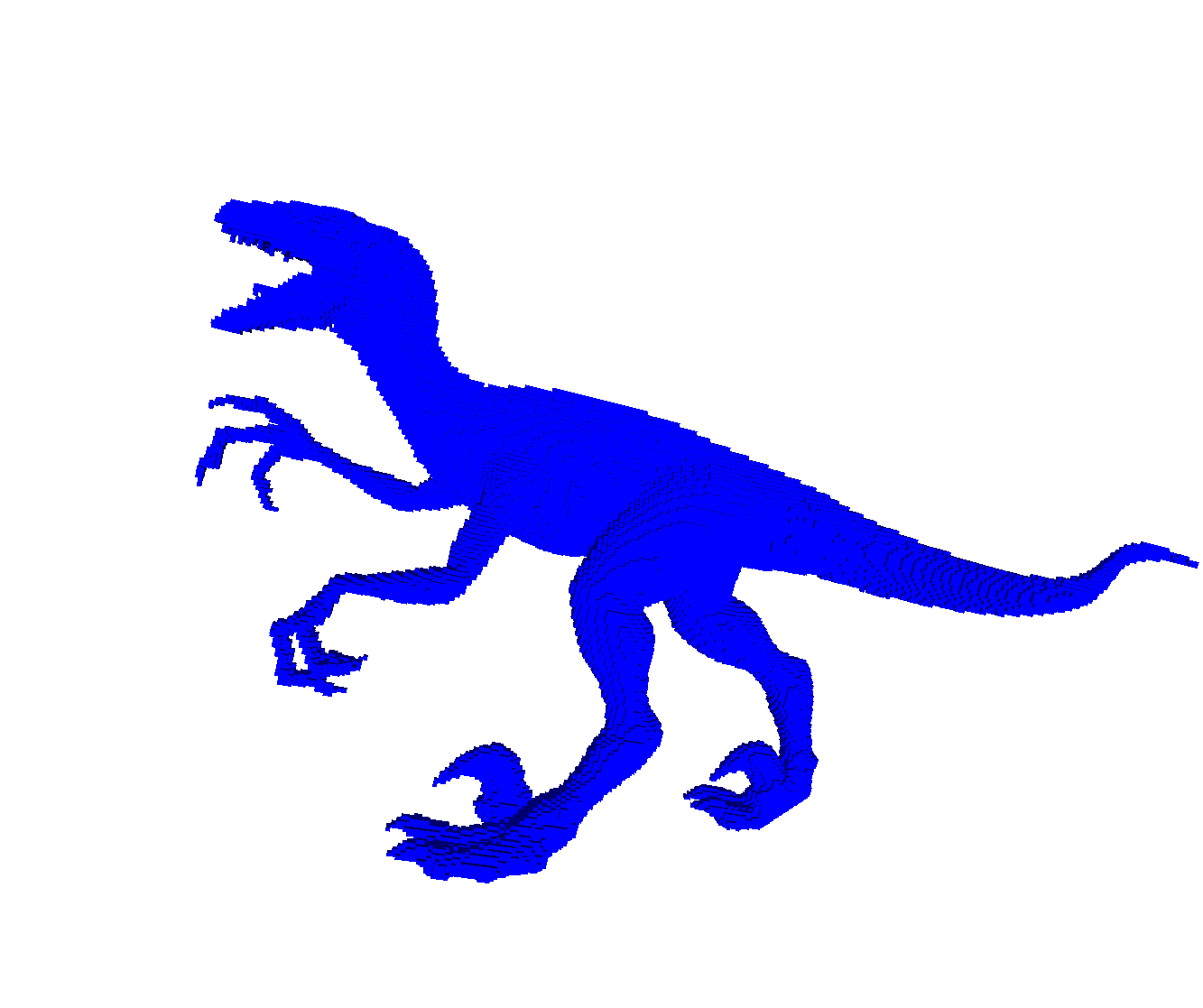}
	\label{sf:raptor}
}
\subfloat[Skeleton]  
	{  
	\includegraphics[trim = 5cm 3cm 0cm 2cm,clip, width=\factorSynthetics\linewidth]{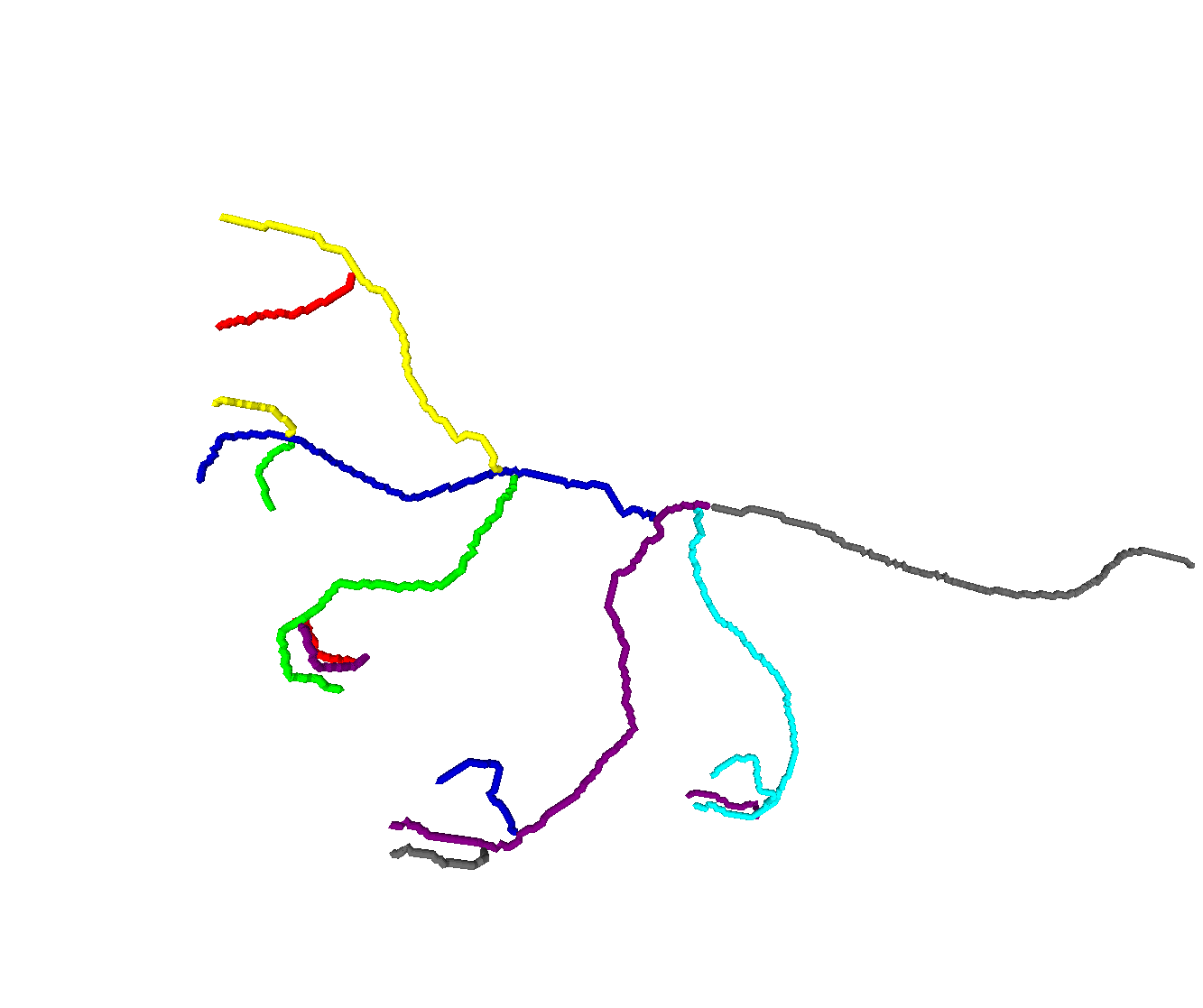}	
}
\caption{Original surfaces and skeletons computed with our method.}
\label{fig:model_third}
\end{figure*}

\begin{figure*}
\subfloat[Surface]  
	{  
	\includegraphics[trim = 5cm 0cm 4cm 2cm,clip, width=\factorSynthetics\linewidth]{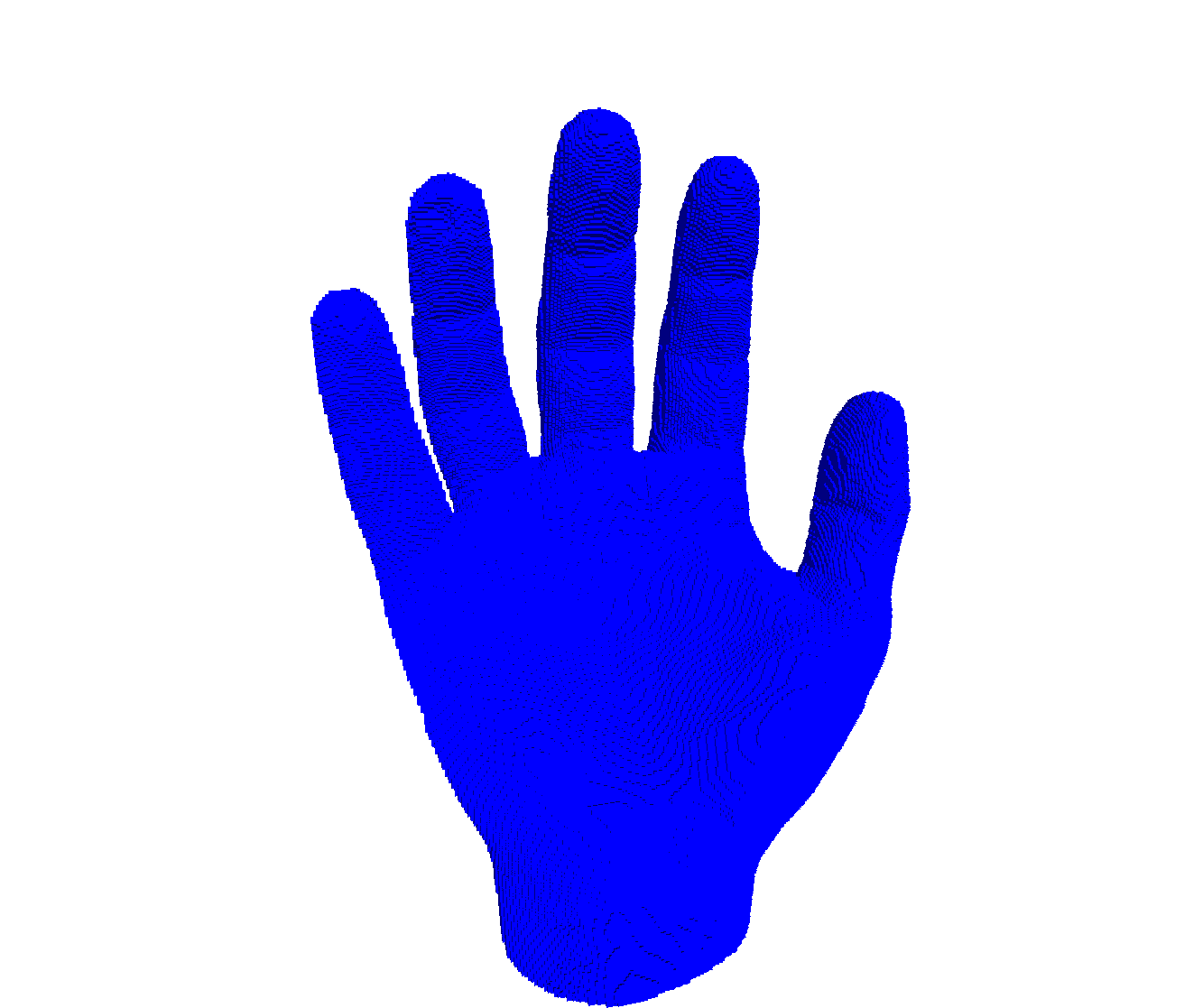}
\label{sf:hand}
}
\subfloat[Skeleton]  
	{  
	\includegraphics[trim = 5cm 0cm 4cm 2cm,clip, width=\factorSynthetics\linewidth]{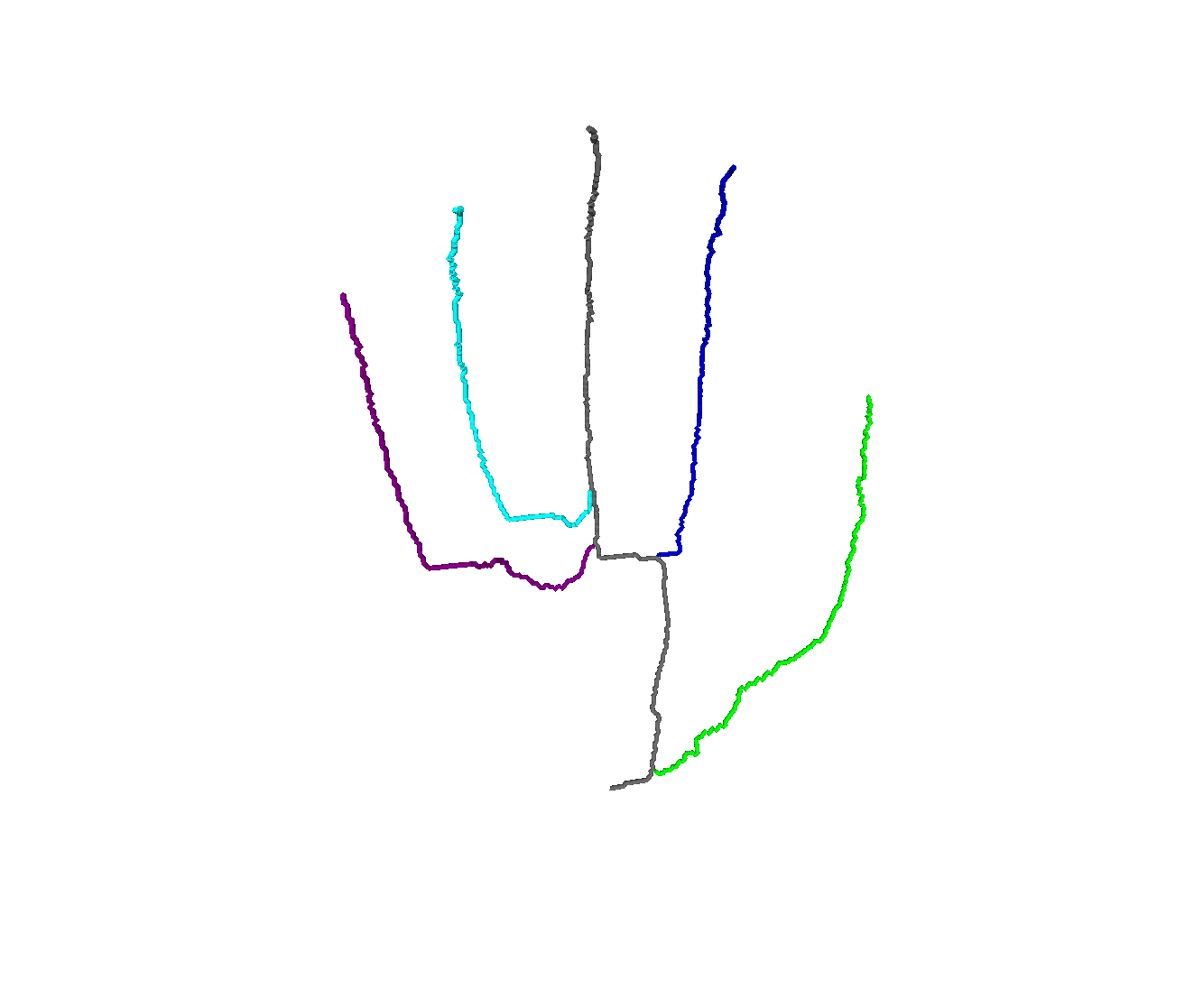}
}

\subfloat[Surface]  
	{  
	\includegraphics[trim = 5cm 0cm 5cm 0cm,clip, width=\factorSynthetics\linewidth]{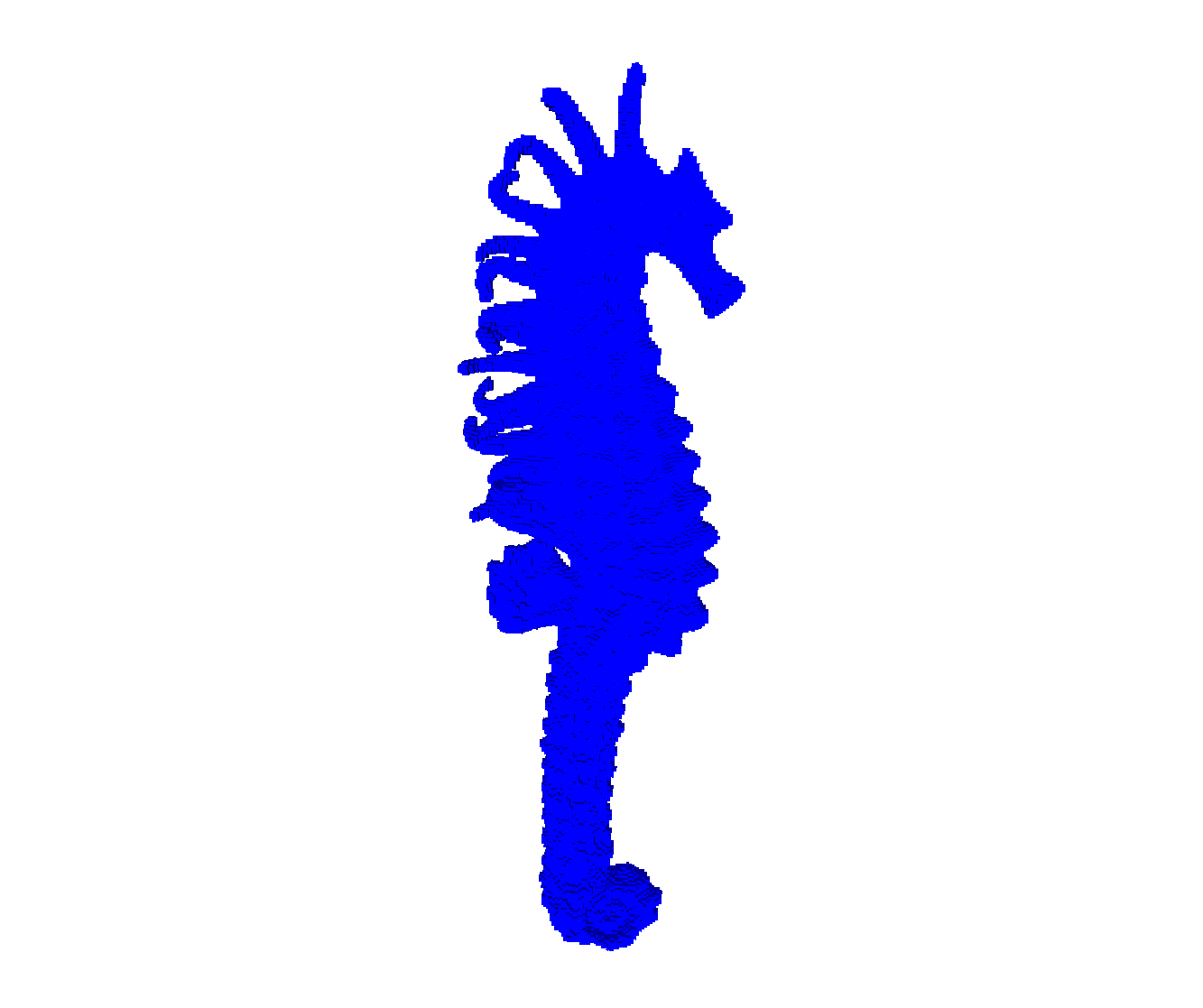}
\label{sf:seahorse}
}
\subfloat[Skeleton]  
	{  
	\includegraphics[trim = 5cm 0cm 5cm 0cm,clip, width=\factorSynthetics\linewidth]{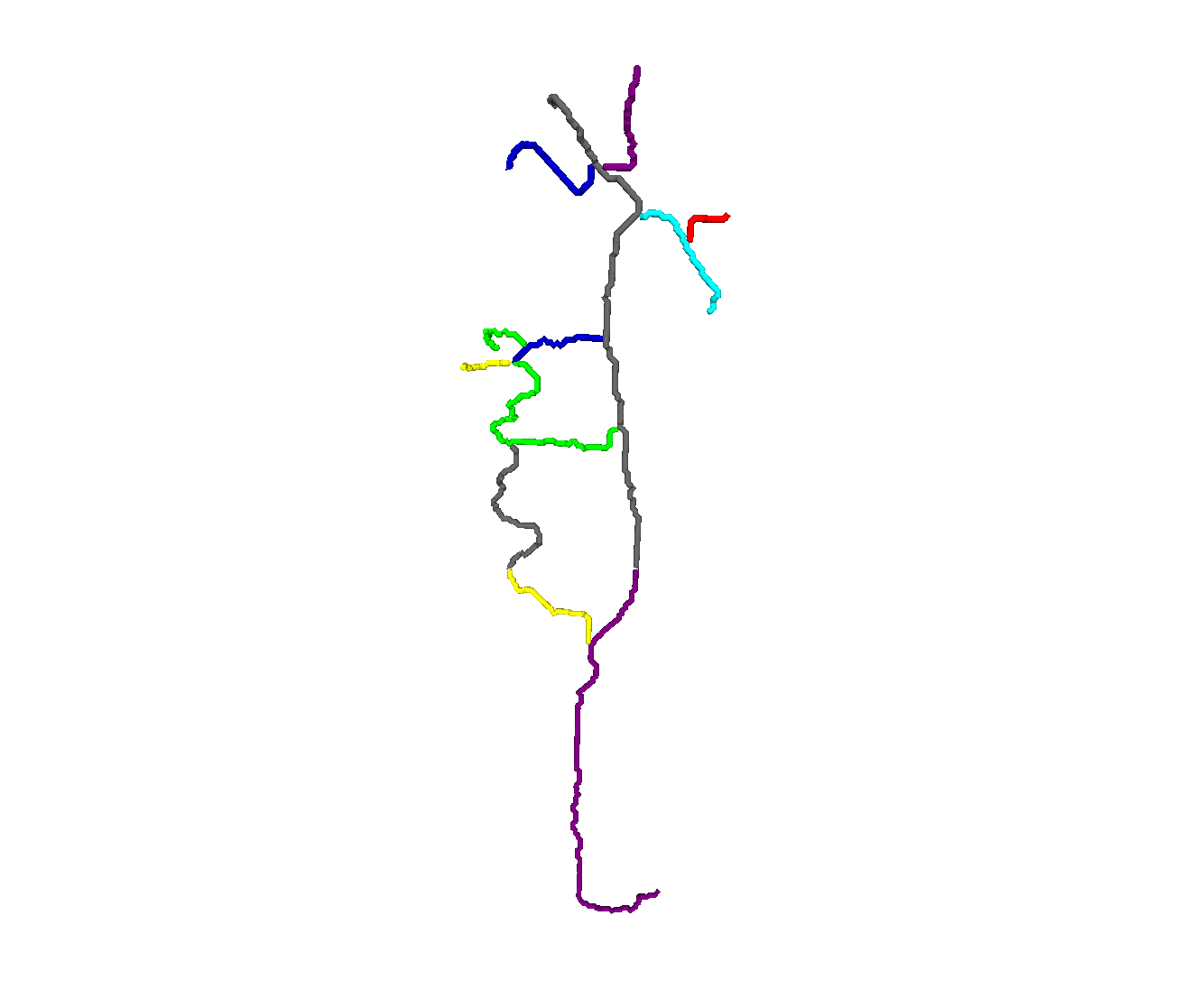}
\label{sf:seahorses}
}
\caption{Original surfaces and skeletons computed with our method.}
\label{fig:model_fourth}
\end{figure*}

\subsection{Influence of parameter $t$ on results }

We performed some experiments with the one user-supplied threshold, $t$ in our method, and show results in Figure \ref{fig:comparison_t}.  When $t = 1$, no segments are discarded. The resulting curve skeleton resembles a more dense version of the thinning result. When $t = 0.0001$, the result resembles the Jin \etalspace result except that loops are preserved. All results generated in this paper for the proposed method, except in this section, were generated with $t = 1e^{-12}$.  Consequently, different values of $t$ may be chosen depending on the application.

\newcommand*{\factorcomp}{0.4}
\begin{figure*}
\centering
\subfloat[Original surface]  
	{  
	\includegraphics[trim = 0cm 0cm 0cm 0cm,clip, width=\factorcomp\linewidth]{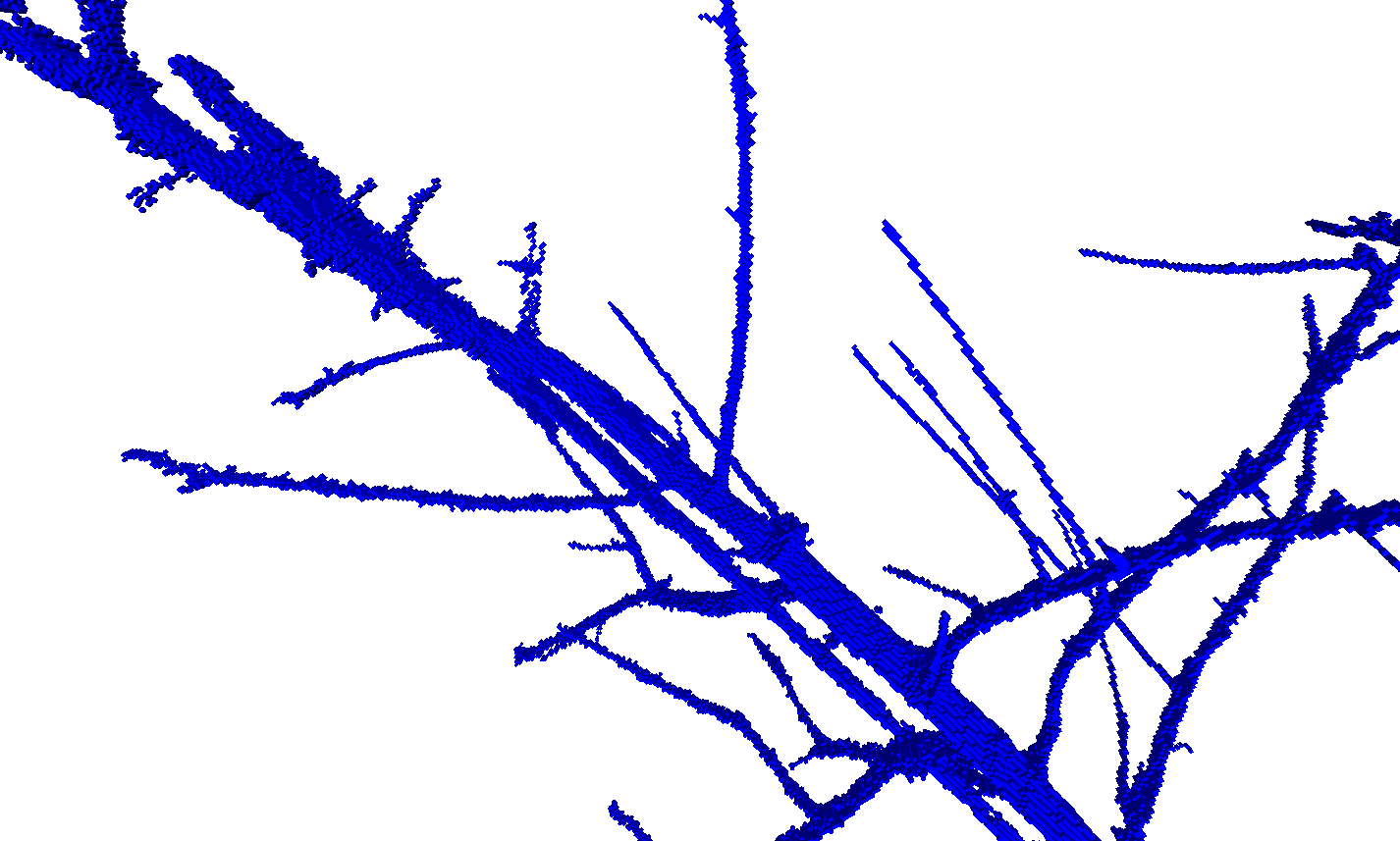}
}
\subfloat[Our method, $t = 1$]  
	{  
	\includegraphics[trim = 0cm 0cm 0cm 0cm,clip, width=\factorcomp\linewidth]{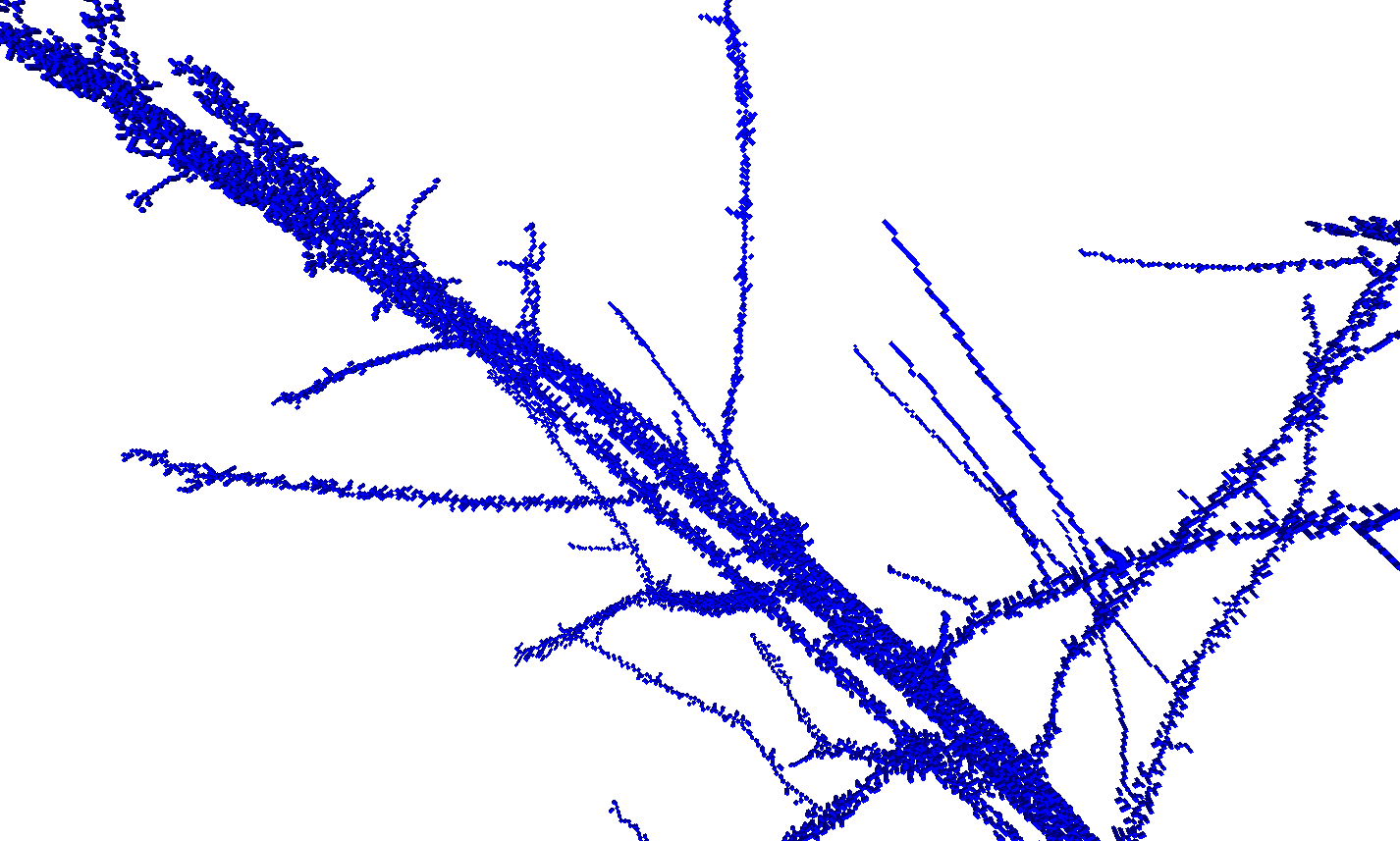}
}

\subfloat[Our method, $t = 0.0001$]  
	{  
	\includegraphics[trim = 0cm 0cm 0cm 0cm,clip, width=\factorcomp\linewidth]{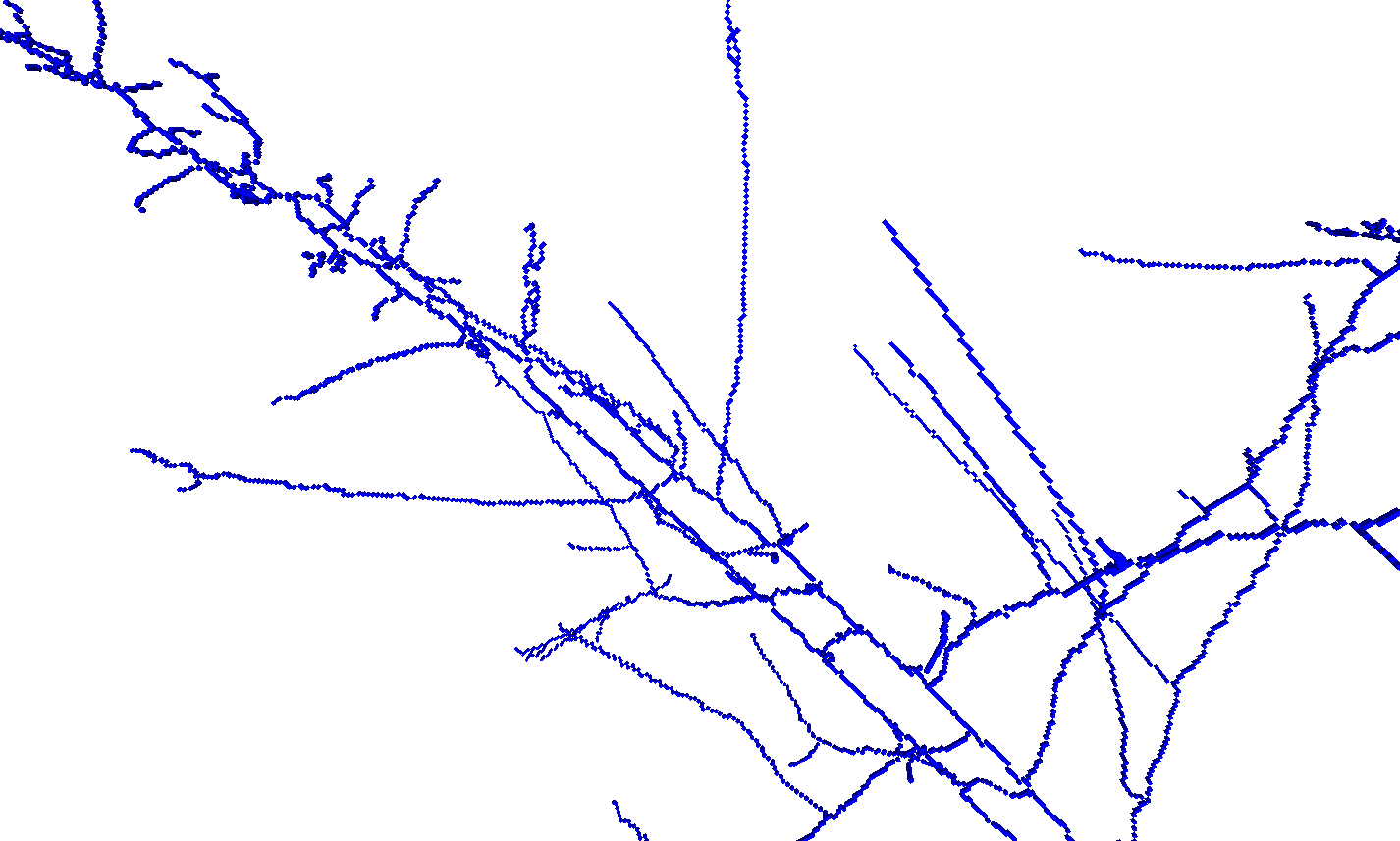}
}
\subfloat[Our method, $t= 1e^{-12}$]  
	{  
	\includegraphics[trim = 0cm 0cm 0cm 0cm,clip, width=\factorcomp\linewidth]{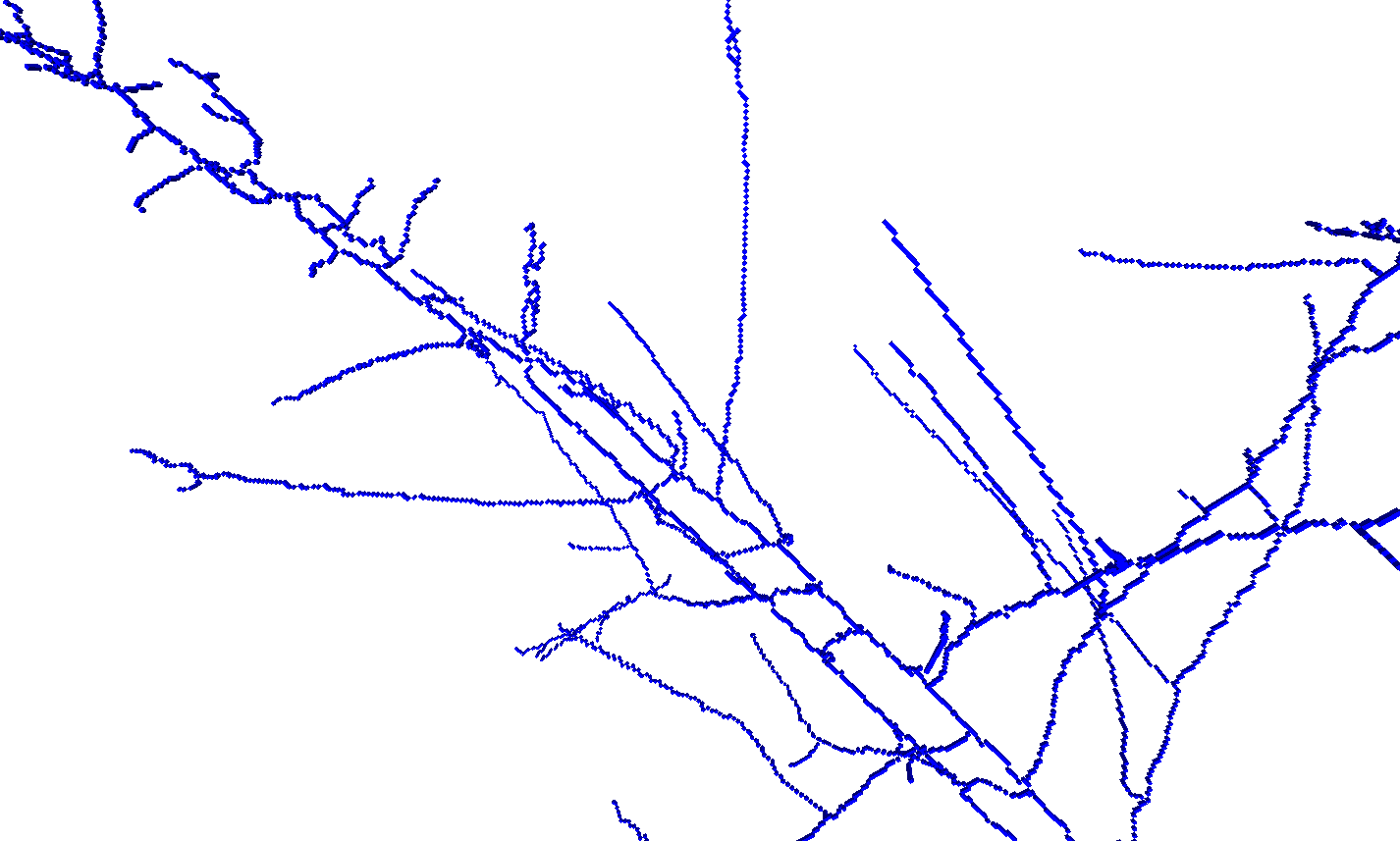}
}
\caption{ Original surface from Dataset B and curve skeletons computed with our method, with varying values of $t$.}
\label{fig:comparison_t}
\end{figure*}

\section{Additional results without comparisons}

To further illustrate the performance of our approach, Figures \ref{fig:plumI} to \ref{fig:peachIII} show additional high resolution images of results obtained using our method.

\newcommand*{\factorExtras}{0.4}
\begin{figure*}
\centering
\subfloat[Surface]  
	{  
	\includegraphics[trim = 0cm 0cm 0cm 0cm,clip, width=\factorExtras\linewidth]{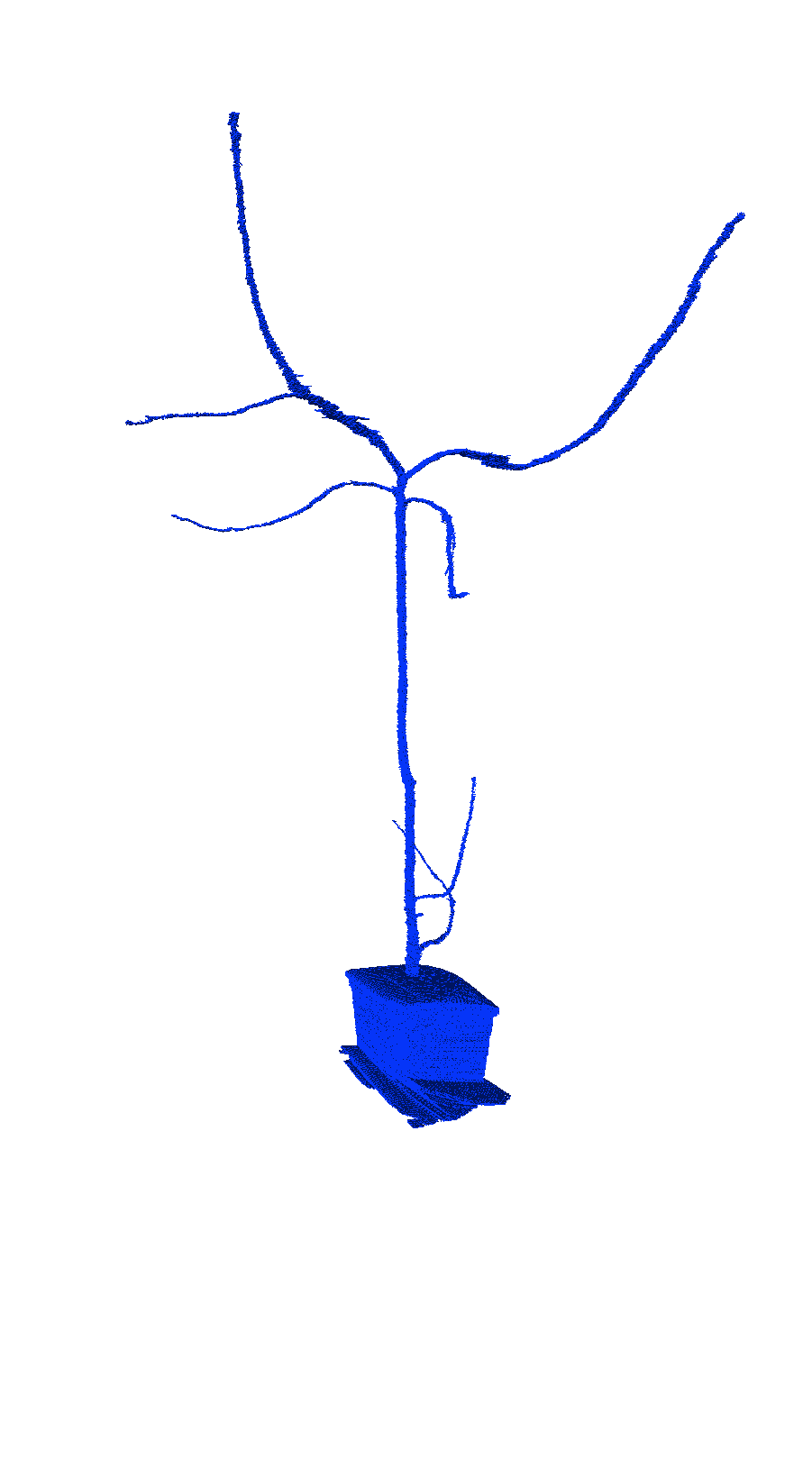}
}
\subfloat[Our method]  
	{  
	\includegraphics[trim = 0cm 0cm 0cm 0cm,clip, width=\factorExtras\linewidth]{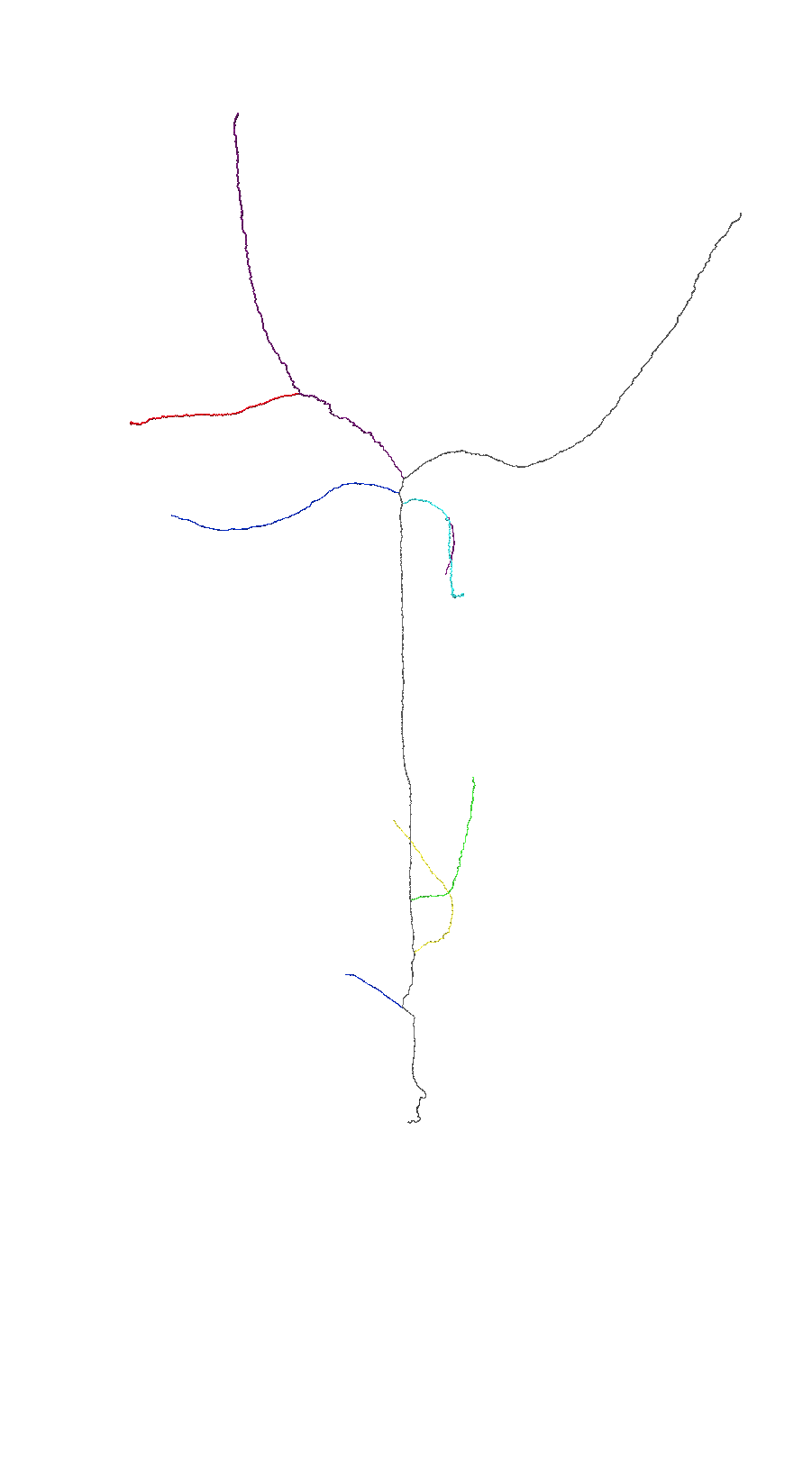}
}
\caption{\textbf{Best viewed in color.} Original surface and curve skeleton computed with our method.}
\label{fig:plumI}
\end{figure*}

\begin{figure*}
\centering
\subfloat[Surface]  
	{  
	\includegraphics[trim = 0cm 0cm 0cm 0cm,clip, width=\factorExtras\linewidth]{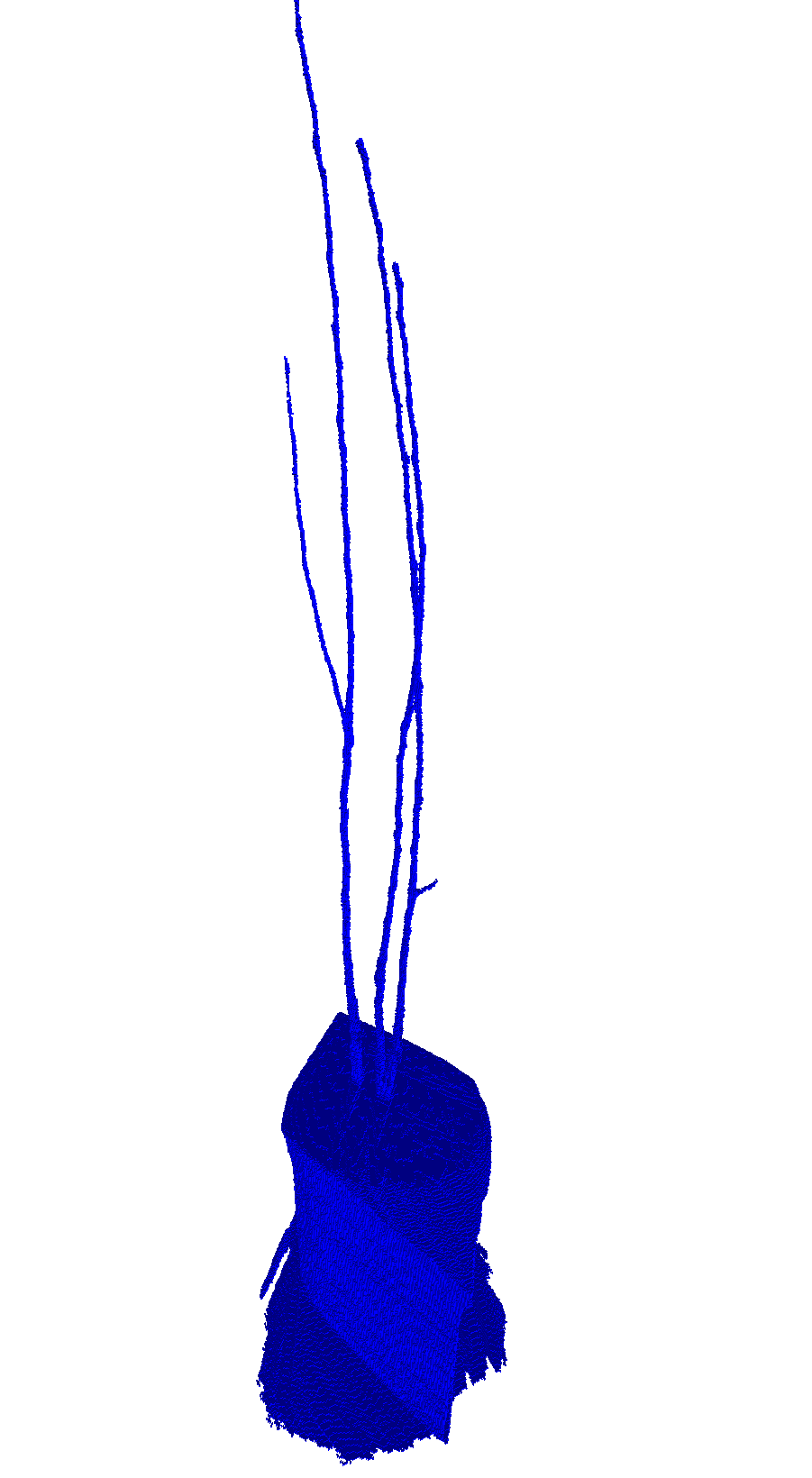}
}
\subfloat[Our method]  
	{  
	\includegraphics[trim = 0cm 0cm 0cm 0cm,clip, width=\factorExtras\linewidth]{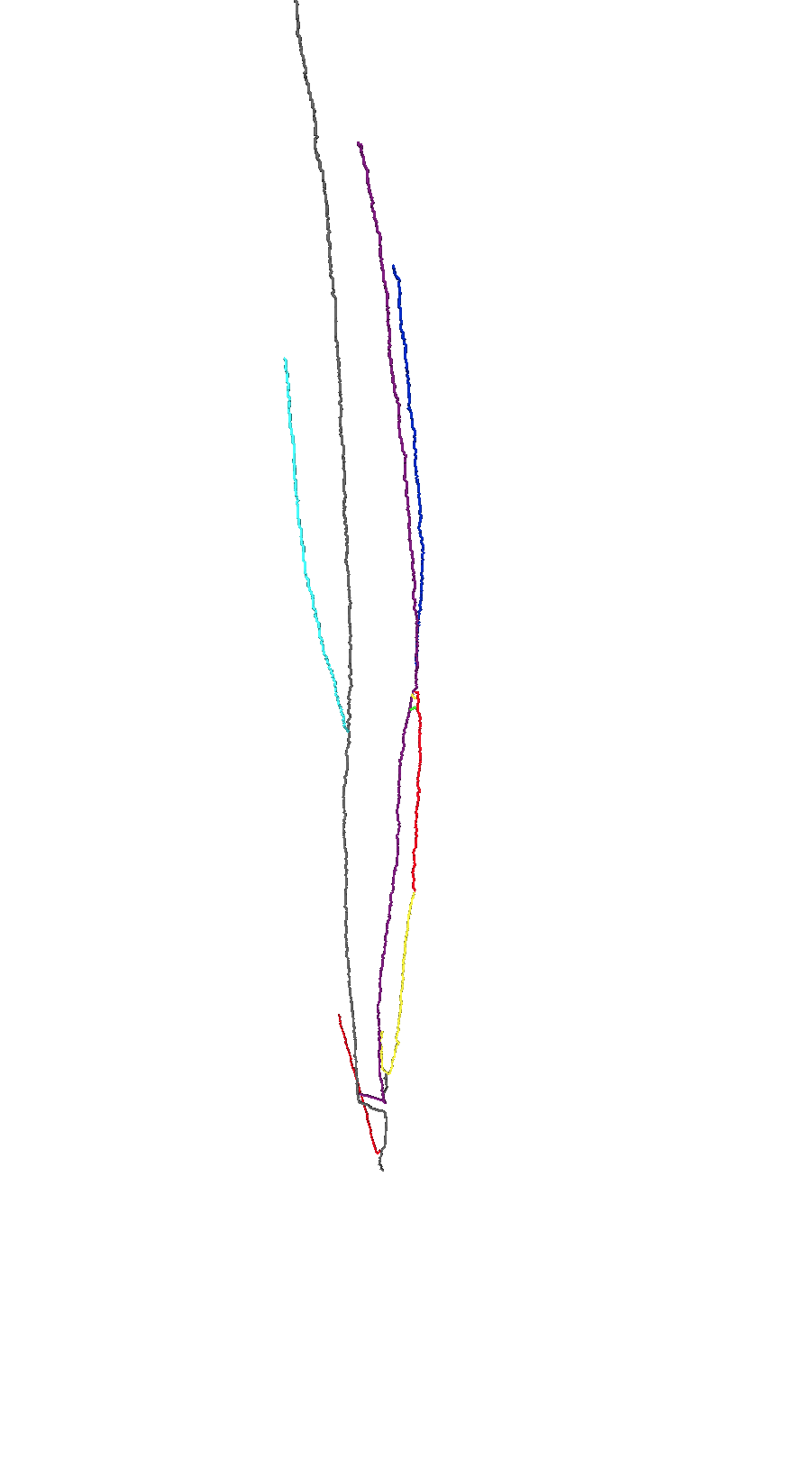}
}
\caption{\textbf{Best viewed in color.} Original surface and curve skeleton computed with our method.}
\label{fig:plumIII}
\end{figure*}

\begin{figure*}
\centering
\subfloat[Surface]  
	{  
	\includegraphics[trim = 0cm 0cm 0cm 0cm,clip, width=\factorExtras\linewidth]{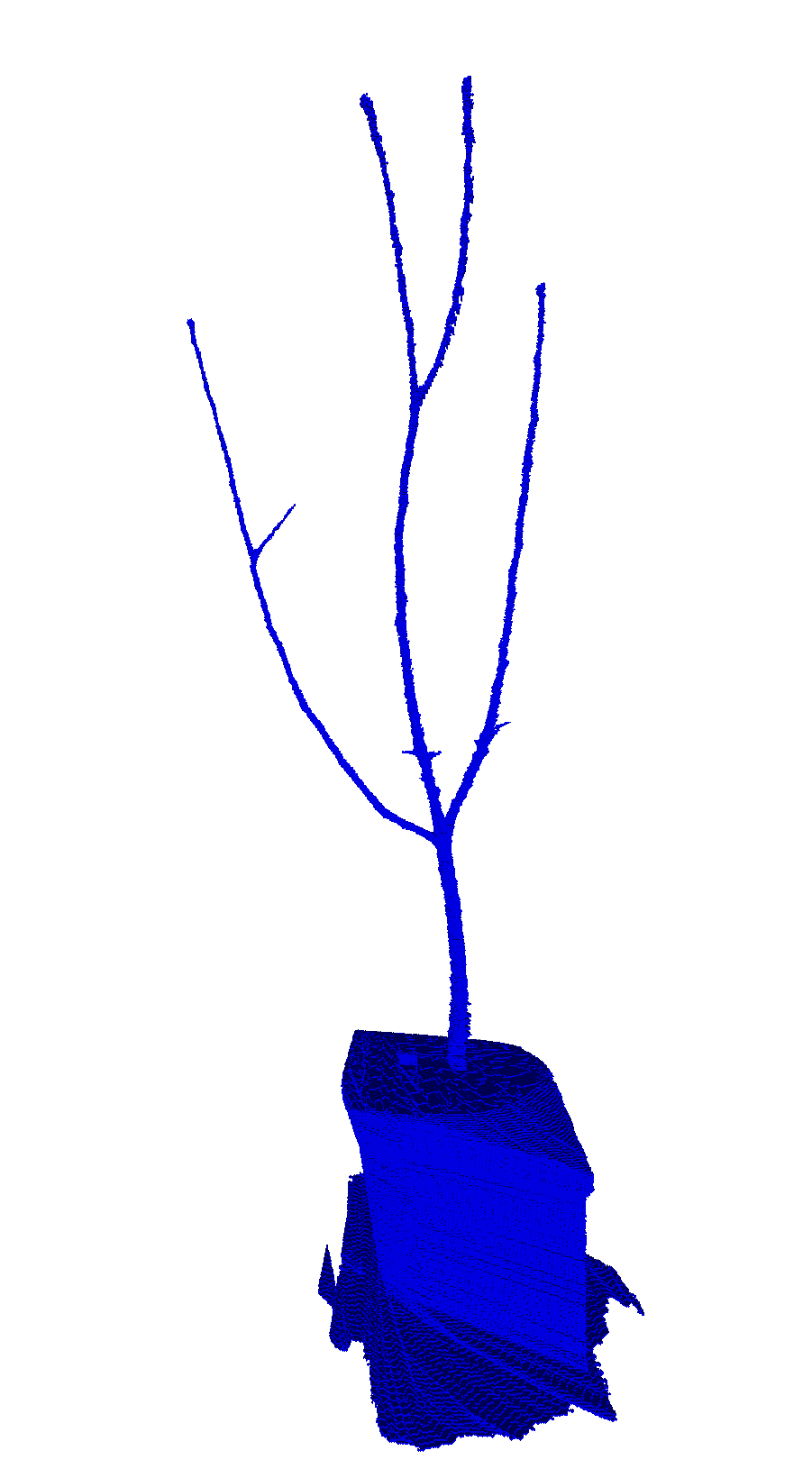}
}
\subfloat[Our method]  
	{  
	\includegraphics[trim = 0cm 0cm 0cm 0cm,clip, width=\factorExtras\linewidth]{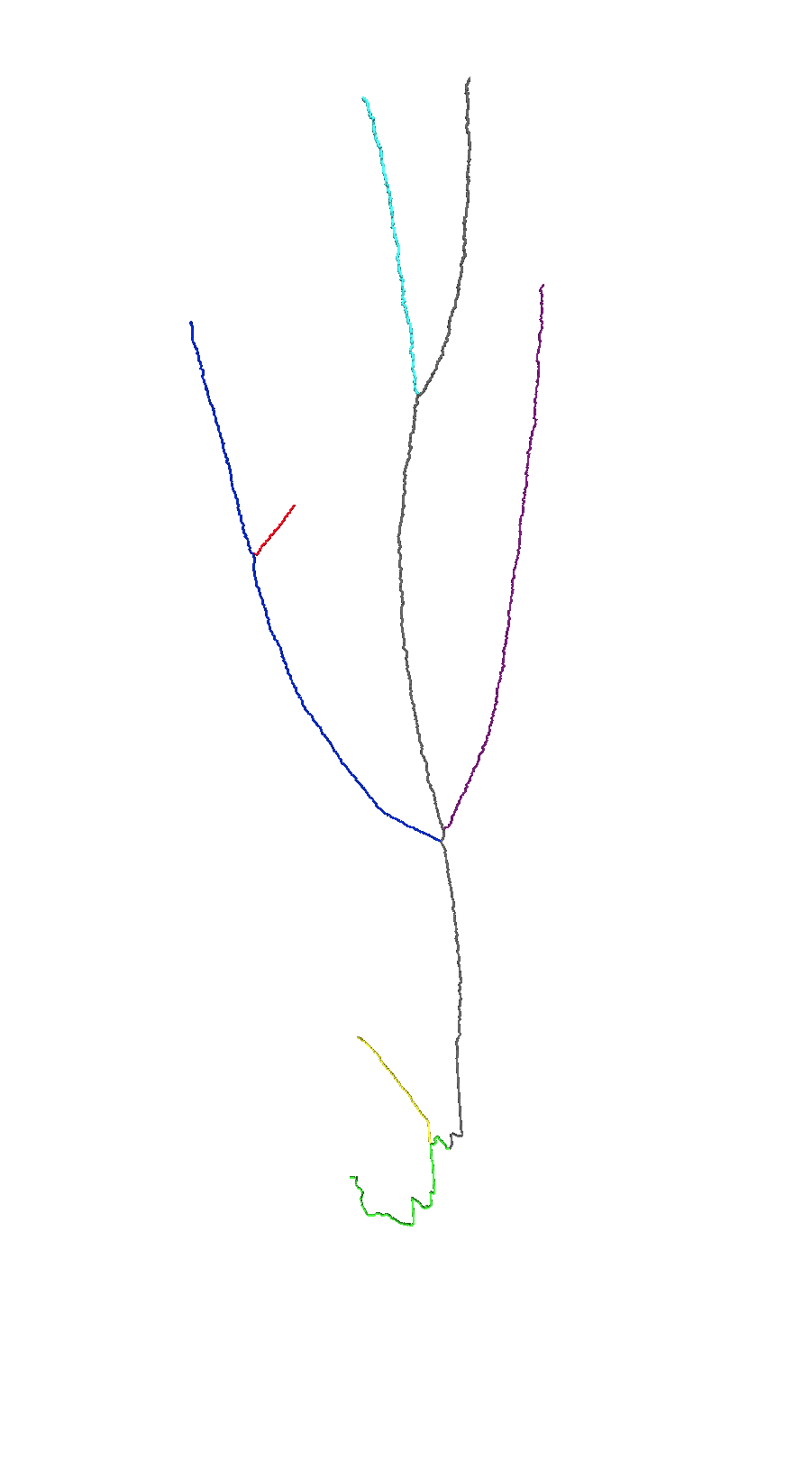}
}
\caption{\textbf{Best viewed in color.} Original surface and curve skeleton computed with our method.}
\label{fig:plumV}
\end{figure*}

\begin{figure*}
\centering
\subfloat[Surface]  
	{  
	\includegraphics[trim = 0cm 0cm 0cm 0cm,clip, width=\factorExtras\linewidth]{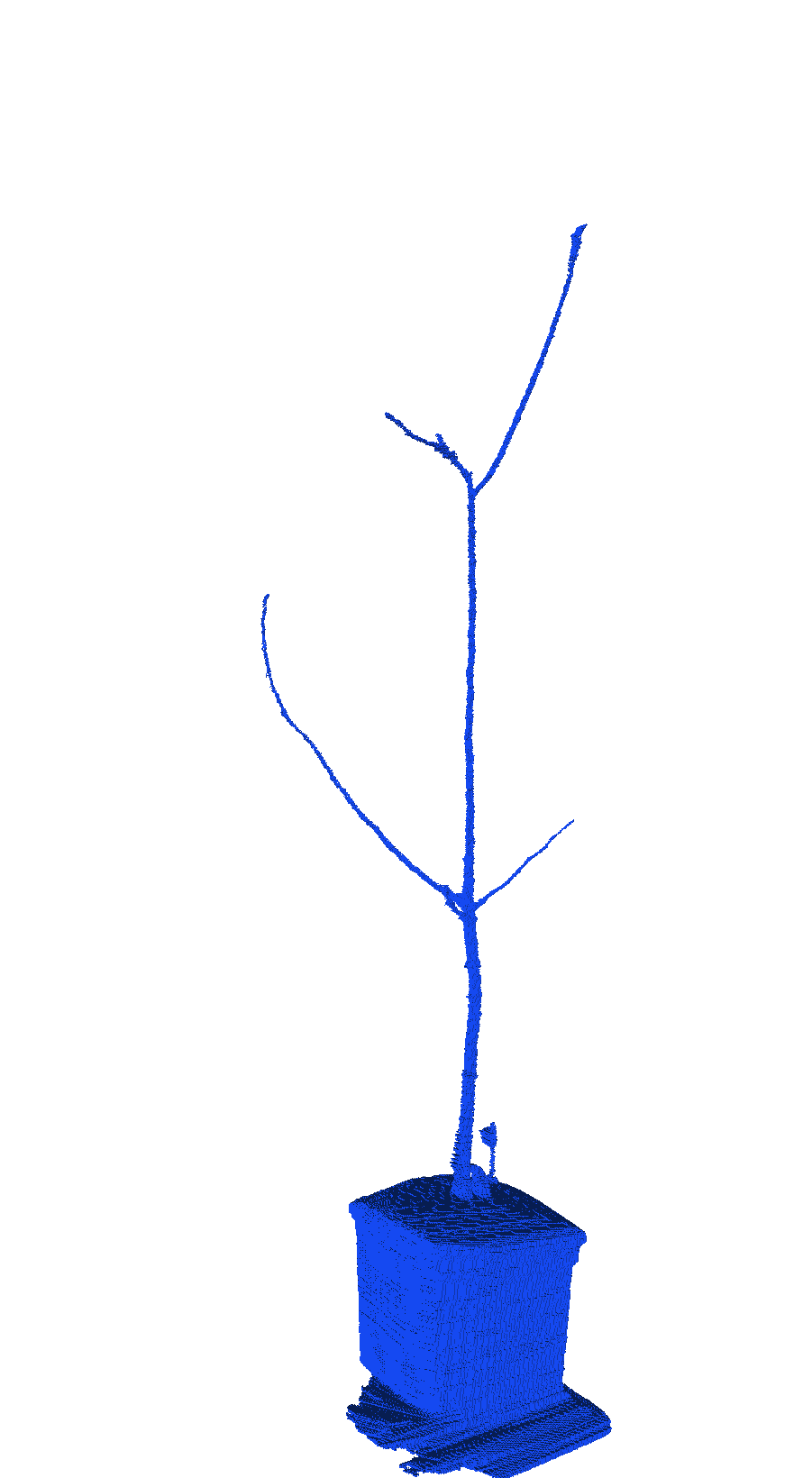}
}
\subfloat[Our method]  
	{  
	\includegraphics[trim = 0cm 0cm 0cm 0cm,clip, width=\factorExtras\linewidth]{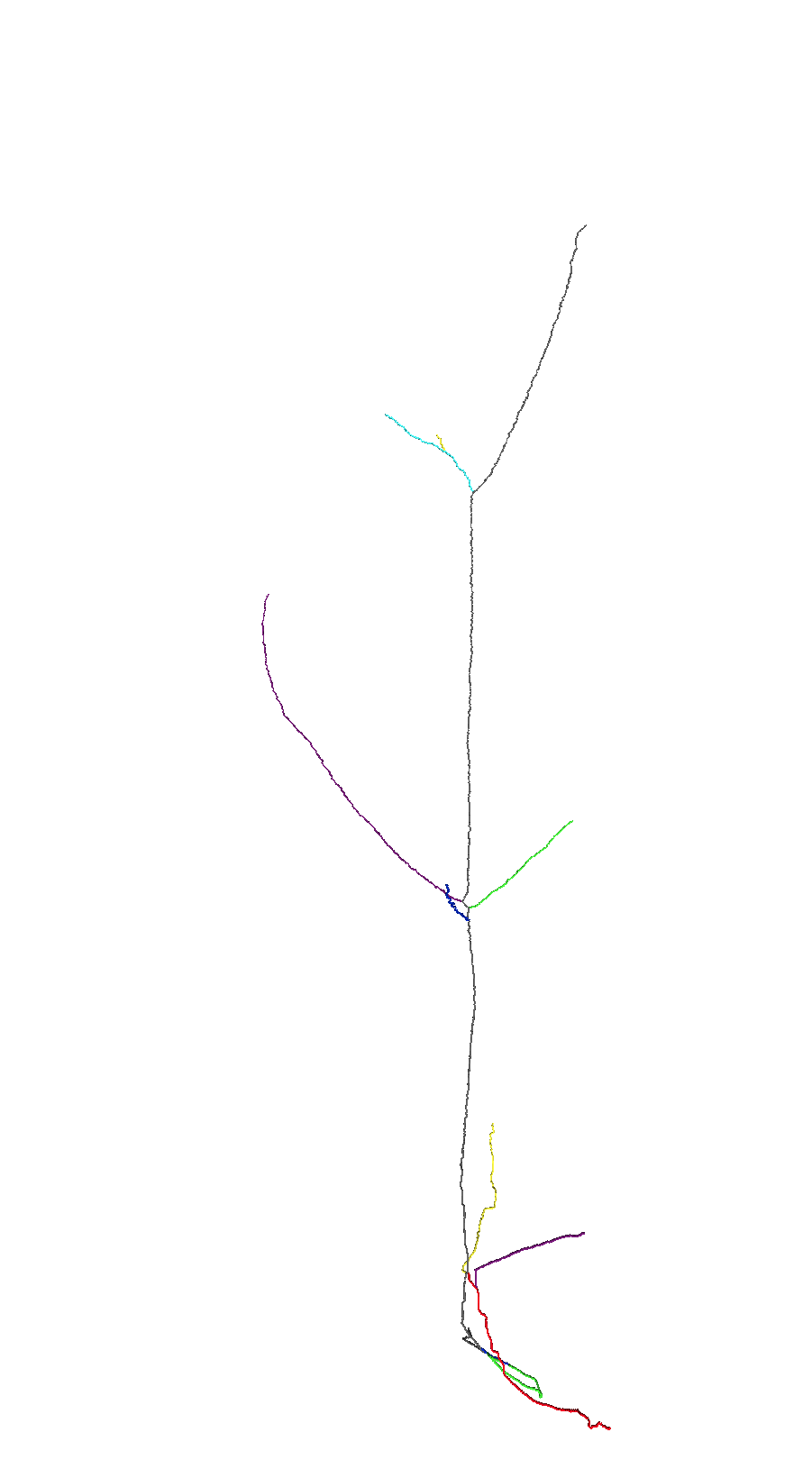}
}
\caption{\textbf{Best viewed in color.} Original surface and curve skeleton computed with our method.}
\label{fig:plumVI}
\end{figure*}

\begin{figure*}
\centering
\subfloat[Surface]  
	{  
	\includegraphics[trim = 0cm 0cm 0cm 0cm,clip, width=\factorExtras\linewidth]{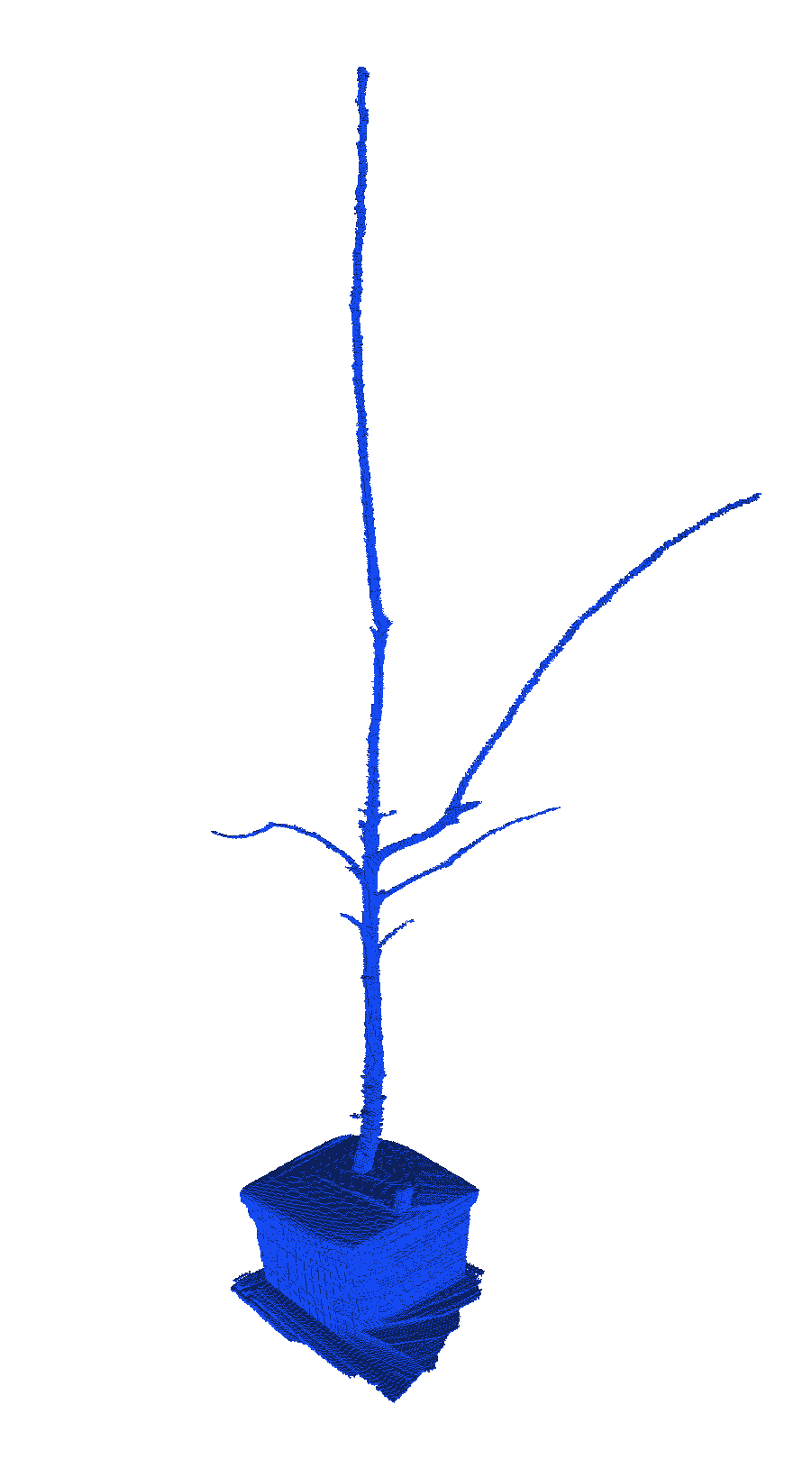}
}
\subfloat[Our method]  
	{  
	\includegraphics[trim = 0cm 0cm 0cm 0cm,clip, width=\factorExtras\linewidth]{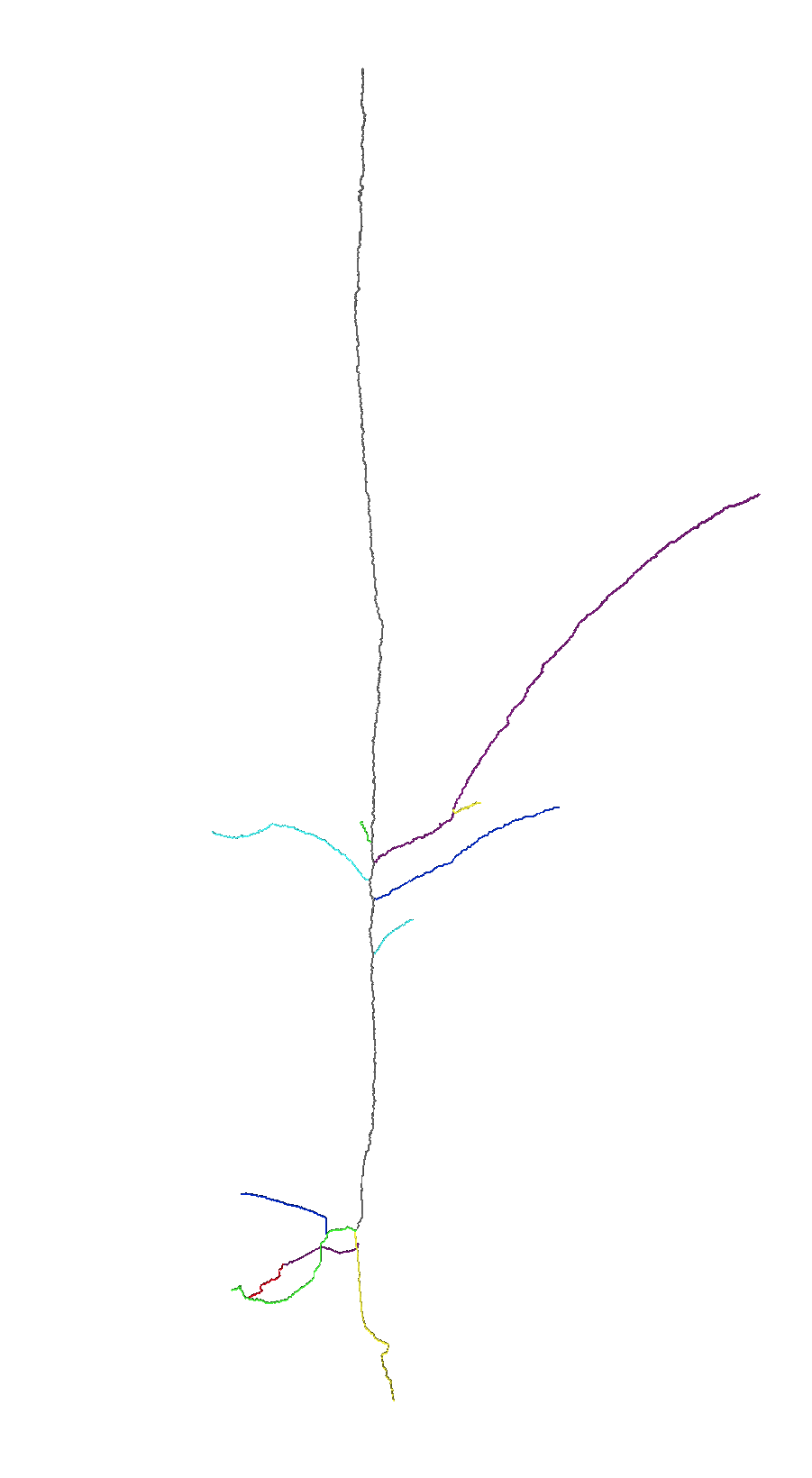}
}
\caption{\textbf{Best viewed in color.} Original surface and curve skeleton computed with our method.}
\label{fig:plumVII}
\end{figure*}

%\begin{landscape}
\begin{figure*}
\centering
\subfloat[Surface]  
	{  
	\includegraphics[trim = 0cm 0cm 0cm 0cm,clip, width=\factorExtras\linewidth]{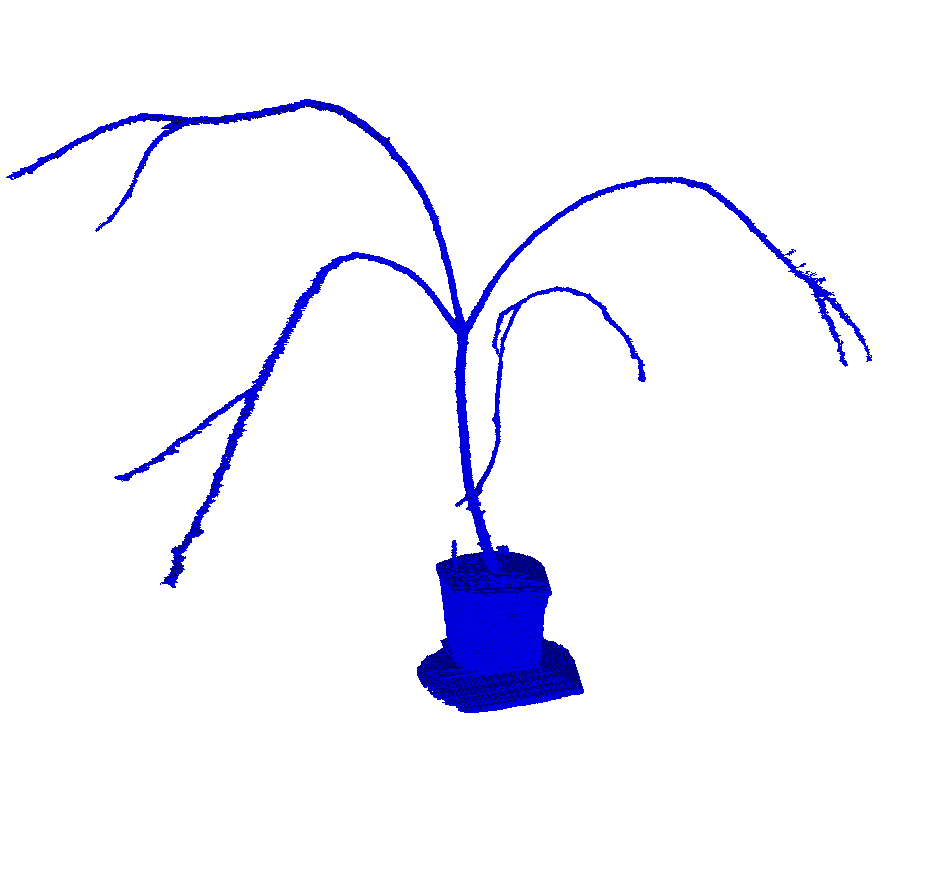}
}
\subfloat[Our method]  
	{  
	\includegraphics[trim = 0cm 0cm 0cm 0cm,clip, width=\factorExtras\linewidth]{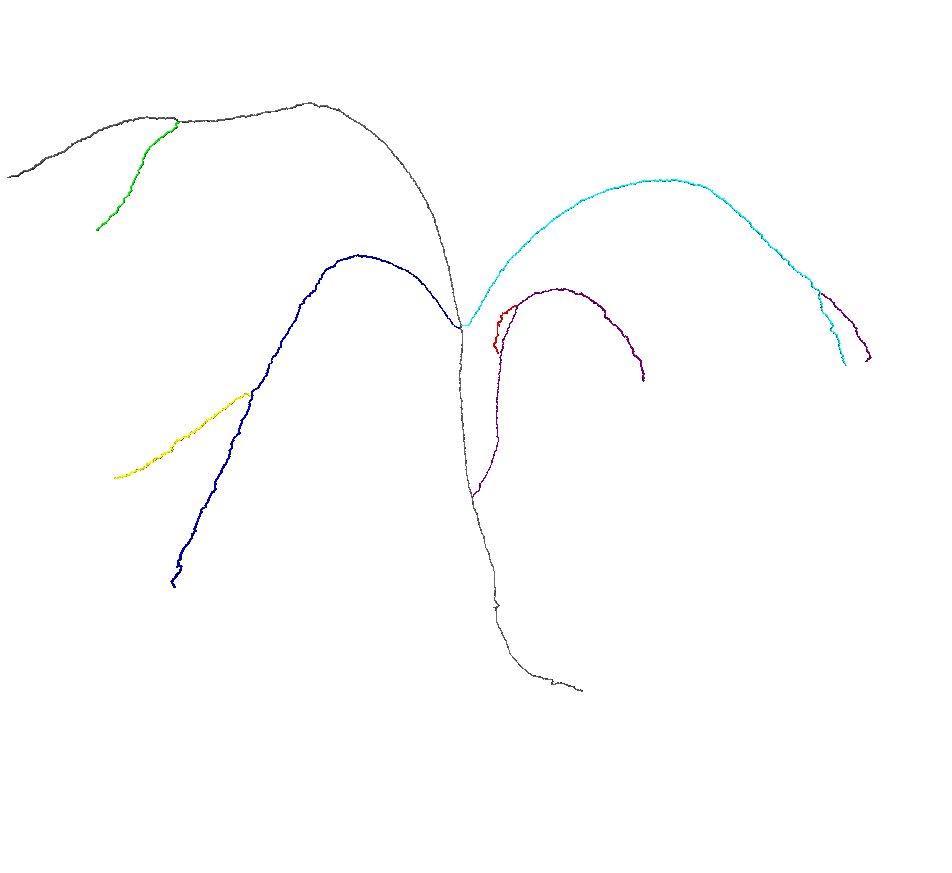}
}
\caption{\textbf{Best viewed in color.} Original surface and curve skeleton computed with our method.}
\label{fig:plumVIII}
\end{figure*}
%\end{landscape}

\begin{figure*}
\centering
\subfloat[Surface]  
	{  
	\includegraphics[trim = 0cm 0cm 0cm 0cm,clip, width=\factorExtras\linewidth]{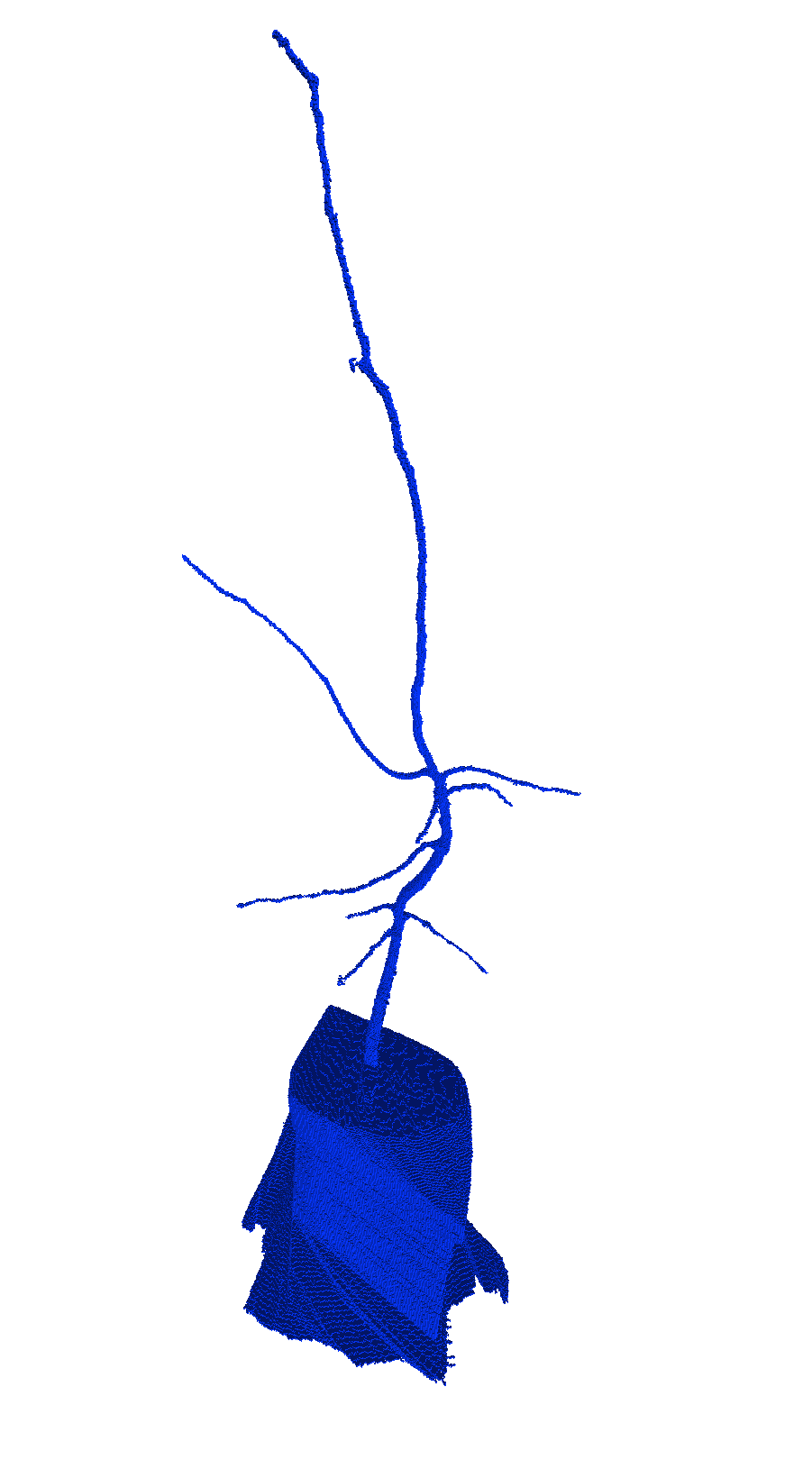}
}
\subfloat[Our method]  
	{  
	\includegraphics[trim = 0cm 0cm 0cm 0cm,clip, width=\factorExtras\linewidth]{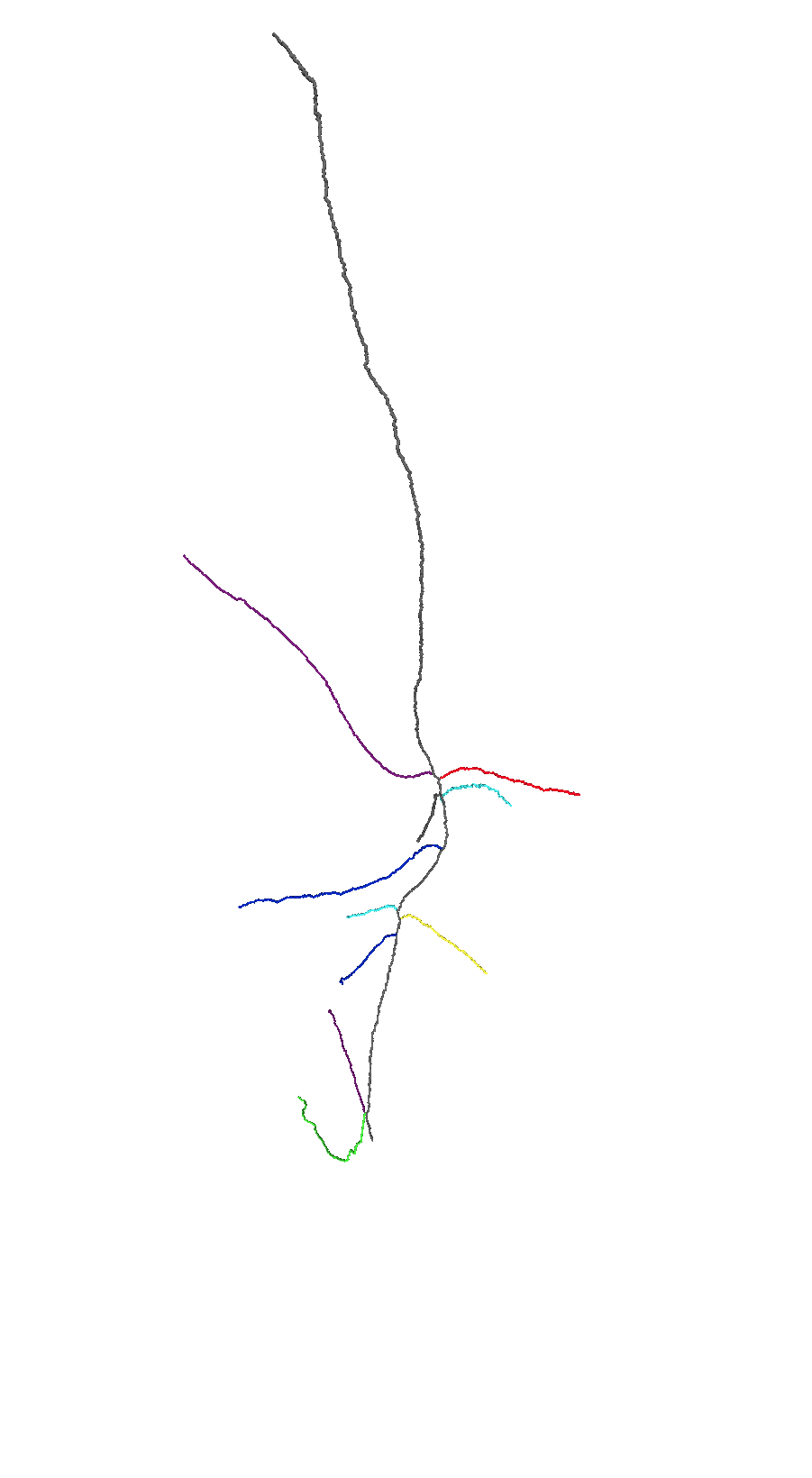}
}
\caption{\textbf{Best viewed in color.} Original surface and curve skeleton computed with our method.}
\label{fig:plumIX}
\end{figure*}

\begin{figure*}
\centering
\subfloat[Surface]  
	{  
	\includegraphics[trim = 0cm 0cm 0cm 0cm,clip, width=\factorExtras\linewidth]{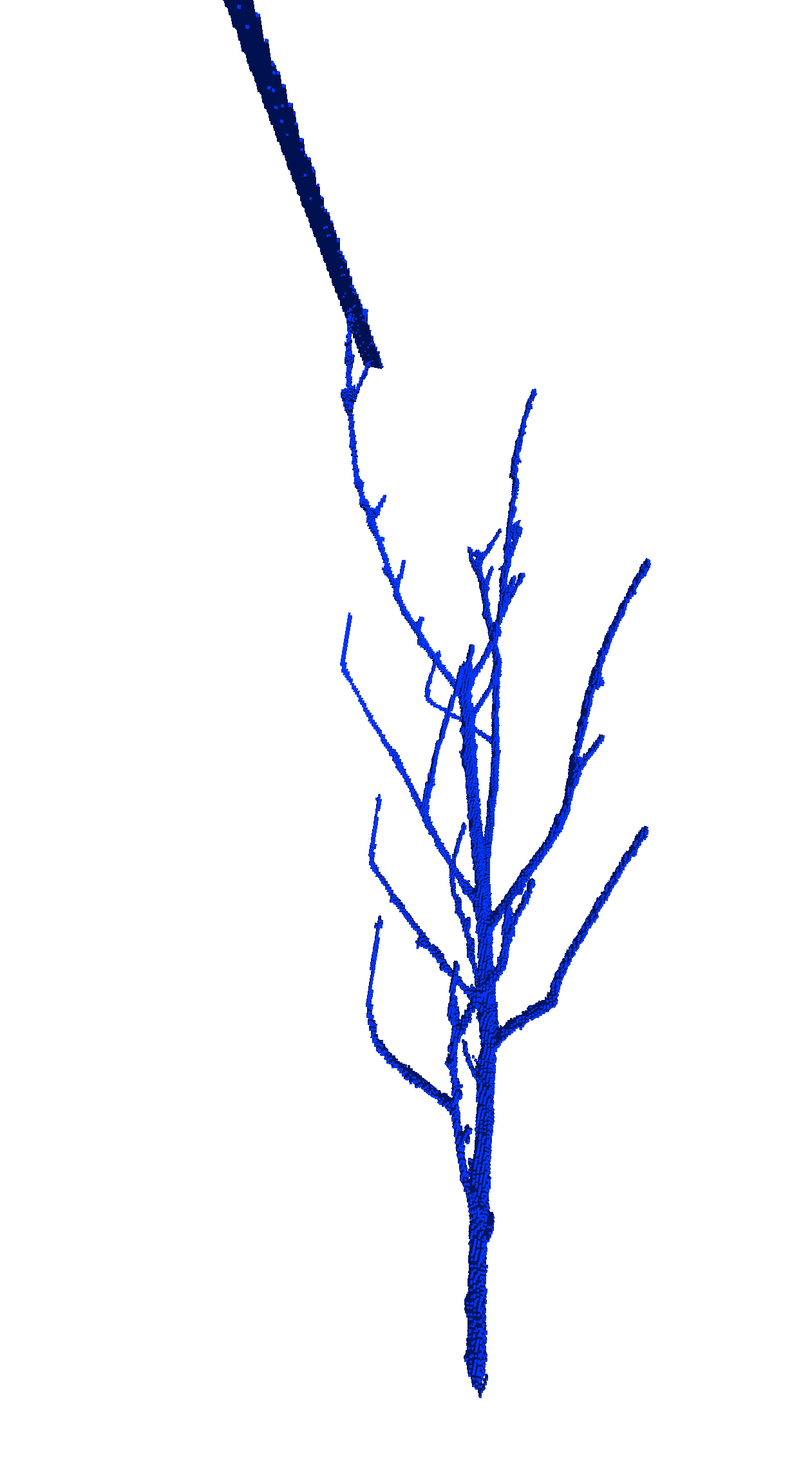}
}
\subfloat[Our method]  
	{  
	\includegraphics[trim = 0cm 0cm 0cm 0cm,clip, width=\factorExtras\linewidth]{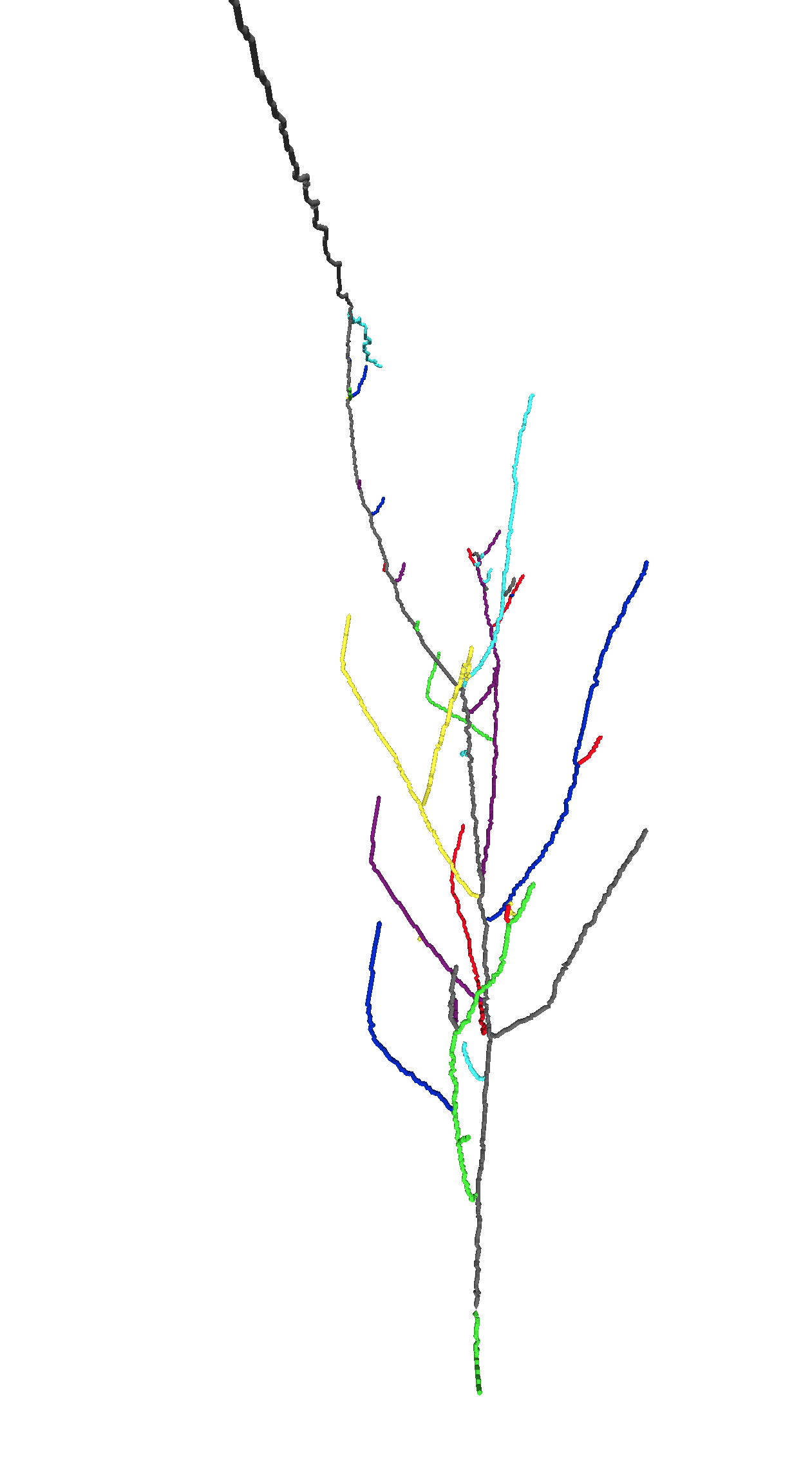}
}
\caption{\textbf{Best viewed in color.} Original surface and curve skeleton computed with our method.}
\label{fig:peachI}
\end{figure*}

%\begin{landscape}
\begin{figure*}
\centering
\subfloat[Surface]  
	{  
	\includegraphics[trim = 0cm 0cm 0cm 0cm,clip, width=\factorExtras\linewidth]{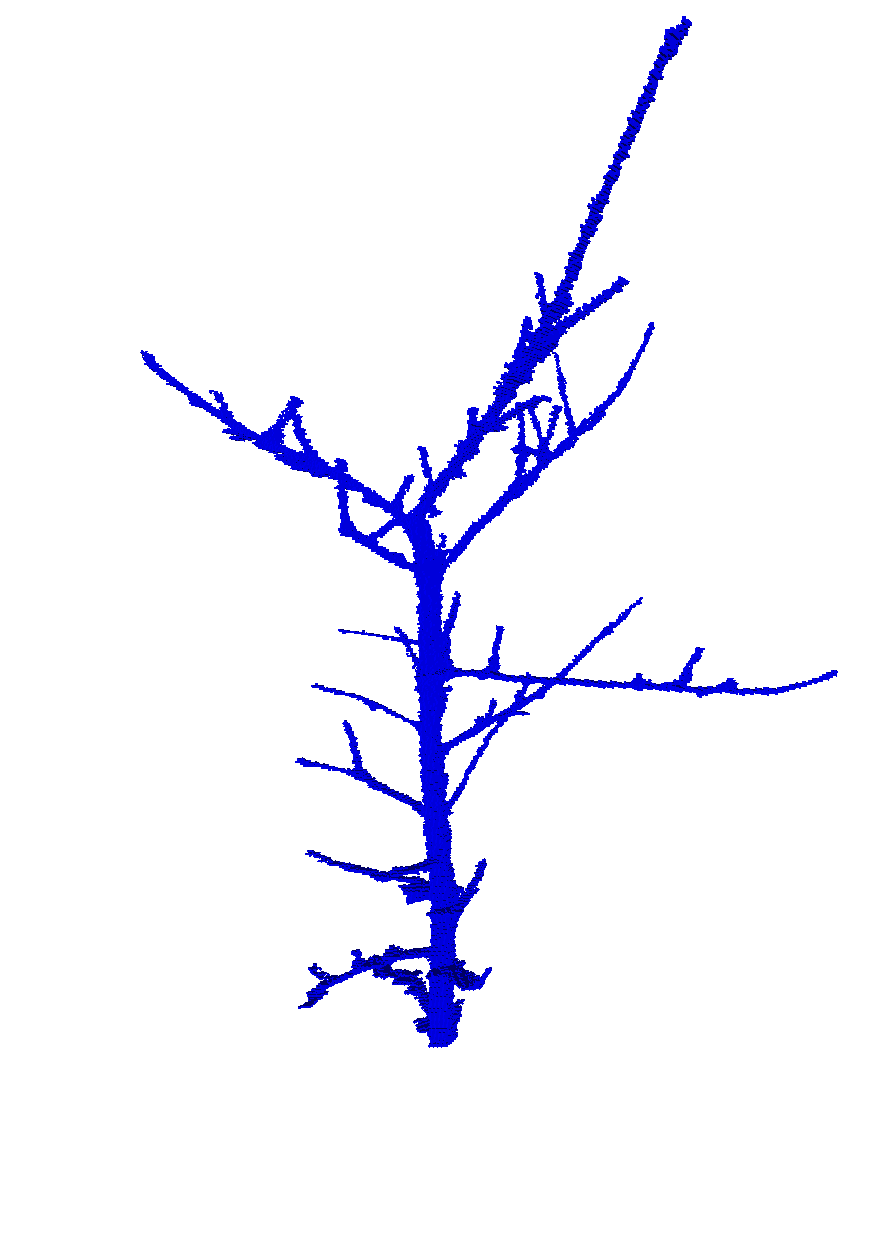}
}
\subfloat[Our method]  
	{  
	\includegraphics[trim = 0cm 0cm 0cm 0cm,clip, width=\factorExtras\linewidth]{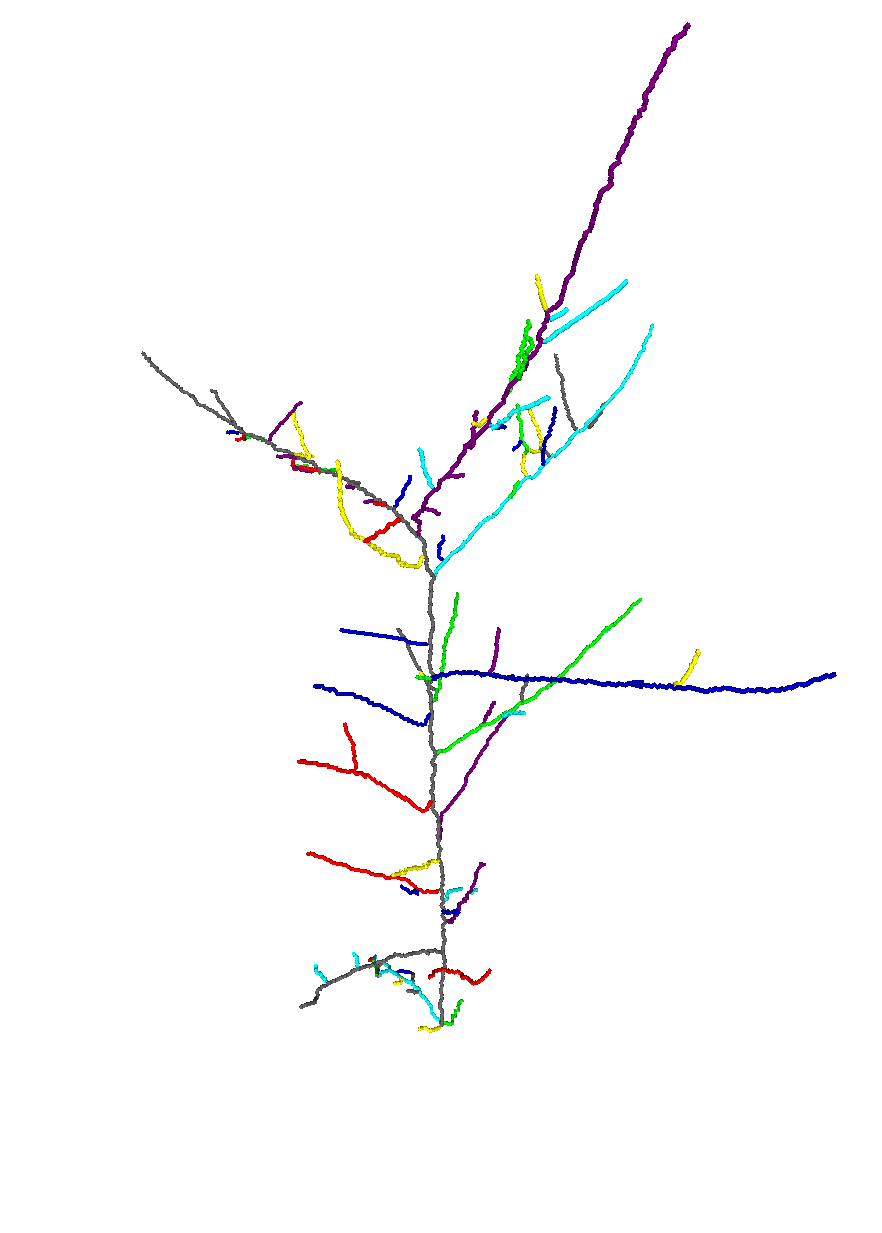}
}
\caption{\textbf{Best viewed in color.} Original surface and curve skeleton computed with our method.}
\label{fig:peachII}
\end{figure*}

\begin{figure*}
\centering
\subfloat[Surface]  
	{  
	\includegraphics[trim = 0cm 0cm 0cm 0cm,clip, width=\factorExtras\linewidth]{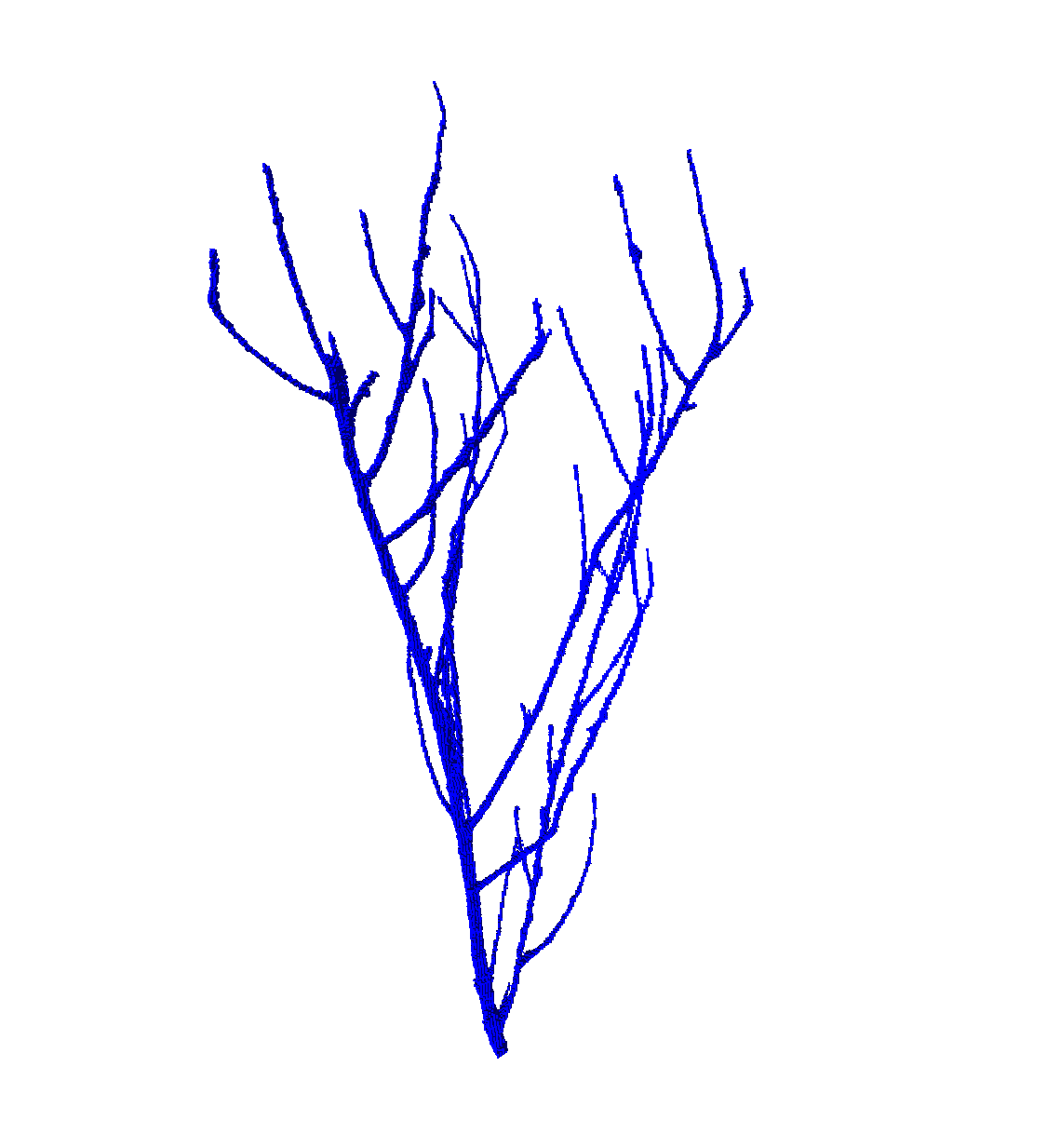}
}
\subfloat[Our method]  
	{  
	\includegraphics[trim = 0cm 0cm 0cm 0cm,clip, width=\factorExtras\linewidth]{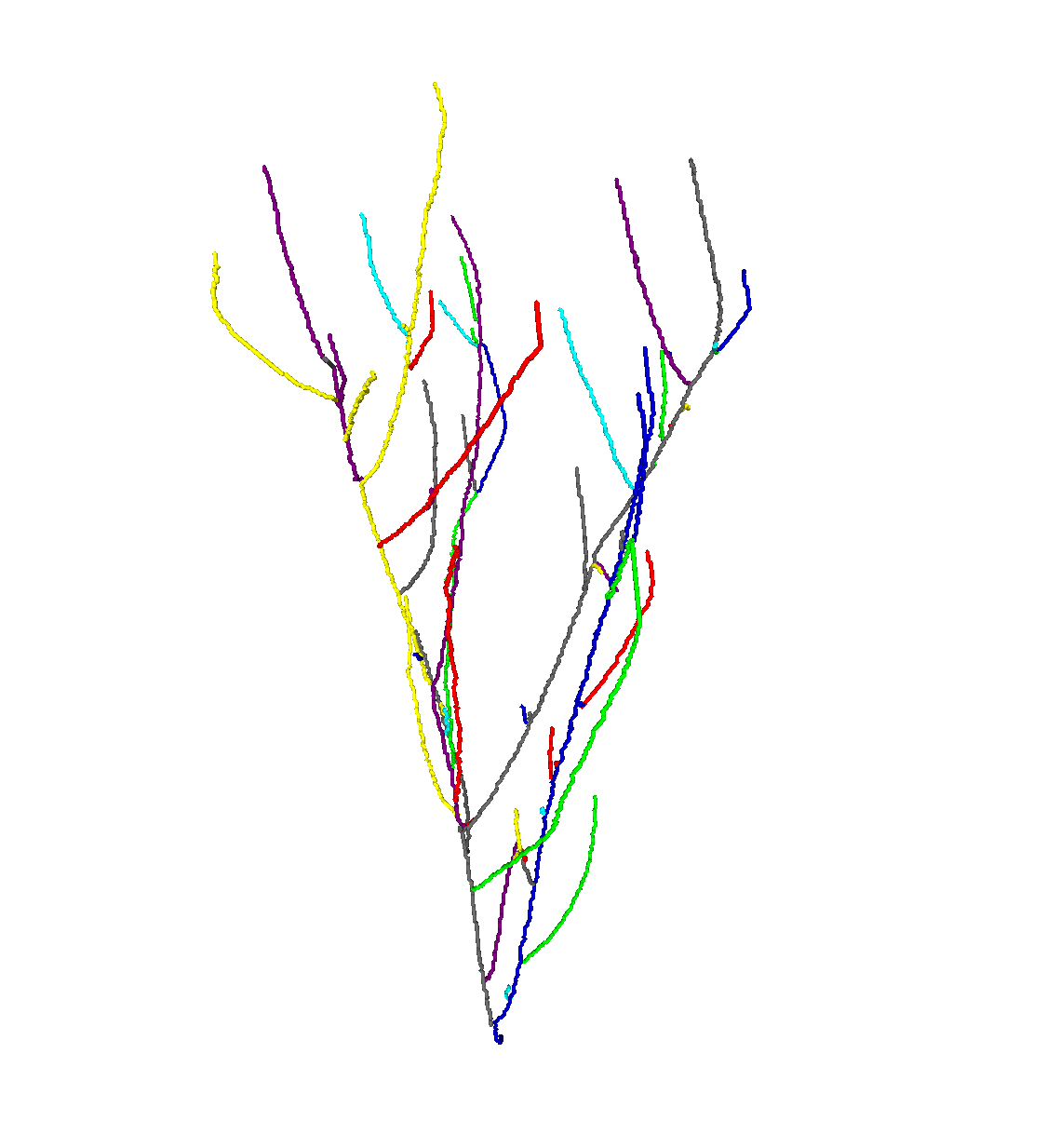}
}
\caption{\textbf{Best viewed in color.} Original surface and curve skeleton computed with our method.}
\label{fig:peachIII}
\end{figure*}

%\clearpage

%{\small
%\bibliographystyle{ieee}
%\bibliography{curveskeleton_supp}
%}

\end{document}